\pdfoutput=1
\documentclass[11pt]{article}

\usepackage[T1]{fontenc}
\usepackage{amsmath,amssymb,amsthm}
\usepackage{bm}
\usepackage{booktabs}
\usepackage{array}
\usepackage{longtable}
\usepackage{graphicx}
\usepackage{float}
\usepackage{multirow}
\usepackage{microtype}
\usepackage[dvipsnames]{xcolor}
\usepackage{algorithm}
\usepackage{algpseudocode}
\usepackage{caption}
\usepackage{subcaption}

\definecolor{hunyuanblue}{HTML}{1E4A8F}
\definecolor{mygreen}{HTML}{2E8B57}

\usepackage{hunyuan}

\usepackage[capitalize,noabbrev]{cleveref}
\usepackage[most]{tcolorbox}
\usepackage{fancyhdr}
\usepackage{xurl}
\usepackage{etoc}              
\usepackage{fontawesome5}
\usepackage{twemojis}

\crefname{appendix}{Appendix}{Appendices}
\Crefname{appendix}{Appendix}{Appendices}
\numberwithin{theorem}{section}

\makeatletter
\let\c@example\relax

\makeatother
\theoremstyle{definition}
\newtheorem{example}{Example}
\theoremstyle{plain}
\AddToHook{env/corollary/begin}{%
  \crefalias{theorem}{corollary}%
  \crefalias{section}{corollary}%
}
\newcommand{\refex}[1]{Example~\ref{#1}}
\renewcommand{\eqref}[1]{\text{Equation}~(\ref{#1})}

\makeatletter
\long\def\@makecaption#1#2{%
  \vskip 10pt
  \setbox\@tempboxa\hbox{#1: #2}%
  \ifdim \wd\@tempboxa >\hsize
    \noindent #1: #2\par
  \else
    \hbox to\hsize{\hfil\box\@tempboxa\hfil}%
  \fi}
\makeatother

\hypersetup{
  colorlinks=true,
  linkcolor=hunyuanblue,
  citecolor=hunyuanblue,
  urlcolor=hunyuanblue,
}

\makeatletter
\def\section{\@startsiction{section}{1}{\z@}{-0.24in}{0.10in}
             {\large\bf\raggedright\color{hunyuanblue}}}
\def\subsection{\@startsection{subsection}{2}{\z@}{-0.20in}{0.08in}
                {\normalsize\bf\raggedright\color{hunyuanblue}}}
\makeatother

\fancypagestyle{plain}{%
  \fancyhf{}%
  \fancyfoot[C]{\thepage}%
}
\fancypagestyle{firststyle}{%
  \fancyhf{}%
  \fancyhead[L]{%
    \IfFileExists{figs/hunyuan-logo.png}
      {\raisebox{-18.5mm}[0pt][0pt]{\includegraphics[height=9mm]{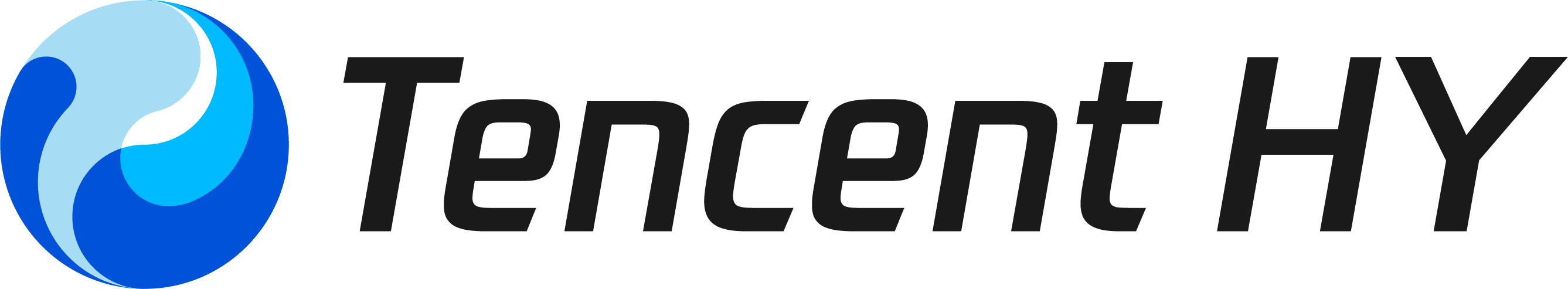}}}
      {\raisebox{-14mm}[0pt][0pt]{\textcolor{hunyuanblue}{\sffamily\bfseries HUNYUAN}}}}%
  \fancyfoot[C]{\thepage}%
}

\makeatletter
\newcommand{\appendixtitle}{%
  \clearpage
  \thispagestyle{plain}%
  \vbox{%
    \hsize\textwidth
    \linewidth\hsize
    \vskip 0.1in
    {\color{hunyuanblue}\hrule height 0.6pt}%
    \vskip 4mm
    \centering
    {\LARGE\bfseries Appendix\par}%
    \vskip 4mm
    {\color{hunyuanblue}\hrule height 0.6pt}%
    \vskip 0.3in\@minus0.1in
  }%
}
\makeatother

\definecolor{abstractbg}{HTML}{F0F7FC}

\usepackage[frozencache,cachedir=minted-cache]{minted}
\tcbuselibrary{minted}
\setminted[python]{linenos, breaklines, fontsize=\footnotesize, xleftmargin=2em}

\definecolor{metablue}{HTML}{0064E0}
\definecolor{metafg}{HTML}{1C2B33}
\definecolor{metabg}{HTML}{F1F4F7}

\newtcolorbox[auto counter]{pbox}[2][]{
  colback=white,
  title={Code~\thetcbcounter: #2},
  #1, fonttitle=\sffamily, fontupper=\sffamily,
  arc=10pt, colframe=metabg, coltitle=metafg, colbacktitle=metabg,
  toptitle=0.25cm, bottomtitle=0.125cm
}
\crefname{tcb@cnt@pbox}{code}{code}
\Crefname{tcb@cnt@pbox}{Code}{Code}

\newcommand{\sg}{\operatorname{sg}}
\newcommand{\clip}{\operatorname{clip}}
\newcommand{\mean}{\operatorname{mean}}
\newcommand{\Rollout}{\textsc{Rollout}}
\newcommand{\AddNoise}{\textsc{AddNoise}}
\newcommand{\dec}{\mathcal{D}}
\newcommand{\xh}{\hat{x}}                         
\newcommand{\xhth}{\hat{x}_\theta}                 
\newcommand{\vph}{v_\phi}                         
\newcommand{\vcond}{v_{\mathrm{cond}}}            
\newcommand{\xt}{\tilde{x}_t}                     
\newcommand{\xr}{\tilde{x}_r}                     
\newcommand{\xz}{\tilde{x}_\theta}                
\newcommand{\xzr}{\tilde{x}_{\theta,r}}           
\newcommand{\xind}{\tilde{x}_{\mathrm{ind}}}      
\newcommand{\xaux}{\tilde{x}_{\mathrm{aux}}}      
\newcommand{\xu}{\tilde{x}_{\theta,\mathrm{u}}}    
\newcommand{\xv}{\tilde{x}_{\theta,\mathrm{v}}}    
\newcommand{\xoldu}{\tilde{x}_{\mathrm{old},\mathrm{u}}}
\newcommand{\xindr}{\tilde{x}_{\mathrm{ind},r}}   
\newcommand{\xrealr}{\tilde{x}_{\mathrm{real},r}} 
\newcommand{\Xmap}[3]{X\big[#1,\,#2,\,#3\big]}    
\newcommand{\Jterm}[3]{J\big[#1,\,#2,\,#3\big]}   
\newcommand{\vind}{v_{\mathrm{ind}}}              
\newcommand{\xreal}{\tilde{x}_{\mathrm{real}}}    
\newcommand{\uth}{u_\theta}
\newcommand{\uold}{u_{\mathrm{old}}}
\newcommand{\xreg}{\tilde{x}_{\mathrm{reg}}}       
\newcommand{\xbase}{\tilde{x}_{\mathrm{base}}}     
\newcommand{\xrwd}{\tilde{x}_{\mathrm{rwd}}}       
\newcommand{\gu}{g_{\mathrm{u}}}
\newcommand{\gv}{g_{\mathrm{v}}}
\newcommand{\Nel}{N}   
\newcommand{\xoldur}{{\xind|_{J=0}}}             
\newcommand{\norm}[1]{\left\lVert #1 \right\rVert}
\newcommand{\msp}[1]{\mean_{\mathrm{sp}}\!\left[ #1 \right]}
\newsavebox{\videosemanticbox}
{\catcode`\,=\active \gdef,{\mathchar"613B\penalty0\relax}}
\newcommand{\breakablecommas}{\mathcode`\,="8000\relax}

\begin{document}

\thispagestyle{firststyle}
\vspace*{0.25cm}
{\color{hunyuanblue}\hrule height 0.6pt}
\vskip 6mm
\begin{center}
{\LARGE\bfseries FewStepRL: Fine-tuning Few-step Diffusion Models with\\
Reinforcement Learning\par}
\end{center}
\vskip 3mm
{\color{hunyuanblue}\hrule height 0.6pt}
\vskip 6mm

\begin{center}
\textbf{Haitao Lin$^{1, \star, \dagger}$, Minsi Ren$^{2, \star}$, Yang Li$^{1, \dagger}$,\\ Xingtong Ge$^{1}$,  Zheyuan Hu$^{1}$, Peiyan Hu$^{2,3}$,  Tailin Wu$^{2,\ddagger}$}\\[2mm]
{\small {Affiliation: $^1$Tencent Hunyuan, $^2$ Westlake University, $^3$ University of Chinese Academy of Sciences}}
\\{\small $^{\star}$Equal contribution; $^{\dagger}$Project Leader;}
{\small $^{\ddagger}$Corresponding author: \texttt{wutailin@westlake.edu.cn}}
\end{center}
\vskip 6mm

\begin{tcolorbox}[
  colframe=abstractbg,
  colback=abstractbg,
  boxrule=0pt,
  arc=2mm,
  enhanced,
  top=12pt,
  bottom=12pt,
  left=15pt,
  right=15pt,
  width=\textwidth,
]
\textbf{Abstract.}\quad
Reinforcement learning (RL) has become a standard post-training method for
diffusion models, but most RL objectives roll out the current policy at every iteration, and
with a many-step sampler these rollouts typically dominate the cost of training. Few-step
models cut this cost, yet they model an average velocity over an interval,
whereas existing RL objectives are built for an instantaneous velocity with a
many-step sampler. We adapt RL objectives to few-step models systematically in
three phases, namely a transformation, a decomposition and an adaptation. The
transformation to $x$-space expresses each objective we consider through a common policy
prediction, at the clean end by default and at an explicitly marked level for
transition or flow-map objectives, so that objectives stated over densities,
velocities or rewards become comparable. The decomposition writes the gradient
of each of these objectives as the sum of a self-regression toward a reference
and a reward augmentation, so that a method reduces, at the gradient level, to a choice of anchors and
weights. The adaptation computes each anchor from a few-step model at the end
of its transport or along its tangent, with the instantaneous velocity
recovered from the network itself or from an auxiliary network. This yields
few-step versions of FlowGRPO, AWM, DiffusionNFT, DiffusionOPSD and
ReFL in one modular codebase. In our experiments, reward-gradient-based
methods are the most effective where the reward is differentiable and its
gradient affordable; otherwise DiffusionNFT-style methods generally reach
high rewards in relatively few iterations despite a sometimes higher per-iteration
cost.
Extra
inference steps show no systematic aggregate benefit in our experiments, although selected
outputs visibly improve with two to four times as many sampling steps as in
training.

\vskip 8pt
\textcolor{hunyuanblue}{\faGlobe}~\textbf{Project Page:} \href{https://edapinenut.github.io/fewsteprl.page/}{\texttt{fewsteprl.page}}\\[3pt]
\textcolor{hunyuanblue}{\faGithub}~\textbf{Code:} \href{https://edapinenut.github.io/fewsteprl.page/coming-soon.html}{\texttt{fewsteprl-sd35}} (images), \href{https://edapinenut.github.io/fewsteprl.page/coming-soon.html}{\texttt{fewsteprl-wan21}} (videos)\\[3pt]
\raisebox{-0.2ex}{\twemoji[height=1.05em]{1f917}}~\textbf{Checkpoints:} \href{https://edapinenut.github.io/fewsteprl.page/coming-soon.html}{Hugging Face (coming soon)}
\end{tcolorbox}
\clearpage

\etocsettagdepth{mainmatter}{subsubsection}
\etocsettagdepth{appendix}{none}
\tableofcontents
\etocdepthtag.toc{mainmatter}

\renewcommand{\arraystretch}{1.3}

\section{Introduction}\label{sec:intro}

Diffusion models and their flow-matching counterparts generate data by
integrating a learned transport from noise to samples, and have become a
dominant approach to image and video synthesis
\citep{ho2020ddpm,song2021scoresde,lipman2023flowmatching,liu2023rectifiedflow,rombach2022ldm,esser2024sd3}.
Because the learned field is accurate only locally, sampling typically takes tens of
network evaluations \citep{song2021ddim}.

Few-step models remove most of this cost by learning the transport over long
intervals directly. Consistency models and MeanFlow, which models the average
velocity over an interval, and on-policy distilled flow maps such as DMD and
AnyFlow generate in one to four evaluations
\citep{song2023consistency,geng2025meanflow,geng2026improvedmeanflow,yin2024dmd,gu2026anyflow}.

Reinforcement learning (RL) has become a standard post-training method for
diffusion models. Fine-tuning against reward models of human preference, text rendering or
composition measurably improves the outputs
\citep{black2024ddpo,fan2023dpok,clark2024draft,xu2023imagereward,liu2025flowgrpo,xue2026awm,zheng2026diffusionnft}.
This gain, however, typically requires samples from the current policy at every
iteration and must be protected against reward hacking, in which the policy
exploits the reward model instead of improving.

\begin{figure}[tbp]
\centering
\begin{subfigure}[t]{0.32\textwidth}
\centering
\includegraphics[width=\linewidth]{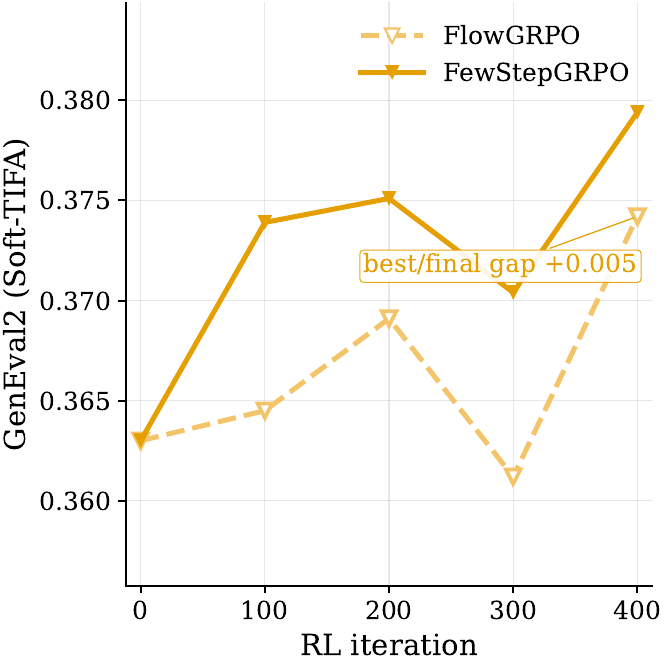}
\caption{FewStepGRPO (\S\ref{sec:fewstep-grpo}) vs.\ FlowGRPO}
\end{subfigure}\hfill
\begin{subfigure}[t]{0.32\textwidth}
\centering
\includegraphics[width=\linewidth]{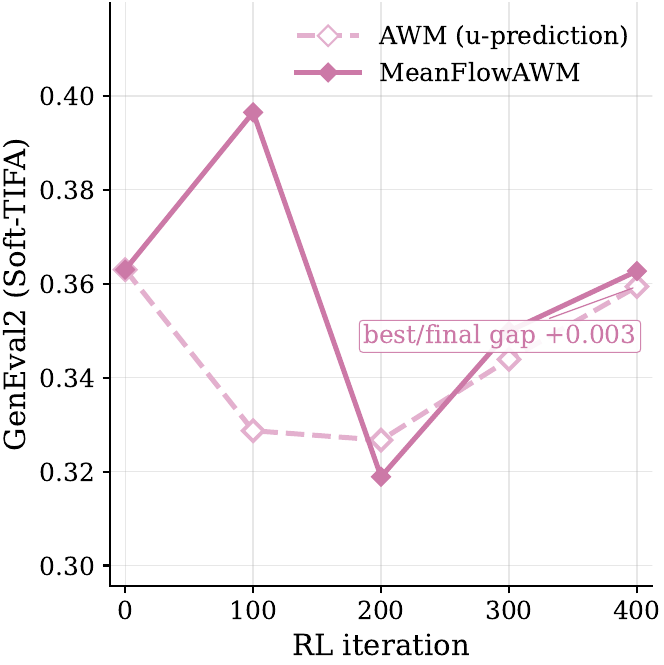}
\caption{MeanFlowAWM (\S\ref{sec:fewstep-awm}) vs.\ AWM}
\end{subfigure}\hfill
\begin{subfigure}[t]{0.32\textwidth}
\centering
\includegraphics[width=\linewidth]{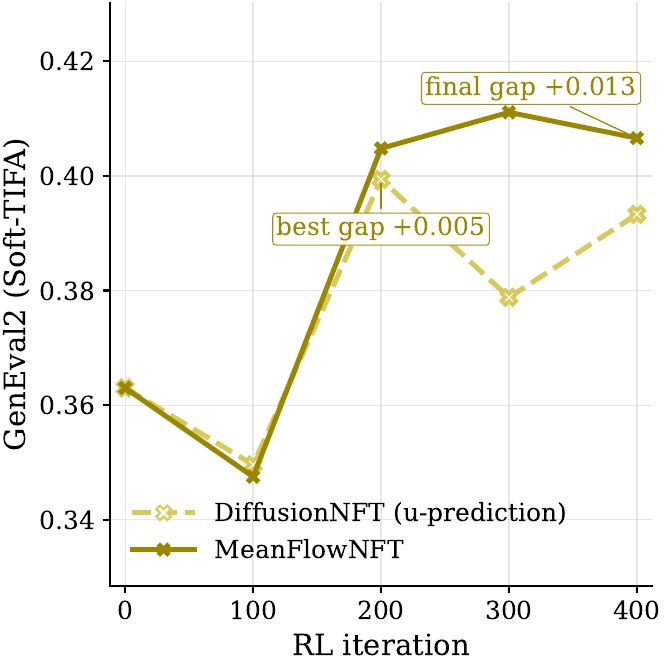}
\caption{MeanFlowNFT (\S\ref{sec:fewstep-nft}) vs.\ DiffusionNFT}
\end{subfigure}
\caption{Held-out GenEval2 against the RL iteration in the mixed-reward
setting of \S\ref{sec:exp-setup}, from the shared pre-RL few-step checkpoint.
Dashed: the original objective applied to the distilled model with the
average velocity substituted for the instantaneous velocity (FlowGRPO, AWM,
DiffusionNFT). Solid: the
corresponding few-step adaptation of this paper (FewStepGRPO, MeanFlowAWM,
MeanFlowNFT; \S\ref{sec:impl}). Each annotated gap is the adaptation minus
the baseline, at the best checkpoint of the baseline and at
iteration $400$; the two coincide when the baseline is best at iteration
$400$.}
\label{fig:motivation}
\end{figure}

With a many-step sampler, these rollouts typically
account for most of the
time and memory of training, and more so for long video sequences. A few-step
model shrinks this cost, but it models an average velocity over an interval,
whereas the RL objectives above are built for an instantaneous velocity or for
the transitions of a many-step sampler. Applied directly, they leave open which
network evaluation should be differentiated, how the reference policy should be
evaluated, and which parts of the interval are trained. Prior work typically resolves these
choices for one objective at a time
\citep{huang2026meanflownft,huang2026rtdmd,li2026flowmapgrpo}; what is, to our knowledge, missing
is a systematic way to adapt RL objectives to few-step models.
\Cref{fig:motivation} illustrates the effect of leaving them unresolved. Substituting
the average velocity of a distilled few-step model for the instantaneous
velocity required by FlowGRPO, AWM and DiffusionNFT, i.e., applying the
original objective to the distilled model without modification, gives a lower best GenEval2 than the
adapted objective derived in this paper in each of the three cases.

This paper builds one in three phases (\Cref{fig:overview}), namely a
transformation, a decomposition and an adaptation. The transformation maps each
objective to
$x$-space. Under a linear noise schedule, a velocity defines an affine map
between noise levels, so each objective we consider can be expressed through a
common policy prediction, at the clean end by default and at an explicitly
marked level for transition or flow-map objectives. In
$x$-space, objectives stated over densities, velocities or rewards become
directly comparable. The decomposition splits the loss gradient. 
The gradient of every objective we consider is the sum of a self-regression
term, which keeps the policy close to a reference, and a reward-augmentation
term, which contains the reward, so that, at the gradient level, a method reduces to a choice of
anchors and weights. The adaptation extends the anchors to few-step models. We
show
how each anchor is computed from a few-step model, either at the end of its
transport or along the tangent of the trajectory, with the instantaneous
velocity recovered from the network itself, as in MeanFlow, or from an
auxiliary network, as in DMD.

All methods share one codebase, which is represented by a single training step
with several method-specific stages, namely rollout, reward and advantage, anchors
and weights, and loss (plus auxiliary fitting and the refresh of the old
policy), where each method overrides only the hooks in which it differs
(\S\ref{sec:impl}). The methods are therefore compared under shared rollout,
reward and evaluation code, and a new variant is specified mainly by its
anchors and weights.
Experimentally, we compare seven implementations on image generation with an
AnyFlow-distilled SD3.5-Medium \citep{stabilityai2024sd35,gu2026anyflow}, under
a non-differentiable OCR reward and a differentiable mixed reward, and on video
generation with an AnyFlow-distilled Wan2.1-T2V-1.3B
\citep{wan2025wan,gu2026anyflow}; we further test whether the fine-tuned
few-step models improve when sampled with more steps. 

The contributions of this paper are as follows.
\begin{itemize}
\item \textbf{A general view.} After the transformation to
$x$-space, the gradients of the RL objectives considered here share one
decoupled form, the sum of a self-regression term and a reward-augmentation
term, so that their gradients differ only in their anchors and weights.
\item \textbf{A systematic adaptation to few-step models.} Each of these anchors can be
computed from an average-velocity model at the end of the transport or along its
tangent, which yields few-step versions of FlowGRPO, AWM, DiffusionNFT,
DiffusionOPSD and ReFL within one modular codebase.
\item \textbf{Empirical guidance with experimental analysis.} Where the reward is differentiable and its
gradient affordable, reward-gradient-based methods are the most effective in our experiments, and ReFL
alone is a strong baseline. Otherwise, group-advantage methods apply. Among
them, DiffusionNFT-style methods generally reach high rewards in relatively few
iterations, even when one iteration is expensive, and DMD-based variants are
faster per iteration on image but slower on video, depending on the relative
cost of forward and backward passes, and often coincide with larger
held-out degradation. Additional inference steps bring little aggregate
improvement in generation quality, while visual gains appear in selected outputs with
more sampling steps than training.
\end{itemize}

\section{Background}\label{sec:background}

\paragraph{Diffusion and flow-based generators.}
Diffusion models learn to reverse a fixed noising process
\citep{ho2020ddpm}; score-based models cast this reversal as a reverse-time
SDE whose probability-flow ODE gives a deterministic sampler, of which DDIM is
a discrete instance \citep{song2021scoresde,song2021ddim}. Flow matching and
rectified flow instead regress the velocity of a prescribed interpolation
between noise and data, the latter straightening the paths so that fewer
integration steps suffice \citep{lipman2023flowmatching,liu2023rectifiedflow}.
These models were scaled by moving to a latent space \citep{rombach2022ldm}, by
transformer backbones \citep{peebles2023dit,ma2024sit} and by larger, more
efficient architectures \citep{podell2024sdxl,esser2024sd3,xie2025sana15},
with classifier-free guidance as a standard way to trade diversity for
prompt fidelity \citep{ho2021classifierfree}. Related generators include
current text-to-image systems
\citep{openai2023dalle2,betker2023dalle3,saharia2022imagen,blackforestlabs2024flux,blackforestlabs2025flux2,seedream2025seedream4,imageteam2025zimage,qin2025lumina2},
multimodal models that both generate and understand images in one network
\citep{chen2025januspro,ma2025janusflow,xie2025showo,wang2024emu3,hurst2024gpt4osystemcard,chen2025sharegpt4oimage},
trained on web-scale image--text data
\citep{schuhmann2022laion5b,huggingface2024texttoimage2m}, and video
generators such as Wan and LongCat-Video
\citep{wan2025wan,meituan2025longcatvideo}. These models typically learn the
instantaneous velocity of the sampling flow, or an affine transformation of it
such as the score or the predicted noise. Since this velocity is instantaneous,
the sampler integrates it with many small steps, typically 32 or 64, which
makes inference costly.

\paragraph{Few-step generation.}
Sampling cost has been cut along several lines. Progressive distillation halves
the number of steps repeatedly by training a student to match two teacher
steps with one \citep{salimans2022progressive}. Consistency models map every
point of a trajectory to its endpoint, and were extended to latent space,
arbitrary trajectory jumps, multiple sub-intervals and segmented trajectories
\citep{song2023consistency,luo2023lcm,kim2024ctm,wang2024phasedconsistency,ren2024hypersd};
consistency has also been used as a training regulariser against sampling
drift and to learn from noisy data
\citep{daras2023consistentdiffusion,daras2024ambientdiffusion}. Shortcut
models condition on the step size and learn to take large steps directly
\citep{frans2025shortcut}. Adversarial distillation adds a discriminator so
that one to four steps produce sharp samples
\citep{wang2023diffusiongan,sauer2024add,sauer2024ladd,lin2024sdxllightning,chadebec2025flashdiffusion, lu2025scm}.
Distribution matching distillation (DMD) trains the generator so that its
output distribution matches the teacher's, using the difference between a
teacher score and the score of an auxiliary network fitted to the generator's
samples \citep{yin2024dmd,yin2024dmd2}, an idea shared with Diff-Instruct and
variational score distillation \citep{luo2023diffinstruct,wang2023prolificdreamer};
later work scales it to flow models, splits it into
sub-intervals, decouples guidance from matching, moves it to continuous time,
regularises it with consistency, and adapts it to video
\citep{ge2025senseflow,fan2026phaseddmd,liu2026decoupleddmd,liu2026cdm,zheng2026rcm,ge2026salt,nie2026tmd,luo2025tdm}.
Flow maps learn the transport between any two noise levels. MeanFlow
parameterises the average velocity over an interval and trains it through its
identity with the instantaneous velocity
\citep{geng2025meanflow,lin2025esc, geng2026improvedmeanflow}, and AnyFlow distils such a
map on policy for any step count \citep{gu2026anyflow}. Unlike the many-step
generators above, these flow-map models learn an average velocity over an
interval, which can be read as the direction of a secant of the sampling
trajectory, differing substantially from the instantaneous velocity where the
trajectory is curved.

\paragraph{Reinforcement learning for diffusion models.}
RL fine-tuning of generators builds on policy optimisation
\citep{levine2018probabilisticrl,schulman2017ppo}, preference optimisation
\citep{rafailov2023dpo} and group-relative advantages, which normalise rewards
within a group of samples for the same prompt
\citep{shao2024deepseekmath,guo2025deepseekr1}. For diffusion models, early
work fine-tunes on human feedback \citep{lee2023aligning}, treats denoising as
a Markov decision process optimised by policy gradients
\citep{black2024ddpo,fan2023dpok}, or optimises pairwise preferences directly
\citep{wallace2024diffusiondpo}. FlowGRPO brings group-relative policy
optimisation to flow models by turning the ODE sampler into an SDE with
tractable transition likelihoods \citep{liu2025flowgrpo}; follow-ups extend it
to video, mix ODE and SDE steps, weight timesteps, and preserve the noise
coefficients of the sampler
\citep{xue2025dancegrpo,li2026mixgrpo,he2026tempflowgrpo,wang2025cps}. A
second family avoids transition likelihoods by acting on the forward process:
AWM weights the pretraining loss by the advantage \citep{xue2026awm},
reward-weighted and policy-gradient flow matching reweight or rescale the
velocity regression \citep{fan2025rewardweightedflow,mcallister2026flowmatchingpolicygradients},
adjoint matching is scaled to RL post-training \citep{bergmeister2026ram}, and
DiffusionNFT contrasts positive and negative implicit policies built from the
model and the frozen policy that generated its rollouts
\citep{zheng2026diffusionnft}, following NFT in
language models \citep{chen2026nft}. A third family backpropagates
differentiable rewards through the sampler, as in DRaFT and ReFL
\citep{clark2024draft,xu2023imagereward,wang2026rewardbackprop}; the recently
proposed DiffusionOPSD \citep{zhou2026diffusionopsd} belongs to this family
but constructs detached points with normalised reward-gradient steps and fits
the policy to them, thereby forming a surrogate gradient. These methods are
built for an instantaneous velocity, and most of them need on-policy
rollouts at every iteration, which are expensive with the many-step samplers discussed above; how
to apply RL to few-step models is therefore an important question.

\paragraph{RL for few-step generators.}
Closest to this paper are methods that fine-tune few-step models with rewards.
Several combine DMD with RL, by adding a reward term to distillation, guiding
distillation with reward gradients, or recasting distribution matching itself
as a reward \citep{chen2026flashdmd,jiang2026dmdr,dong2026guidingdmd,fan2026rdm};
others tune distilled models with non-differentiable rewards,
Kullback--Leibler (KL) regularised
alignment or pairwise sample optimisation
\citep{luo2026tdmr1,luo2025diffinstructpp,miao2025pso}. For flow maps,
Flow-Map GRPO composes stochastic steps of a flow-map generator
\citep{li2026flowmapgrpo}, MeanFlowNFT applies DiffusionNFT to an
average-velocity model through its self-induced instantaneous field
\citep{huang2026meanflownft}, and RTDMD optimises a reward-tilted distribution
matching objective \citep{huang2026rtdmd}. These works mostly adapt one objective
to one kind of few-step model; we instead give a common construction that
extends FlowGRPO, AWM, DiffusionNFT, DiffusionOPSD and ReFL to
average-velocity models, and compare them under one implementation.

\section{Preliminaries}\label{sec:prelim}


\subsection{Few-step Diffusion Models with On-Policy Distillation}

We write $t\in[0,1]$ for the noise level, with data at $t=0$ and Gaussian
noise at $t=1$. A plain $x_t$ denotes a point at level $t$ on a sampling
trajectory, and $\xt$ denotes a renoised state at level $t$ obtained by adding
fresh noise to a clean sample (\eqref{eq:xt-source}); every map below applies
to either. The setting we focus on throughout is
\emph{distillation}, where a pretrained
instantaneous-velocity field $v_{\mathrm{real}}(\xt,t)$ is available and frozen,
which we call the teacher,
and the student $\uth(\xt,t,r)$ must reproduce its transport in one or a few
network calls. The teacher is an ordinary flow-matching model, trained by
regressing onto the conditional velocity $\vcond=\epsilon-x$ of a clean sample
$x$ and Gaussian noise $\epsilon\sim\mathcal N(0,I)$, and it provides
the student's regression target in \eqref{eq:mfloss}
\citep{lipman2023flowmatching,liu2023rectifiedflow}. The student is the
average-velocity generator 
\citep{geng2025meanflow,geng2026improvedmeanflow,gu2026anyflow},
\begin{equation}
    \uth(x_t,t,r)\;=\;\frac{1}{t-r}\int_{r}^{t} v_\theta(x_\nu,\nu)\,
    \mathrm d\nu ,
    \label{eq:avgvel}
\end{equation}
where $v_\theta$ is the instantaneous velocity induced by the student, $x_\nu$
is the state at level $\nu$ on the probability-flow trajectory of $v_\theta$
through $x_t$, and $\uth(x_t,t,t):=v_\theta(x_t,t)$ closes the
definition at $r=t$. Throughout, we identify this induced field with the
flow-matching marginal velocity of the student's sample distribution
$p_\theta$, $v_\theta(x_t,t)=\mathbb E[\epsilon-\hat x\mid x_t]$ with
$\hat x\sim p_\theta$ (\Cref{app:notation}); the two coincide when the flow
map is exact.

Unlike off-policy methods, which query the model at
$\xt=\AddNoise(x_0,t)$ with $x_0\sim p_{\mathrm{data}}$, \emph{on-policy}
training draws the noised state from the student's own distribution:
\begin{equation}
\xt=\AddNoise\big(\xhth,t\big)=(1-t)\,\xhth+t\,\epsilon,
\qquad
\xhth\sim p_\theta(x),\quad \epsilon\sim\mathcal N(0,I) .
\label{eq:xt-source}
\end{equation}
The subscript $\theta$ in \eqref{eq:xt-source} is kept only where the
sampler is differentiated.
The samples $\xhth$ are produced by a rollout operator
$\Rollout$, an ODE or SDE solver
that maps Gaussian noise at level $1$ to a clean sample at level $0$ for a
text prompt $c$ in $M_{\mathrm{step}}$ steps, on a grid
$1=t_0>t_1>\dots>t_{M_{\mathrm{step}}}=0$:
\begin{equation}
\xhth=\Rollout\big(\epsilon,\,\uth,\,c\big),
\qquad
\hat x_{\theta,t_0}=\epsilon\sim\mathcal N(0,I).
\label{eq:rollout-def}
\end{equation}
For an instantaneous field, one Euler step of the solver between $t_j$ and
$t_{j+1}$ is
\begin{equation}
\hat x_{\theta,t_{j+1}}
=\hat x_{\theta,t_j}-(t_j-t_{j+1})\,v_\theta\big(\hat x_{\theta,t_j},t_j\big),
\label{eq:rollout-euler}
\end{equation}
which is only first-order accurate, so $M_{\mathrm{step}}$ must be large. A
displacement is the time integral of the velocity, and by \eqref{eq:avgvel}
$(t-r)\,\uth$ equals this integral over $[r,t]$. With $\uth$ as the
generator, the Euler step is replaced by an exact jump,
\begin{equation}
\hat x_{\theta,r}=\hat x_{\theta,t}-(t-r)\,\uth\big(\hat x_{\theta,t},t,r\big),
\label{eq:rollout-op}
\end{equation}
so a few-step sampler is a chain of $M_{\mathrm{step}}$ such jumps, each a
single subtraction rather than an integration. Taking the whole interval at
once, $M_{\mathrm{step}}=1$, $t=1$ and $r=0$, recovers the one-call generator.

\paragraph{MeanFlow.} According to the definition of $\uth$ in
\eqref{eq:avgvel} \citep{geng2025meanflow}, multiplying by $(t-r)$ and
differentiating in $t$ eliminates the integral and
leaves the \emph{MeanFlow identity}
\begin{equation}
\uth(\xt,t,r)\;=\;v_\theta(\xt,t)\;-\;(t-r)\,\partial_t\uth(\xt,t,r),
\label{eq:mfid}
\end{equation}
where $\partial_t\uth=\mathrm d\uth/\mathrm dt$ is the total derivative along
the trajectory $x_\nu$ of \eqref{eq:avgvel}; the implementation estimates it
by a directional-difference estimate (\eqref{eq:cd}). The identity follows
from the definition \eqref{eq:avgvel} and imposes no additional constraint; it
expresses the average velocity as the instantaneous velocity minus $(t-r)$
times the total derivative of the average velocity with respect to $t$.

Every use of \eqref{eq:mfid} treats this correction as detached, so we define
it once with the stop-gradient $\sg(\cdot)$, which is the identity in the
forward pass and zero in the backward pass,
\begin{equation}
\Jterm{u}{t}{r}\;:=\;(t-r)\,\sg\big(\partial_t u\big),
\label{eq:jvp}
\end{equation}
the \emph{detached Jacobian-vector term}. It gives the gap between an average velocity
over $[r,t]$ and the instantaneous one at $t$, measured on the network itself
and frozen before it is used. Substituting the teacher for $v_\theta$ in
\eqref{eq:mfid} then turns the identity into a regression whose target is
assembled entirely from quantities available at the current state,
\begin{equation}
\mathcal L_{\mathrm{MF}}
=\mathbb E_{\,\xt,\,t,\,r}\Big[\norm{\uth(\xt,t,r)
-\big(v_{\mathrm{real}}(\xt,t)-\Jterm{\uth}{t}{r}\big)}^{2}\Big],
\label{eq:mfloss}
\end{equation}
in which $v_{\mathrm{real}}$ is frozen and $J$ contains a stop-gradient,
so $\uth$ is the only differentiated quantity.
Setting $r=t$ recovers instantaneous-velocity distillation onto $v_{\mathrm{real}}$;
the case $r<t$ trains the flow map and hence the one-step transport
$\xzr=\xt-(t-r)\,\uth(\xt,t,r)$ to level $r$.

\paragraph{DMD.} Distribution matching distillation
\citep{yin2024dmd,yin2024dmd2} does not regress $\uth$ onto a target. It
maintains a second, \emph{auxiliary} velocity $\vph(\xt,t)$ with its own
parameters $\phi$, fitted online by flow matching to the samples generated by
the student. Its target is the conditional velocity of each sample instead of
the teacher, because $\vph$ is meant to approximate the velocity of $p_\theta$
instead of $p_{\mathrm{data}}$:
\begin{equation}
\mathcal L_{\mathrm{aux}}
=\mathbb E_{\,\xh\sim p_\theta,\ \epsilon,\ t}
\Big[\norm{\vph(\xt,t)-\vcond}^{2}\Big],
\qquad \xt=\AddNoise(\xh,t),\quad \vcond=\epsilon-\xh .
\label{eq:auxloss}
\end{equation}
The difference between the two fields then gives the score difference in the
gradient of the reverse KL divergence, which is used to
train $\uth(\xt,t,r)$:
\begin{equation}
\nabla_\theta\mathcal L_{\mathrm{DMD}}
=\mathbb E_{\,\epsilon,\,t\in(0,1)}\Big[
\frac{1-t}{t}\,\sg\big[v_{\mathrm{real}}(\xt,t)-\vph(\xt,t)\big]^{\!\top}
\frac{\partial\,\xhth}{\partial\theta}\Big],
\label{eq:dmdloss}
\end{equation}
The order of the two fields is fixed by the conversion between them. The
score of the linear path is $\nabla\log p_t(\xt)=-\mathbb E[\epsilon\mid\xt]/t$ with
$\mathbb E[\epsilon\mid\xt]=\xt+(1-t)v$, so the difference
$\nabla\log p^{\phi}_t-\nabla\log p^{\mathrm{real}}_t$ of the scores of the two
fields, which the reverse KL produces, equals
$\frac{1-t}{t}\,(v_{\mathrm{real}}-\vph)$. Converting scores to velocities reverses the
order of the two fields and introduces the score-to-velocity factor $\frac{1-t}{t}$, which is positive on
$(0,1)$ and is absorbed into a per-$t$ weight in practice. Descending
\eqref{eq:dmdloss} therefore applies the
direction $+(\vph-v_{\mathrm{real}})$, reducing the mismatch to the data field
relative to the auxiliary field that tracks $p_\theta$.

The term $\partial\xhth/\partial\theta$ requires a differentiable sampler, and
the sampler used in practice is stochastic. A step at level $t$ does not move
directly to the next level; it first maps the state to $r=0$ with one network
evaluation and then renoises the result to an intermediate level $t'$ with
$0<t'<t$:
\begin{equation}
\hat x_{\theta,0}=\hat x_{\theta,t}-t\,\uth\big(\hat x_{\theta,t},t,0\big),
\qquad
\hat x_{\theta,t'}=(1-t')\,\hat x_{\theta,0}+t'\,\epsilon',
\qquad \epsilon'\sim\mathcal N(0,I) .
\label{eq:sde-step}
\end{equation}
Unlike the deterministic chain \eqref{eq:rollout-op}, this sampler draws fresh
noise $\epsilon'$ at every step, so the state at level $t'$ is not a
deterministic function of the previous state.

The levels follow a fixed grid of $M_{\mathrm{step}}+1$ values spanning
$M_{\mathrm{step}}$ intervals; a step count $\ell\in\{1,\dots,M_{\mathrm{step}}\}$ is drawn
per update, and only the \emph{last} step remains in the computation graph,
with all earlier steps detached. The backward pass therefore requires one
network evaluation for any $\ell$. Moreover, the differentiated evaluation is
$\uth(\cdot,t,0)$ at the level that the grid assigns to step $\ell$, so
randomising $\ell$ changes this level across updates. Over training, the
gradient therefore reaches $\uth(\cdot,t,0)$ at every grid level $t$ instead
of a single one.

\subsection{Reinforcement Learning for Diffusion Models}\label{sec:rl}

\paragraph{FlowGRPO.}\label{sec:grpo}
FlowGRPO combines a GRPO-style group baseline with a PPO transition objective
\citep{schulman2017ppo,shao2024deepseekmath,liu2025flowgrpo}. It scores a
\emph{group} of $K$ rollouts that share a
prompt $c$, and turns the raw scores into a group-normalised advantage. Writing
$R^{(i)}=R\big(\dec(\xh^{(i)}),c\big)$ for the endpoint reward of member $i$
($\dec$ the decoder), the advantage standardises the reward within the group,
\begin{equation}
A^{(i)}
=\frac{R^{(i)}-\operatorname{mean}\big(\{R\}\big)}
{\operatorname{std}\big(\{R\}\big)+\varepsilon_{\mathrm{std}}},
\qquad i=1,\dots,K .
\label{eq:adv-prelim}
\end{equation}
Here $\{R\}$ denotes the $K$ rewards of the group and $\varepsilon_{\mathrm{std}}>0$ is a small
floor. The advantage is a detached per-sample scalar with zero mean across the
group; no value
network is fitted, and the group mean serves as the baseline. AWM below uses
the same $A^{(i)}$; DiffusionNFT uses a bounded transform of it.

FlowGRPO formulates the stochastic differential equation (SDE) sampler as a
Markov decision process, whose state is the current noisy sample $x_t$ (a
point on the sampling trajectory, in the convention of
\S\ref{sec:prelim}) together
with the prompt and the level and whose action is the next state produced by
the sampler; the Gaussian transition $p^{\mathrm{tr}}_\theta(x_r\mid x_t)$ then
provides a likelihood, which the deterministic sampler lacks. Optimisation is
proximal policy optimisation on the realised action
$x_r$,
\begin{equation}
\mathcal L_{\mathrm{GRPO}}
=-\,\mathbb E_{\,i,\,t}\Big[\min\big\{A^{(i)}\rho_{\mathrm{tr}}^{(i)},\;
A^{(i)}\clip\big(\rho_{\mathrm{tr}}^{(i)},1\pm\varepsilon_{\mathrm{clip}}\big)\big\}\Big],
\qquad
\rho_{\mathrm{tr}}
=\frac{p^{\mathrm{tr}}_\theta(x_r\mid x_t)}
{p^{\mathrm{tr}}_{\mathrm{old}}(x_r\mid x_t)} .
\label{eq:grpoloss}
\end{equation}
Here the \emph{old policy} is the frozen policy, with parameters
$\theta_{\mathrm{old}}$, that generated the rollouts on which the current policy
is trained; it receives no gradient. In our implementation,
$\theta_{\mathrm{old}}$ is an exponential moving average (EMA) of $\theta$,
refreshed after every outer iteration (\S\ref{sec:impl-base}) with the
EMA-decay ratio as the weight kept on the previous $\theta_{\mathrm{old}}$; an
EMA-decay ratio of $0.0$ therefore copies $\theta$, so that
$\theta_{\mathrm{old}}=\theta$ at the start of every outer iteration. The
subscript $\mathrm{old}$ marks its quantities, so
$p^{\mathrm{tr}}_{\mathrm{old}}$ is its transition density, $v_{\mathrm{old}}$
its instantaneous velocity and $p_{\mathrm{old}}$ the distribution of its
rollouts. The clipping radius is $\varepsilon_{\mathrm{clip}}$. For later
clipped surrogates, define the detached
\emph{unclipped-gradient gate}
\begin{equation}
\chi(z,A)
:=\mathbf 1\!\left[
  (A>0\ \wedge\ z\le1+\varepsilon_{\mathrm{clip}})
  \ \vee\
  (A<0\ \wedge\ z\ge1-\varepsilon_{\mathrm{clip}})
\right],
\qquad
\chi(z,0):=0 .
\label{eq:clip-gate}
\end{equation}
At the two clipping kinks this convention selects the derivative of the
unclipped branch.
The sign of the reward advantage determines the direction of the update of the
mean $\mu_\theta$ of the Gaussian transition (the scoring kernel defined in
\eqref{eq:grpo-v-mean}) relative to the visited noisy state $x_r$.

\paragraph{AWM.} Advantage Weighted Matching \citep{xue2026awm} keeps the
signed advantage but does not evaluate a likelihood on the reverse process.
Each update draws a fresh $(t,\epsilon)$ and constructs both the noised state
and the regression target from the rollout sample,
\begin{equation}
\xt^{(i)}=\AddNoise\big(\xh^{(i)},t\big)=(1-t)\,\xh^{(i)}+t\,\epsilon,
\qquad
\vcond^{(i)}=\epsilon-\xh^{(i)} ,
\label{eq:cond-vel}
\end{equation}
where $\vcond^{(i)}$ is the \emph{conditional velocity}, i.e., the velocity of
the straight path from $\xh^{(i)}$ to $\epsilon$, which is computed from the two
endpoints without a network evaluation. AWM then observes that the
flow-matching loss against that target is, for a suitable time weight $w(t)$, an evidence-lower-bound
surrogate for the sequence-level log-likelihood of the clean sample,
\begin{equation}
\log p_\theta\big(\xh^{(i)}\mid c\big)
\;\ge\;
\log\hat p_\theta\big(\xh^{(i)}\mid c\big)+\mathrm{const},
\qquad
\log\hat p_\theta\big(\xh^{(i)}\mid c\big)
:=-\,\mathbb E_{\,t,\,\epsilon}\Big[w(t)\,
\norm{v_\theta\big(\xt^{(i)},t\big)-\vcond^{(i)}}^{2}\Big],
\label{eq:awm-elbo}
\end{equation}
so that the policy-gradient term $A^{(i)}\nabla_\theta\log\hat p_\theta$
reduces to the pretraining objective reweighted per sample:
\begin{equation}
\mathcal L_{\mathrm{AWM}}
=\mathbb E_{\,i,\,t,\,\epsilon}
\Big[A^{(i)}\,w(t)\,
\norm{v_\theta\big(\xt^{(i)},t\big)-\vcond^{(i)}}^{2}\Big],
\label{eq:awmloss}
\end{equation}
with the time weight $w(t)=1$ in the original AWM (our implementation
effectively uses $w(t)=\tau_{\mathrm{AWM}}^{2}$, \S\ref{sec:fewstep-awm}) and
a velocity-space proximal penalty
$\norm{v_\theta-v_{\mathrm{old}}}^{2}$ against the old policy
$v_{\mathrm{old}}$, called a KL term in AWM, added for stability. Minimising
\eqref{eq:awmloss} decreases the matching residual at $\xh^{(i)}$ when
$A^{(i)}>0$ and increases it when $A^{(i)}<0$. Since no reverse-time
discretisation appears, the sampler can be chosen freely.

\paragraph{DiffusionNFT.} Negative-aware fine-tuning
\citep{zheng2026diffusionnft} builds on an optimality-variable construction
\citep{levine2018probabilisticrl} and requires neither a likelihood nor a signed
weight. It first maps the advantage to a bounded \emph{optimality probability}
with a normaliser $Z_c=\operatorname{std}(\{R\})+\varepsilon_{\mathrm{std}}$,
\begin{equation}
\hat r^{(i)}
=\frac12+\frac12\,\clip\!\left[
\frac{R^{(i)}-\operatorname{mean}\big(\{R\}\big)}{Z_c},\;-1,\;1\right]
=\frac12+\frac12\,\clip\big(A^{(i)},\,-1,\,1\big)\;\in[0,1],
\label{eq:rhat-prelim}
\end{equation}
since the bracketed quotient then equals \eqref{eq:adv-prelim}. Thus $A^{(i)}$ is signed,
unclipped and zero-mean under \eqref{eq:adv-prelim} and enters as a
\emph{multiplier}, whereas $\hat r^{(i)}$ is bounded, monotone in $A^{(i)}$
and equal to $\tfrac12$ when $A^{(i)}=0$, and enters as a
\emph{mixing weight}. A positive advantage gives $\hat r^{(i)}>\tfrac12$ and
hence a larger weight on the positive branch.

Each update then draws a fresh $(t,\epsilon)$ and forms the noised state
$\xt^{(i)}$ and the conditional velocity $\vcond^{(i)}$ as in
\eqref{eq:cond-vel}; the target is the same as in AWM, and, besides the weight, only the regressed
quantity differs. Two {implicit} policies are built from one
trainable network $v_\theta$ and the frozen old policy
$v_{\mathrm{old}}$, both evaluated at that same $\xt^{(i)}$ and mixed by a
branch coefficient $\beta$:
\begin{equation}
v^{\pm}_\theta\big(\xt^{(i)},t\big)
:=(1\mp\beta)\,v_{\mathrm{old}}\big(\xt^{(i)},t\big)
\pm\beta\,v_\theta\big(\xt^{(i)},t\big).
\label{eq:nft-implicit}
\end{equation}
Both are regressed onto the same target $\vcond^{(i)}$, mixed by $\hat r^{(i)}$:
\begin{equation}
\mathcal L_{\mathrm{NFT}}
=\mathbb E_{\,i,\,t,\,\epsilon}
\Big[\hat r^{(i)}\,\norm{v^{+}_\theta-\vcond^{(i)}}^{2}
+\big(1-\hat r^{(i)}\big)\,\norm{v^{-}_\theta-\vcond^{(i)}}^{2}\Big].
\label{eq:nftloss}
\end{equation}
Both branches are affine in the same trainable network $v_\theta$; since
$v_{\mathrm{old}}$, $\vcond^{(i)}$ and $\hat r^{(i)}$ are detached, $v_\theta$
is the only differentiated quantity in \eqref{eq:nftloss}.

\section{Framework}\label{sec:frame}
\begin{figure}[t]
\centering
\includegraphics[width=\textwidth]{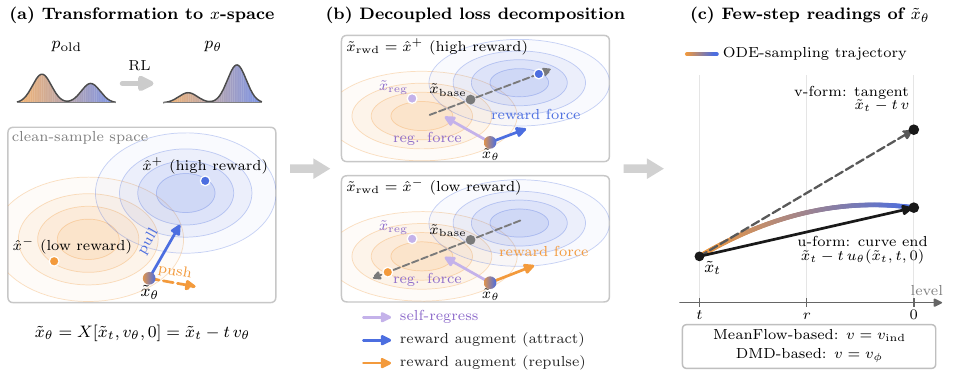}
\caption{Overview of the framework. (a)~RL increases the relative mass of the
high-reward mode under the rollout distribution $p_{\mathrm{old}}$ of the
old policy. The default
objective is evaluated at the clean end, on the policy prediction
(renoise-and-denoise sample)
$\xz=\Xmap{\xt}{v_\theta}{0}=\xt-t\,v_\theta$,
whose reward augmentation attracts it toward high-reward rollout samples and
repels it from low-reward ones; in the figure, $\hat x^{+}$ and $\hat x^{-}$
denote a high-reward and a low-reward rollout sample $\xh^{(i)}$,
respectively. (b)~The RL-loss gradient can be decoupled into a
self-regression term, visualised as a regression force toward the regression
anchor $\xreg$, and a reward-augmentation term, visualised as a reward force
along the displacement $\xrwd-\xbase$ from the baseline anchor $\xbase$, which
lies between the two modes, to the rewarded anchor $\xrwd$. A high-reward
sample produces attraction along that displacement (blue); a low-reward
sample produces repulsion against it (orange), so both updates favour higher reward
(\S\ref{sec:grad}). (c)~For a few-step model, the prediction and the anchors
$\{\tilde x_{\mathrm{rwd}},\tilde x_{\mathrm{base}}, \tilde x_{\mathrm{reg}}\}$
are computed either at the end
of the transport curve, $\xt-t\,u$ (u-form), or along the tangent
at $\xt$, $\xt-t\,v$ (v-form). The instantaneous velocity is recovered from
the network itself for
MeanFlow, $v=u+(t-r)\,\sg(\partial_t u)$ with the detached correction
evaluated on the old policy in the implemented methods (\eqref{eq:vpred-nft}),
and from an auxiliary network for DMD, $v=\vph$, fitted so that
$\vph(\xt,t)\approx\mathbb E\big[\epsilon-\xh\mid\xt\big]$
(\S\ref{sec:fewstep}).}
\label{fig:overview}
\end{figure}

Each RL method of \S\ref{sec:rl} increases the relative probability of
high-reward samples under the rollout distribution $p_{\mathrm{old}}$ of the
old policy. The methods
differ in the space in which the update is formulated (transition densities,
velocities, or rewards differentiated through the sample) and in how a
few-step network computes the quantities that the update requires. This
section resolves both differences in three steps, summarised in
\Cref{fig:overview}.
\begin{enumerate}
\item \emph{Transformation to $x$-space} (\S\ref{sec:xspace},
\Cref{fig:overview}a). Under the linear noise schedule a velocity transports a state
affinely, so each objective of \S\ref{sec:prelim} can be expressed through a
common prediction in $x$-space. By default this prediction is the policy
prediction
$\xz$; MeanFlow and FewStepGRPO keep an explicitly marked level-$r$ quantity.
In $x$-space the methods become directly comparable.
\item \emph{Decomposition of the gradient} (\S\ref{sec:grad}, \Cref{fig:overview}b). In
$x$-space, the gradient of each of these losses is the transposed Jacobian
$(\partial\xz/\partial\theta)^{\top}$ applied to the sum of a reward-free
self-regression toward an anchor $\xreg$, which limits how far the prediction
moves, and a reward augmentation along $\xrwd-\xbase$, which contains the
reward. A method is then specified by the prediction, three anchors
$(\xreg,\xrwd,\xbase)$ and three weights $(W_{\mathrm{reg}},W_{\mathrm{rwd}},\omega)$
instead of a new loss.
\item \emph{Adaptation to few-step models} (\S\ref{sec:fewstep},
\Cref{fig:overview}c). The framework requires instantaneous velocities,
whereas a few-step model outputs an average velocity. Its anchors are
therefore computed either at the end of the transport curve (u-form) or along
the tangent at the current state (v-form). The latter recovers the
instantaneous velocity from the network itself, as in MeanFlow, or from an
auxiliary network, as in DMD.
\end{enumerate}
\S\ref{sec:impl} then specifies the predictions, anchors and weights of each
family.

\subsection{Transformation to x-space}\label{sec:xspace}

The objectives of \S\ref{sec:prelim} are written in transition-density or
velocity space. Under the linear noise schedule, a velocity instead identifies a
point in $x$-space at any chosen level $r$. This $x$-space includes noisy
level-$r$ points; clean samples correspond to $r=0$.

For any point $x_t$ at level $t$ (in particular a renoised state $\xt$),
define the affine transport
\begin{equation}
\Xmap{x_t}{v}{r}\;:=\;x_t-(t-r)\,v .
\label{eq:xmap}
\end{equation}
For two velocities evaluated at the same $x_t$ and transported to the same
level $r<t$,
\begin{equation}
\norm{v_a-v_b}^{2}
\;=\;\frac{1}{(t-r)^{2}}\,
\norm{\Xmap{x_t}{v_a}{r}-\Xmap{x_t}{v_b}{r}}^{2}.
\label{eq:xidentity}
\end{equation}
Thus a velocity MSE equals a scaled MSE between transported points in
$x$-space. At the clean level $r=0$ the factor is $t^{-2}$, and we omit the
level subscript.
In particular, for $\xt=(1-t)x+t\epsilon$ and
$\vcond=\epsilon-x$, \eqref{eq:xmap} gives
$\Xmap{\xt}{\vcond}{0}=x$.

For the reward objectives below, a detached rollout endpoint $\xh^{(i)}$ is
re-noised to $\xt^{(i)}$, and the policy provides the current policy prediction
\begin{equation}
\xz^{(i)}\;=\;\Xmap{\xt^{(i)}}{v_\theta}{0}\;=\;\xt^{(i)}-t\,v_\theta ,
\label{eq:xz-inst}
\end{equation}
while the conditional target maps back to $\xh^{(i)}$. Hence the loss compares
predictions at the same state $\xt^{(i)}$, and only $\xz^{(i)}$ is
differentiated. \S\ref{sec:fewstep} constructs this prediction when the network
outputs an average velocity instead of an instantaneous one.
Analysing a loss in $x$-space therefore shows how the prediction $\xz$ must
change to reduce the loss, i.e., which points $\xz$ moves toward and which it
moves away from. The same map applied to the old policy gives the
\emph{old policy's prediction}
\begin{equation}
\tilde x^{(i)}_{\mathrm{old}}=\Xmap{\xt^{(i)}}{v_{\mathrm{old}}}{0},
\label{eq:xold-def}
\end{equation}
which serves as an anchor in several methods below; its few-step variants,
the self-induced prediction of \refex{ex:ra-mfdmd} and the u-prediction and
$J$-free query of \S\ref{sec:fewstep}, are collected in \Cref{app:notation}.
\begin{example}[AWM in x-space]
    The target of \eqref{eq:awmloss} is
    $\vcond^{(i)}$, and \eqref{eq:xmap} sends it to the rollout endpoint
    itself, $\Xmap{\xt^{(i)}}{\vcond^{(i)}}{0}=\xh^{(i)}$. Applying
    \eqref{eq:xidentity},
    \begin{equation}
    \mathcal L_{\mathrm{AWM}}
    =\mathbb E_{\,i,\,t,\,\epsilon}\Big[
    A^{(i)}\,\frac{w(t)}{t^{2}}\;\norm{\xz^{(i)}-\xh^{(i)}}^{2}\Big].
    \label{eq:awm-x}
    \end{equation}
    The method is a squared residual between the current prediction and the
    model's sample, weighted by the signed advantage. Taking $w(t)=t^{2}$
    cancels the prefactor and gives an $x$-space weight independent of $t$;
    this is the choice our few-step implementation makes up to a floor
    (\S\ref{sec:fewstep-awm}).
\end{example}
\begin{example}[DiffusionNFT in x-space]
The target of \eqref{eq:nftloss} is again $\vcond^{(i)}$, so \eqref{eq:xmap}
sends it to the same rollout endpoint $\xh^{(i)}$. Only the prediction
differs; the two implicit policies \eqref{eq:nft-implicit} are mapped to
\begin{equation}
\tilde x^{\pm}_\theta
=\Xmap{\xt^{(i)}}{v^{\pm}_\theta}{0}
=\tilde x^{(i)}_{\mathrm{old}}\pm\beta\big(\xz^{(i)}-\tilde x^{(i)}_{\mathrm{old}}\big),
\label{eq:nft-xbranch}
\end{equation}
with $\tilde x^{(i)}_{\mathrm{old}}$ as in \eqref{eq:xold-def}.
Applying \eqref{eq:xidentity} to each branch,
\begin{equation}
\mathcal L_{\mathrm{NFT}}
=\mathbb E_{\,i,\,t,\,\epsilon}\Big[
\frac{1}{t^{2}}\Big(
\hat r^{(i)}\,\norm{\tilde x^{+}_\theta-\xh^{(i)}}^{2}
+\big(1-\hat r^{(i)}\big)\,\norm{\tilde x^{-}_\theta-\xh^{(i)}}^{2}
\Big)\Big].
\label{eq:nft-x}
\end{equation}
The method mixes two squared residuals to the model's sample with weights
$\hat r^{(i)}$ and $1-\hat r^{(i)}$. The positive branch receives more weight
when the reward of the sample is above the group mean, and the negative branch
when it is below. The common prefactor $t^{-2}$ can be absorbed into the
weights, so the mixing itself is independent of $t$.
\end{example}

\subsection{The Decoupled Loss Decomposition}\label{sec:grad}

Under the transformation \eqref{eq:xmap}, the gradient of each objective
considered here is the transposed Jacobian
$(\partial\xz^{(i)}/\partial\theta)^{\top}$ applied to the sum of two vectors
in $x$-space (at the clean level by default and at level $r$ in the explicit
exceptions). We adopt this form as the general framework,
\begin{equation}
\nabla_\theta\mathcal L^{(i)}
=2\left(\frac{\partial\xz^{(i)}}{\partial\theta}\right)^{\!\!\top}
\Big[\;
\underbrace{W_{\mathrm{reg}}^{(i)}\big(\xz^{(i)}-\xreg^{(i)}\big)}
  _{\text{self-regression}}
\;-\;
\underbrace{W_{\mathrm{rwd}}^{(i)}\;\omega^{(i)}
\big(\xrwd^{(i)}-\xbase^{(i)}\big)}
  _{\text{reward augmentation}}
\Big],
\label{eq:generalgrad}
\end{equation}
with $W_{\mathrm{reg}}^{(i)},W_{\mathrm{rwd}}^{(i)}\ge0$ and
$\omega^{(i)}\in\mathbb R$ detached scalars. The factor $2$ from differentiating a
square is kept outside the bracket, so all weights below are defined relative
to \eqref{eq:generalgrad}; for example, the loss $W\norm{\xz-\xrwd}^{2}$
corresponds to $W_{\mathrm{rwd}}=W$. The prediction $\xz^{(i)}$ is
constructed by first noising the rollout sample $\xh^{(i)}$ to level $t$ and
then transporting the resulting state $\xt^{(i)}$ back to $r=0$ by
\eqref{eq:xmap}. Here $\omega^{(i)}$ is the signed reward coefficient used in the
general loss \eqref{eq:general}, and $|\omega^{(i)}|$ is its magnitude. For methods whose
coefficient follows the advantage, a non-zero $\omega^{(i)}$ has the sign of
$A^{(i)}$ in the theoretical weighting (the implemented DiffusionNFT weighting
of \S\ref{sec:fewstep-nft} can deviate), and a clipping gate may set it to zero.
\paragraph{Self-regression.} The first term does not depend on the reward. It
vanishes at $\xz^{(i)}=\xreg^{(i)}$ and grows with the distance from this
anchor, so its gradient moves the prediction back toward the anchor. This
self-regression term limits how far the reward-augmentation term can move the
prediction and therefore helps protect against reward hacking.

The regression anchor is therefore chosen as a prediction that the policy
should stay close to. A common choice is the old policy's prediction
$\tilde x_{\mathrm{old}}^{(i)}=\Xmap{\xt^{(i)}}{v_{\mathrm{old}}}{0}$, so that
the term penalises the deviation of the current prediction from that of the
policy used for sampling. Since $\xz^{(i)}$ is computed from $\xt^{(i)}$, the
anchor $\xreg^{(i)}$ must be computed from the same $\xt^{(i)}$; otherwise, the
difference $\xz^{(i)}-\xreg^{(i)}$ would mix a change of the velocity field with
a change of the state and would no longer measure the deviation that the term
is meant to bound.

\paragraph{Reward augmentation.} For fixed anchors, the second term contributes
a vector that does not depend on the current prediction and contains the
reward. Under gradient descent it moves $\xz^{(i)}$ along
$+\omega^{(i)}(\xrwd^{(i)}-\xbase^{(i)})$, a displacement whose
\emph{direction} is set by the sign of $\omega^{(i)}$ and whose coefficient has
magnitude $|\omega^{(i)}|$. The magnitude of the full term also depends on
$W_{\mathrm{rwd}}^{(i)}$ and on the anchor displacement
$\xrwd^{(i)}-\xbase^{(i)}$.

The rewarded anchor $\xrwd^{(i)}$ is usually the rollout sample, since
transporting $\xt^{(i)}$ with the conditional velocity gives
$\Xmap{\xt^{(i)}}{\vcond^{(i)}}{0}=\xh^{(i)}$. This choice is natural because
$A^{(i)}$ is obtained by scoring $\xh^{(i)}$ and provides no direct information about
other points. For methods whose coefficient follows the advantage, the sign of
$\omega^{(i)}$ determines whether the prediction moves toward $\xrwd^{(i)}$ or
away from it.

The other end, $\xbase^{(i)}$, is the baseline anchor. The displacement
$\xrwd^{(i)}-\xbase^{(i)}$ points from the baseline anchor to the rewarded
anchor, so $\xbase^{(i)}$ sets the reference point and the line along which the
reward acts. Admissible choices for $\xbase^{(i)}$ include
the old policy's prediction $\tilde x_{\mathrm{old}}^{(i)}$ and the
detached current prediction $\sg(\xz^{(i)})$. In the RL methods, $\xbase^{(i)}$ is also evaluated
at the state $\xt^{(i)}$ from which $\xz^{(i)}$ is computed, so that the two
anchors define a displacement in the space on which the update acts.

\paragraph{An intuitive example.} Consider a group-advantage-based method with
the group-mean advantage of \eqref{eq:adv-prelim} and a
group in which $A^{(1)}>A^{(2)}>0>A^{(3)}$. Members $1$ and $2$ both score
above the group mean, so $\omega^{(1)},\omega^{(2)}>0$ and both predictions are
attracted \emph{toward} their rewarded samples. Member $2$ is not penalised for
scoring lower than member $1$; it only receives a coefficient of smaller or equal magnitude, since
$|\omega^{(1)}|\ge|\omega^{(2)}|$ in AWM and in the theoretical DiffusionNFT
weighting below. Member $3$ scores below the mean, so $\omega^{(3)}<0$ and its
prediction is repelled \emph{away} from $\xrwd^{(3)}$, in the direction
$\xbase^{(3)}-\xrwd^{(3)}$. Whether a prediction is attracted or repelled is
therefore determined by the baseline-subtracted advantage $A^{(i)}$ and not by
the raw reward; a sample with a high reward is still repelled if it scores
below its group mean.

\paragraph{Recovering the loss.}  For any $V_\theta$ and any detached $U$,
\begin{equation}
-\left(\frac{\partial V_\theta}{\partial\theta}\right)^{\!\!\top}U
\;=\;\nabla_\theta\,\tfrac12\norm{V_\theta-\sg\big(V_\theta+U\big)}^{2}.
\label{eq:sgid}
\end{equation}
Therefore, applying \eqref{eq:sgid} with $V_\theta=\xz^{(i)}$ and
$U=g^{(i)}:=\omega^{(i)}(\xrwd^{(i)}-\xbase^{(i)})$ writes
\eqref{eq:generalgrad} as a sum of two squares,
\begin{equation}
\mathcal L^{(i)}
=\underbrace{W_{\mathrm{reg}}^{(i)}\norm{\xz^{(i)}-\xreg^{(i)}}^{2}}
  _{\textstyle \mathcal L^{(i)}_{\mathrm{reg}}}
\;+\;
\underbrace{W_{\mathrm{rwd}}^{(i)}\norm{\xz^{(i)}-\sg\big(\xz^{(i)}+g^{(i)}\big)}^{2}}
  _{\textstyle \mathcal L^{(i)}_{\mathrm{rwd}}},
\label{eq:general}
\end{equation}
which is the form \S\ref{sec:impl} implements.

\begin{example}[Decoupling AWM loss] Applying the same decomposition to
\eqref{eq:awm-x} gives a limiting case of \eqref{eq:generalgrad}.
Differentiating it and rewriting the residual as a difference of anchors,
\begin{equation}
\begin{aligned}
\nabla_\theta\mathcal L^{(i)}_{\mathrm{AWM}}
&=2\,\frac{w(t)}{t^{2}}\,A^{(i)}
\left(\frac{\partial\xz^{(i)}}{\partial\theta}\right)^{\!\!\top}
\big(\xz^{(i)}-\xh^{(i)}\big)\\[2pt]
&=-2\left(\frac{\partial\xz^{(i)}}{\partial\theta}\right)^{\!\!\top}
\frac{w(t)}{t^{2}}\,A^{(i)}
\,\big(\xh^{(i)}-\sg(\xz^{(i)})\big),
\end{aligned}
\label{eq:awm-gradx}
\end{equation}
which is \eqref{eq:generalgrad} with $W_{\mathrm{reg}}=0$:
\begin{equation}
W_{\mathrm{reg}}^{(i)}=0,
\qquad
\xrwd^{(i)}=\xh^{(i)},
\qquad
\xbase^{(i)}=\sg\big(\xz^{(i)}\big),
\qquad
W_{\mathrm{rwd}}^{(i)}=\frac{w(t)}{t^{2}},
\qquad
\omega^{(i)}=A^{(i)} .
\label{eq:awm-decomp}
\end{equation}
The anchor $\xreg^{(i)}$ is arbitrary, since it is multiplied by zero.
Compared with the DiffusionNFT decomposition (\refex{ex:nft-decomp} below,
\eqref{eq:nft-decomp}), there are three main
differences. First, the baseline anchor is the detached current prediction, so
one end of the reward displacement is fixed and the other moves with the
policy. The term is therefore recomputed at every step and is proportional to
the residual $\xh^{(i)}-\sg(\xz^{(i)})$ instead of being a fixed displacement;
this is why AWM can be written as the single weighted square \eqref{eq:awm-x}.
Second, in \eqref{eq:awm-decomp} the magnitude $|\omega^{(i)}|=|A^{(i)}|$ is not clipped, whereas in
DiffusionNFT it saturates at $1/\beta$. Third, with $W_{\mathrm{reg}}^{(i)}=0$
the matching loss contains no self-regression term; this role is played by the
additional velocity-space proximal penalty $\norm{v_\theta-v_{\mathrm{old}}}^{2}$ of
\S\ref{sec:rl}, which under \eqref{eq:xidentity} is a self-regression toward
the old policy's prediction $\tilde x^{(i)}_{\mathrm{old}}$.
\end{example}

\begin{example}[Decoupling DiffusionNFT loss]\label{ex:nft-decomp}
Applying the same decomposition to \eqref{eq:nft-x} yields both terms of
\eqref{eq:generalgrad}. The two branches \eqref{eq:nft-xbranch} are affine in
the same current prediction, so
\begin{equation}
\frac{\partial\tilde x^{+}_\theta}{\partial\theta}
=\beta\,\frac{\partial\xz^{(i)}}{\partial\theta},
\qquad
\frac{\partial\tilde x^{-}_\theta}{\partial\theta}
=-\beta\,\frac{\partial\xz^{(i)}}{\partial\theta} .
\label{eq:nft-branchjac}
\end{equation}
Differentiating \eqref{eq:nft-x} and writing
$\xbase^{(i)}:=\tilde x^{(i)}_{\mathrm{old}}$, the two branches collapse onto
a single Jacobian. Since
$\hat r^{(i)}=\tfrac12+\tfrac12\,\clip(A^{(i)},-1,1)$,
\begin{equation}
\nabla_\theta\mathcal L^{(i)}_{\mathrm{NFT}}
=\frac{2\beta^{2}}{t^{2}}
\left(\frac{\partial\xz^{(i)}}{\partial\theta}\right)^{\!\!\top}
\Big[
\big(\xz^{(i)}-\xbase^{(i)}\big)
-\frac{\clip\big(A^{(i)},-1,1\big)}{\beta}
\big(\xh^{(i)}-\xbase^{(i)}\big)
\Big],
\label{eq:nft-gradx}
\end{equation}
which matches \eqref{eq:generalgrad} term by term, with a self-regression
toward $\xbase^{(i)}$ and a reward augmentation along the fixed direction
$\xh^{(i)}-\xbase^{(i)}$. The corresponding anchors and weights are
\begin{equation}
\xreg^{(i)}=\xbase^{(i)}=\Xmap{\xt^{(i)}}{v_{\mathrm{old}}}{0},
\quad
\xrwd^{(i)}=\xh^{(i)},
\quad
W_{\mathrm{reg}}^{(i)}=W_{\mathrm{rwd}}^{(i)}=\frac{\beta^{2}}{t^{2}},
\quad
\omega^{(i)}=\frac{\clip\big(A^{(i)},-1,1\big)}{\beta} .
\label{eq:nft-decomp}
\end{equation}
Both anchors coincide at the old policy's clean prediction, so
DiffusionNFT regularises toward the point from which it measures the reward.
The rewarded sample is the rollout produced by the model. The two weights are
equal, which allows the published loss to be written as one fused square up to a detached term. The
signed reward coefficient $\omega^{(i)}=\clip(A^{(i)},-1,1)/\beta$ has a
magnitude that increases with $|A^{(i)}|$ and saturates at $1/\beta$ because of
the clip. In comparison, methods with $W_{\mathrm{reg}}^{(i)}=0$, such as AWM above, drop the self-regression
from the matching loss.
\end{example}

\begin{example}[Reward Augmentation for DMD and MeanFlow]\label{ex:ra-mfdmd}
    The decomposition also applies to reward-free distillation losses and
    yields their decoupled form. Neither objective scores a group, so there is
    no advantage and no index $(i)$; both move the policy toward the teacher.
    In the decomposition this is the limiting case
    \[
    \omega\equiv1 ,
    \]
    i.e., a unit positive coefficient for every sample, with the rewarded
    anchor set to the teacher's prediction
    \begin{equation}
    \xreal=\Xmap{\xt}{v_{\mathrm{real}}}{0}=\xt-tv_{\mathrm{real}} .
    \label{eq:distill-rwd}
    \end{equation}
    
    \emph{MeanFlow.} The network output in \eqref{eq:mfloss} is an average
    velocity, so the corresponding instantaneous velocity is recovered
    before \eqref{eq:xmap} is applied. Rearranging the MeanFlow identity
    \eqref{eq:mfid} with \eqref{eq:jvp} recovers it as $\uth+\Jterm{\uth}{t}{r}$.
    The correction $J$ is detached, so we evaluate it, together with the
    average velocity it corrects, on the frozen copy $\uold$ of the network,
    which is $\sg(\uth)$ in distillation and the old policy in RL. In
    particular, in MeanFlow $\uold$ equals $\uth$ in value; we write $\uold$
    only to unify the notation with the RL methods below:
    \begin{equation}
    \vind(\xt,t,r)\;:=\;\uold(\xt,t,r)+\Jterm{\uold}{t}{r},
    \label{eq:vind}
    \end{equation}
    and we call $\vind$ the \emph{self-induced instantaneous velocity}. It is
    computed from the average velocity and its total derivative with
    respect to $t$, without a second network, and coincides with the
    instantaneous velocity of $\uold$ wherever the flow map is exact. Its
    clean prediction is
    \begin{equation}
    \xind=\Xmap{\xt}{\vind}{0}=\xt-t\,\vind ,
    \label{eq:xz-vind}
    \end{equation}
    where the subscript indicates the velocity used. Both $\vind$ and $\xind$
    are detached.
    
    Both fields are now instantaneous, so \eqref{eq:xmap} transports them over the
    same interval. According to \eqref{eq:mfloss}, the prediction in $x$-space is extended from the clean level to any level $r<t$. 
    Transporting with the teacher and with the self-induced velocity to
    level $r$,
    \begin{equation}
    \xrealr=\Xmap{\xt}{v_{\mathrm{real}}}{r}=\xt+(r-t)\,v_{\mathrm{real}},
    \qquad
    \xindr=\Xmap{\xt}{\vind}{r}=\xt+(r-t)\,\vind ,
    \label{eq:x-step}
    \end{equation}
    whose clean ends at $r=0$ are \eqref{eq:distill-rwd} and \eqref{eq:xz-vind}.
    The policy's own transport with the same detached $J=\Jterm{\uold}{t}{r}$ is
    $\Xmap{\xt}{\uth+J}{r}=\xzr+(r-t)J$, with $\xzr=\Xmap{\xt}{\uth}{r}$, and
    equals $\xindr$ in value. For $r<t$, \eqref{eq:mfloss} is
    \begin{equation}
    \mathcal L_{\mathrm{MF}}
    =\mathbb E_{\,\xt,\,t,\,r}\!\left[
    \frac{\norm{\Xmap{\xt}{\uth+J}{r}-\xrealr}^{2}}{(t-r)^{2}}
    \right],\qquad r<t .
    \label{eq:mf-x}
    \end{equation}
    At $r=t$ the quotient is defined by continuity:
    \[
    \lim_{r\uparrow t}
    \frac{\norm{\Xmap{\xt}{\uth+J}{r}-\xrealr}^{2}}{(t-r)^{2}}
    =
    \norm{\uth(\xt,t,t)-v_{\mathrm{real}}(\xt,t)}^{2}.
    \]
    Thus MeanFlow is the squared velocity residual written as a scaled distance
    between level-$r$ predictions. To obtain the loss form from \eqref{eq:generalgrad},
    we differentiate the average velocity, since $J$
    is detached:
    \begin{equation}
    \frac{\partial}{\partial\theta}\Xmap{\xt}{\uth+J}{r}
    =\frac{\partial}{\partial\theta}\Big[\xt+(r-t)\big(\uth+\Jterm{\uold}{t}{r}\big)\Big]
    =\frac{\partial\xzr}{\partial\theta},
    \label{eq:mf-jac}
    \end{equation}
    where $\xzr$ is the one-step transport \eqref{eq:rollout-op} used by the
    sampler. Since $J$ is detached, it shifts the prediction without changing its
    Jacobian. For $r<t$, the gradient therefore corresponds to the reward augmentation,
    \begin{equation}
    \nabla_\theta\mathcal L_{\mathrm{MF}}
    =-\,2\left(\frac{\partial\xzr}{\partial\theta}\right)^{\!\top}
    \frac{1}{(t-r)^{2}}\,\big(\xrealr-\xindr\big),
    \label{eq:mf-gradx}
    \end{equation}
    which is the second term of \eqref{eq:generalgrad}, as
    \begin{equation}
    W_{\mathrm{reg}}=0,\quad
    \xrwd=\xrealr,\quad
    \xbase=\xindr,\quad
    W_{\mathrm{rwd}}=\frac{1}{(t-r)^{2}},\quad
    \omega=1 .
    \label{eq:mf-decomp}
    \end{equation}
    These $x$-space definitions apply for $r<t$. At $r=t$ we use the
    velocity-space limit above; no finite value of
    $W_{\mathrm{rwd}}=(t-r)^{-2}$ is assigned to this case.
    Here the Jacobian is $\partial\xzr/\partial\theta$ instead of
    $\partial\xz/\partial\theta$, because the reward augmentation acts at level
    $r$, which is resampled at every update. 
    
    \emph{DMD.} The DMD gradient \eqref{eq:dmdloss} contains a velocity
    difference, and \eqref{eq:xmap} converts it into a difference of clean
    predictions. With $\xz$ in place of the sampler output $\xhth$, the reverse
    KL of DMD evaluates the two fields at the noised prediction
    $\xt'=\AddNoise\big(\sg(\xz),t\big)$ rather than at $\xt$; we write
    $\xreal=\Xmap{\xt'}{v_{\mathrm{real}}}{0}$ and
    $\xaux=\Xmap{\xt'}{\vph}{0}$ for the teacher's and the auxiliary
    network's predictions at that state, so that
    \begin{equation}
    v_{\mathrm{real}}(\xt',t)-\vph(\xt',t)=-\frac{1}{t}\big(\xreal-\xaux\big).
    \label{eq:dmd-xdiff}
    \end{equation}
    The noised prediction is affine in $\xz$, so its Jacobian passes through to
    $\partial\xz/\partial\theta$,
    \begin{equation}
    \nabla_\theta\mathcal L_{\mathrm{DMD}}
    =-\frac{1-t}{t^2}
    \left(\frac{\partial\xz}{\partial\theta}\right)^{\!\!\top}
    \big(\xreal-\xaux\big).
    \label{eq:dmd-gradx}
    \end{equation}
    With this convention, \eqref{eq:dmd-gradx} is \eqref{eq:generalgrad} with
    \begin{equation}
    \xrwd=\xreal,\quad
    \xbase=\xaux,\quad
    W_{\mathrm{reg}}=0,\quad
    W_{\mathrm{rwd}}=\frac{1-t}{2t^2},\quad
    \omega=1 .
    \label{eq:dmd-decomp}
    \end{equation}
    
    The two distillation implementations therefore differ mainly in (i) how the
    instantaneous velocity of the current model distribution $p_\theta$ is
    estimated and (ii) the state at which $\xreal$ and $\xbase$ are evaluated.
    \begin{itemize}
        \item DMD \emph{parameterises} the instantaneous velocity with a second network $\vph$ with
        parameters $\phi$, fitted online to the student's samples by
        \eqref{eq:auxloss}, and sets $\xbase=\xaux$ at $\xt'$. It uses the
        clean-end prediction because \eqref{eq:dmd-gradx} uses
        $\partial\xz/\partial\theta$; varying $t$ distributes the update over
        noise levels but always trains the same map $\uth(\cdot,t,0)$ to the
        endpoint of the trajectory.
        \item MeanFlow \emph{induces} the instantaneous velocity, computing $\vind=\uold+\Jterm{\uold}{t}{r}$
        from the network itself, and sets $\xbase=\xindr$, whose $r=0$ case is
        the clean end \eqref{eq:xz-vind}. Since $\vind$ is defined at every
        queried $(t,r)$, this restriction disappears; \eqref{eq:mf-gradx} uses
        $\partial\xzr/\partial\theta$, and resampling $r$ at every update trains
        $\uth(\cdot,t,r)$ for all $r\in[0,t]$.
    \end{itemize}
    \end{example}
    
\subsection{Adaptation to Few-Step Models}\label{sec:fewstep}

The methods of \S\ref{sec:rl} are built on an instantaneous velocity, where the clean predictions in
\S\ref{sec:frame} are $\xz^{(i)}=\Xmap{\xt^{(i)}}{v_\theta}{0}$,
 \eqref{eq:general} uses $\partial\xz^{(i)}/\partial\theta$,
and the anchors apply the same map to $v_{\mathrm{old}}$ or $v_{\mathrm{real}}$.
A few-step policy instead outputs an average velocity $\uth(\xt,t,r)$. To evaluate
\eqref{eq:general}, the velocity in \eqref{eq:xmap} can be replaced by a velocity recovered from $\uth$, and the anchors by the matching
predictions of $\uold$ or the teacher. Therefore, methods differ in which predictions and
anchors they use. A rollout sample $\xh^{(i)}$ is already a sample and needs no
reconstruction, and the teacher prediction \eqref{eq:distill-rwd} remains
instantaneous, so only $\xz$ and the old-policy anchors need to be rebuilt.

\paragraph{Building anchors with u-prediction.}

The direct substitution queries only the $r=0$ edge of the domain
$\{0\le r\le t\le1\}$ of the flow map,
\begin{equation}
\xu=\Xmap{\xt}{\uth(\xt,t,0)}{0}=\xt-t\,\uth(\xt,t,0).
\label{eq:upred}
\end{equation}
The policy velocity $v_\theta(\xt,t)$ in \S\ref{sec:frame} is replaced by
$\uth(\xt,t,0)$, and the old policy is treated in the same way,
\begin{equation}
\xoldu=\Xmap{\xt}{\uold(\xt,t,0)}{0}=\xt-t\,\uold(\xt,t,0).
\label{eq:upred-old}
\end{equation}
The current prediction of \eqref{eq:general} is then $\xz\leftarrow\xu$, and
both the regression and the baseline anchor are $\xoldu$, or $\xbase=\sg(\xu)$ in the AWM case
where the baseline follows the policy.

The Jacobian in \eqref{eq:generalgrad} is then $\partial\xu/\partial\theta$,
so each update trains only $\uth(\cdot,t,0)$. The sampler \eqref{eq:rollout-op}
uses $\hat x_{\theta,r}=\hat x_{\theta,t}-(t-r)\,\uth(\cdot,t,r)$ with $r$ not
necessarily zero. Therefore, in u-prediction, the flow map is trained only on its $r=0$
edge and leaves the entries with $r\neq0$, which the sampler uses, without a
gradient.

\paragraph{Building anchors with v-prediction.}

In u-prediction, the instantaneous velocity is not used. Alternatively,
\eqref{eq:general} keeps $\xz=\Xmap{\xt}{v}{0}$, and the remaining question is
how to construct $v$ from a network that outputs an average velocity. Following
\refex{ex:ra-mfdmd}, there are two constructions.

\emph{Self-induced velocity.} \eqref{eq:vind} recovers the instantaneous
velocity from the average velocity itself, as in MeanFlowNFT, which applies
DiffusionNFT, regressed on the forward (noising) process, to an
average-velocity policy \citep{huang2026meanflownft}. The current prediction
adds the same detached $J$ to $\uth$, which gives the \emph{v-prediction}
\begin{equation}
\xv=\Xmap{\xt}{\uth+\Jterm{\uold}{t}{r}}{0}=\xt-t\big(\uth+\Jterm{\uold}{t}{r}\big),
\label{eq:vpred}
\end{equation}
and the old policy gives the self-induced anchor \eqref{eq:xz-vind},
\begin{equation}
\xind=\Xmap{\xt}{\vind}{0}=\xt-t\big(\uold+\Jterm{\uold}{t}{r}\big),
\label{eq:vpred-old}
\end{equation}
so that $\xv-\xind=-t(\uth-\uold)$. Since $J$ is detached, \eqref{eq:mf-jac} still holds; $J$
shifts the prediction but does not change its Jacobian. The pair $(t,r)$ is the
input of the network and $r$ is resampled at every update, so every
$\uth(\cdot,t,r)$ on the trajectory enters $\partial\xz/\partial\theta$. The
loss itself is still evaluated at $r=0$; $r$ enters only as a network input and through $J$, and
does not change the level at which \eqref{eq:general} is evaluated.

\emph{Auxiliary velocity.} The instantaneous velocity can instead be
parameterised by a second network $\vph$, fitted by \eqref{eq:auxloss} to the
samples of the model, so that its population target is the marginal velocity of
$p_\theta$,
\begin{equation}
\vph(\xt,t)\;\approx\;
\mathbb E_{\,\xh\sim p_\theta,\,\epsilon}\big[\epsilon - \hat x\mid\xt\big],
\qquad \xt=(1-t)\,\xh+t\,\epsilon .
\label{eq:vphi-cond}
\end{equation}
In the RL setting of \S\ref{sec:rl}, however, the rollouts are samples from $p_{\mathrm{old}}$
and not from $p_\theta$, so the target is the marginal velocity of
$p_{\mathrm{old}}$,
\begin{equation}
\vph(\xt,t)\;\approx\;
\mathbb E_{\,\xh\sim p_{\mathrm{old}},\,\epsilon}\big[\epsilon - \hat x\mid\xt\big],
\qquad \xt=(1-t)\,\xh+t\,\epsilon .
\label{eq:vphi-cond-old}
\end{equation}
Its clean prediction $\xaux=\Xmap{\xt}{\vph}{0}$ is detached, like $\xind$, and
serves only as an anchor. The current prediction cannot be computed from $\vph$, since
$\partial\xz/\partial\theta$ would then vanish and the reward would not affect
the policy. It is therefore computed from $\uth$, e.g., as $\xu$, $\xv$, or the
\emph{$J$-free $(t,r)$ query} $\Xmap{\xt}{\uth(\xt,t,r)}{0}$, i.e., the
transport of $\xt$ to the clean end with the average velocity queried at
$(t,r)$, which has the same Jacobian as
$\xv$. Its old-policy counterpart is $\xind$ with $J$ dropped,
$\xoldur=\Xmap{\xt}{\uold(\xt,t,r)}{0}$. The self-induced construction requires a Jacobian-vector product with a
stop-gradient; the auxiliary construction requires a second network and
optimiser that must follow a moving target.

Within the framework, a few-step method therefore does not introduce a new loss; it specifies how the
four points of \eqref{eq:general} are computed and which entries
$\uth(\cdot,t,r)$ of the flow map receive a gradient through
$\partial\xz/\partial\theta$. \Cref{tab:fewstep-forms} summarises the three
forms.

\begin{table}[tbp]
\centering
\renewcommand{\arraystretch}{1.25}
\begin{tabular}{@{}lccc@{}}
\toprule
& u-form & v-form, shared-$J$ (\texttt{v-ind}) & v-form, auxiliary (\texttt{v-aux}) \\
\midrule
$\xz$ & $\Xmap{\xt}{\uth(\xt,t,0)}{0}$ & $\Xmap{\xt}{\uth(\xt,t,r)+J}{0}$ & $\Xmap{\xt}{\uth(\xt,t,r)}{0}$ \\
$\xreg$ & $\Xmap{\xt}{\uold(\xt,t,0)}{0}$ & $\Xmap{\xt}{\vind}{0}$ & $\Xmap{\xt}{\uold(\xt,t,r)}{0}$ \\
$\xbase$ & $\Xmap{\xt}{\uold(\xt,t,0)}{0}$ & $\Xmap{\xt}{\vind}{0}$ & $\Xmap{\xt}{\vph}{0}$ \\
$\xrwd$ (rollout) & $\Xmap{\xt}{\vcond}{0}$ & $\Xmap{\xt}{\vcond}{0}$ & $\Xmap{\xt}{\vcond}{0}$ \\
trained $(t,r)$ & $(t,0)$  & $(t,r)$ with $0\leq r\leq t$ & $(t,r)$; $\vph$ queried at $(\xt,t)$ only \\
\bottomrule
\end{tabular}
\caption{The three few-step forms of the prediction and the anchors, with
$J=\Jterm{\uold}{t}{r}$, $\vind=\uold+J$ and $\Xmap{\xt}{\vcond}{0}=\xh$. The
labels \texttt{v-ind} and \texttt{v-aux} are used in \S\ref{sec:impl}.}
\label{tab:fewstep-forms}
\end{table}

The first column trains only the $r=0$ edge of the map; the second and the
third train all of it. The third column uses the auxiliary velocity only for
the baseline and computes $\xz$ and $\xreg$ at the same $(t,r)$ without $J$;
$\xz$ has the same Jacobian as $\xv$.

\begin{example}[DiffusionNFT with u-prediction]\label{ex:nft-upred}
With u-prediction, the weights and the rewarded sample of
\eqref{eq:nft-decomp} are unchanged, $W_{\mathrm{reg}}=W_{\mathrm{rwd}}=\beta^{2}/t^{2}$ and
$\xrwd=\xh$, and both anchors become the $r=0$ prediction of the old
policy,
\begin{equation}
\xz=\xu=\Xmap{\xt}{\uth}{0},
\qquad
\xreg=\xbase=\xoldu=\Xmap{\xt}{\uold}{0}.
\label{eq:upred-nft}
\end{equation}
Starting from the two squares of \eqref{eq:general},
\begin{equation}
\mathcal L_{\mathrm{reg}}
=
W_{\mathrm{reg}}\norm{\Xmap{\xt}{\uth}{0}-\Xmap{\xt}{\uold}{0}}^{2},
\label{eq:upred-nft-lreg}
\end{equation}
\begin{equation}
\begin{aligned}
\mathcal L_{\mathrm{rwd}}=
W_{\mathrm{rwd}}\norm{
\Xmap{\xt}{\uth}{0}
-
\sg\big(
\Xmap{\xt}{\uth}{0}
+\gu
\big)
}^{2},
\end{aligned}
\label{eq:upred-nft-lrwd}
\end{equation}
where $\xz=\Xmap{\xt}{\uth}{0}$ and
$\gu=\omega\big(\xh-\Xmap{\xt}{\uold}{0}\big)$.
The copy of $\xz$ inside $\sg$ does not cancel with the copy outside. No $J$
appears, the two anchors coincide, and $\partial\xz/\partial\theta$ involves
only $r=0$. This is DiffusionNFT evaluated on the $r=0$ edge $(t,0)$.
\end{example}

\begin{example}[DiffusionNFT with v-prediction]\label{ex:nft-vpred}
The decomposition \eqref{eq:nft-decomp} carries over, and the self-induced
construction below keeps $\xreg=\xbase$. The remaining choice is how to recover
the instantaneous velocity used in \eqref{eq:xmap} from an average-velocity
network, and \eqref{eq:vpred}--(\ref{eq:vphi-cond-old}) give two ways to do so.
The first is to use the self-induced velocity
$\vind=\uold+\Jterm{\uold}{t}{r}$, which adds the Jacobian-vector term
$\Jterm{\uold}{t}{r}$ to every prediction of the policy and of the old-policy
anchor, as in \eqref{eq:vpred}--(\ref{eq:vpred-old}); this construction gives
the \emph{MeanFlowNFT}-style variant \citep{huang2026meanflownft}. The second
is to fit an auxiliary network $\vph$ to the marginal
velocity of the rollouts, \eqref{eq:vphi-cond}--(\ref{eq:vphi-cond-old}), and
compute the baseline as $\xaux$, which leads to
\emph{DMDNFT}. In both cases the current prediction $\xz$ is computed from
$\uth$, since otherwise $\partial\xz/\partial\theta$ would vanish.

\textit{MeanFlowNFT} queries the network at $(t,r)$ with $0\leq r\leq t$, so
every prediction of the policy is a v-prediction. Following prior
directional-difference flow-map work \citep{nie2026tmd,gu2026anyflow}, the
MeanFlowNFT construction evaluates one derivative on the frozen old policy and
shares it between the current and the old policy
\citep{huang2026meanflownft}, which gives the four points
\begin{equation}
\begin{aligned}
\xz
&=
\xv=\Xmap{\xt}{\uth+\Jterm{\uold}{t}{r}}{0},
\\
\xreg
=
\xbase
&=
\xind=\Xmap{\xt}{\uold+\Jterm{\uold}{t}{r}}{0}
\\
\xrwd
&=
\xh .
\end{aligned}
\label{eq:vpred-nft}
\end{equation}
Starting from the two squares of \eqref{eq:general},
\begin{equation}
\begin{aligned}
\mathcal L_{\mathrm{reg}}
&=
W_{\mathrm{reg}}\norm{
\Xmap{\xt}{\uth+\Jterm{\uold}{t}{r}}{0}
-
\Xmap{\xt}{\uold+\Jterm{\uold}{t}{r}}{0}
}^{2}
\\
&=
W_{\mathrm{reg}}\norm{
\Xmap{\xt}{\uth(\xt,t,r)}{0}
-
\Xmap{\xt}{\uold(\xt,t,r)}{0}
}^{2},
\end{aligned}
\label{eq:vpred-nft-lreg}
\end{equation}
\begin{equation}
\begin{aligned}
\mathcal L_{\mathrm{rwd}}
&=
W_{\mathrm{rwd}}\norm{
\Xmap{\xt}{\uth+\Jterm{\uold}{t}{r}}{0}
-
\sg\big(
\Xmap{\xt}{\uth+\Jterm{\uold}{t}{r}}{0}
+\gv
\big)
}^{2}
\\
&=
W_{\mathrm{rwd}}\norm{
\Xmap{\xt}{\uth(\xt,t,r)}{0}
-
\sg\big(
\Xmap{\xt}{\uth(\xt,t,r)}{0}
+\gv
\big)
}^{2}.
\end{aligned}
\label{eq:vpred-nft-lrwd}
\end{equation}
The prediction $\xz$ is still the v-prediction of \eqref{eq:vpred-nft}.
In $\mathcal L_{\mathrm{reg}}$, the shared detached $J$ cancels between the
current and old-policy predictions; in $\mathcal L_{\mathrm{rwd}}$, it cancels
between the two copies of the current prediction. 
The reward displacement still contains the old-policy prediction,
\[
\gv=\omega\big(\xh-\Xmap{\xt}{\uold+\Jterm{\uold}{t}{r}}{0}\big),
\]
and therefore retains $J$. Since $J$ is detached, the Jacobian is that of
$\Xmap{\xt}{\uth(\xt,t,r)}{0}$. Resampling $r$ then updates the family $\uth(\cdot,t,r)$
instead of only its $r=0$ edge.

\textit{DMDNFT} is the alternative construction. In the losses, only the baseline changes,
with $\vph$ fitted online to the old-policy rollouts by
\eqref{eq:vphi-cond-old}. The current prediction is still computed from $\uth$,
since substituting $\vph$ would give $\partial\xv/\partial\theta=0$. As in
\eqref{eq:mf-jac}, $J$ is detached, so
\begin{equation}
\frac{\partial\xv}{\partial\theta}
=\frac{\partial}{\partial\theta}\Xmap{\xt}{\uth+\Jterm{\uold}{t}{r}}{0}
=\frac{\partial}{\partial\theta}\Xmap{\xt}{\uth}{0}.
\label{eq:vpred-nft-phi-jac}
\end{equation}
Therefore, the four points are
\begin{equation}
\begin{aligned}
\xz
&=\Xmap{\xt}{\uth(\xt,t,r)}{0},
\\
\xreg
&=\xoldur=\Xmap{\xt}{\uold(\xt,t,r)}{0},
\\
\xbase
&=\xaux,
\\
\xrwd
&=\xh .
\end{aligned}
\label{eq:vpred-nft-phi}
\end{equation}
\end{example}

\begin{example}[Few-step DiffusionOPSD]\label{ex:opsd}\textit{DiffusionOPSD} keeps the
training scheme of DiffusionNFT but \emph{constructs} the point toward which
the policy should move, which requires a reward $R(\dec(\cdot),c)$ that is
differentiable with respect to the clean sample. Three clean points are
constructed at one query state $\xt^{(i)}=\AddNoise(\xh^{(i)},t)$ at a small
noise level $t$ (drawn from a law concentrated below $0.5$,
\S\ref{sec:exp-setup}). The first is the old policy's prediction at this state,
given by \eqref{eq:xold-def},
\begin{equation}
\tilde x^{(i)}_{\mathrm{old}}
=\Xmap{\xt^{(i)}}{v_{\mathrm{old}}}{0}.
\label{eq:opsd-anchor}
\end{equation}
The other two points are obtained by $M_{\mathrm{tgt}}$ normalised
reward-gradient steps in $x$-space, in the ascent and in the descent direction.
Suppressing $(i)$ on the iterates and indexing them by $[m]$,
$\tilde x^{[0]}_{\pm}=\tilde x^{(i)}_{\mathrm{old}}$ and
\begin{equation}
\tilde x^{[m+1]}_{\pm}
=\tilde x^{[m]}_{\pm}\pm h\,
\frac{\nabla R\big(\dec(\tilde x^{[m]}_{\pm}),c\big)}
{\norm{\nabla R\big(\dec(\tilde x^{[m]}_{\pm}),c\big)}+\varepsilon_g},
\qquad
h=\frac{\eta\,\kappa\,\norm{\tilde x^{(i)}_{\mathrm{old}}}}{M_{\mathrm{tgt}}},
\label{eq:opsd-target}
\end{equation}
for $m=0,\dots,M_{\mathrm{tgt}}-1$, with a small floor $\varepsilon_g>0$, the
relative trust radius $\kappa>0$ and the total step length $\eta>0$ in units
of that radius. Each
iterate is projected onto the ball of
radius $\kappa\norm{\tilde x^{(i)}_{\mathrm{old}}}$ around the anchor; before
projection, $M_{\mathrm{tgt}}h$ is $\eta$ times that radius. The final points
$\tilde x^{(i)}_{\pm}=\sg(\tilde x^{[M_{\mathrm{tgt}}]}_{\pm})$ are detached.
For $\eta\le1$ the projection is inactive, and their displacements are
\begin{equation}
\Delta_{\mathrm{rwd}}\tilde x^{(i)}_{\pm}
:=\tilde x^{(i)}_{\pm}-\tilde x^{(i)}_{\mathrm{old}}
=\pm h\sum_{m=0}^{M_{\mathrm{tgt}}-1}
\frac{\nabla R\big(\dec(\tilde x^{[m]}_{\pm}),c\big)}
{\norm{\nabla R\big(\dec(\tilde x^{[m]}_{\pm}),c\big)}+\varepsilon_g}.
\label{eq:opsd-increment}
\end{equation} The labels $\pm$ indicate
the directions of construction. We construct the loss function in three steps:

\emph{(i)}~Fitting reuses the two $x$-space branches of
\eqref{eq:nft-xbranch}, now regressed onto the constructed points instead of
the rollout sample:
\begin{equation}
\begin{aligned}
\tilde x^{\pm}_\theta
&=\tilde x^{(i)}_{\mathrm{old}}\pm\beta\big(\xz^{(i)}-\tilde x^{(i)}_{\mathrm{old}}\big),
\\
\mathcal L_{\mathrm{OPSD}}
=\mathbb E_{\,i}\big[\mathcal L^{(i)}_{\mathrm{OPSD}}\big],
\qquad
\mathcal L^{(i)}_{\mathrm{OPSD}}
&=a_+\norm{\tilde x^{+}_\theta-\tilde x^{(i)}_{+}}^{2}
+a_-\norm{\tilde x^{-}_\theta-\tilde x^{(i)}_{-}}^{2},
\end{aligned}
\label{eq:opsdloss}
\end{equation}
with $a_+=\hat r^{(i)}/\gamma_+$ and $a_-=(1-\hat r^{(i)})/\gamma_-$ non-negative
and detached, $\hat r^{(i)}$ the weight of \eqref{eq:rhat-prelim} and
$\gamma_\pm=\max\{\sg\,\msp{|\tilde x^{\pm}_\theta-\tilde x^{(i)}_{\pm}|},
\varepsilon_\gamma\}$ the adaptive normalisers with floor $\varepsilon_\gamma>0$, and where 
$\msp{\cdot}$ denotes the per-sample mean, i.e., the average over the
$\Nel$ elements of one sample.

\emph{(ii)}~To decouple the loss into \eqref{eq:general} with gradient form
\eqref{eq:generalgrad}, note that the two branches are reflections of each
other about the anchor, so both residuals are affine in the same displacement
$\xz^{(i)}-\tilde x^{(i)}_{\mathrm{old}}$. Expanding the squares of
$\mathcal L^{(i)}_{\mathrm{OPSD}}$ and discarding the $\theta$-free remainder
therefore splits it into one square and two inner products, each written
directly on the three constructed points:
\begin{equation}
\begin{aligned}
\mathcal L^{(i)}_{\mathrm{OPSD}}
=\;&\underbrace{\beta^{2}(a_++a_-)\,
\norm{\xz^{(i)}-\tilde x^{(i)}_{\mathrm{old}}}^{2}}_{\text{self-regression}}\\
&-\underbrace{2\beta a_+\big(\xz^{(i)}-\tilde x^{(i)}_{\mathrm{old}}\big)^{\!\top}
\Delta_{\mathrm{rwd}}\tilde x^{(i)}_{+}}
_{\text{reward augmentation, positive}}
+\underbrace{2\beta a_-\big(\xz^{(i)}-\tilde x^{(i)}_{\mathrm{old}}\big)^{\!\top}
\Delta_{\mathrm{rwd}}\tilde x^{(i)}_{-}}
_{\text{reward augmentation, negative}}
\;+\;\mathrm{const},
\end{aligned}
\label{eq:opsd-x}
\end{equation}
the constant being $a_+\norm{\Delta_{\mathrm{rwd}}\tilde x^{(i)}_{+}}^{2}
+a_-\norm{\Delta_{\mathrm{rwd}}\tilde x^{(i)}_{-}}^{2}$, which has no
gradient. Differentiating gives three terms with the same Jacobian,
\begin{equation}
\nabla_\theta\mathcal L^{(i)}_{\mathrm{OPSD}}
=2\left(\frac{\partial\xz^{(i)}}{\partial\theta}\right)^{\!\!\top}
\Big[\beta^{2}(a_++a_-)\big(\xz^{(i)}-\tilde x^{(i)}_{\mathrm{old}}\big)
-\beta a_+\Delta_{\mathrm{rwd}}\tilde x^{(i)}_{+}
+\beta a_-\Delta_{\mathrm{rwd}}\tilde x^{(i)}_{-}\Big].
\label{eq:opsd-gradx}
\end{equation}
Equations~(\ref{eq:opsd-x}) and~(\ref{eq:opsd-gradx}) contain one
self-regression term and two reward-augmentation terms, indexed by $k=1,2$:
\begin{equation}
\begin{aligned}
&\xreg^{(i)}=\tilde x^{(i)}_{\mathrm{old}},
\qquad
W_{\mathrm{reg}}=W_{\mathrm{rwd}}=\beta^{2}(a_++a_-),\\[4pt]
&\tilde x^{(i)}_{\mathrm{rwd},1}=\tilde x^{(i)}_{+},
\quad
\tilde x^{(i)}_{\mathrm{base},1}=\tilde x^{(i)}_{\mathrm{old}},
\quad
\omega_1=
\frac{a_+}{\beta(a_++a_-)},\\[4pt]
&\tilde x^{(i)}_{\mathrm{rwd},2}=\tilde x^{(i)}_{-},
\quad
\tilde x^{(i)}_{\mathrm{base},2}=\tilde x^{(i)}_{\mathrm{old}},
\quad
\omega_2=
-\frac{a_-}{\beta(a_++a_-)}.
\end{aligned}
\label{eq:opsd-decomp}
\end{equation}
The self-regression is toward the old policy's prediction, as in
\eqref{eq:nft-decomp}. The two reward terms share this baseline,
$\tilde x^{(i)}_{\mathrm{base},1}=\tilde x^{(i)}_{\mathrm{base},2}=\xreg^{(i)}$,
and differ only in the constructed rewarded point $\tilde x^{(i)}_{\mathrm{rwd},k}$ and
its signed reward coefficient. The signs of $\omega_1$ and $\omega_2$ are not
determined by the endpoint advantage $A^{(i)}$; no advantage is defined for the
constructed points. They are fixed to be non-negative for the
ascent branch and non-positive for the descent branch.

Applying \eqref{eq:sgid} to each reward term separately gives a loss with
gradient \eqref{eq:opsd-gradx}, in the form \eqref{eq:general} with two reward
squares,
\begin{equation}
\begin{aligned}
\mathcal L^{(i)}_{\mathrm{OPSD}}
&=
\mathcal L^{(i)}_{\mathrm{reg}}
+\mathcal L^{(i)}_{\mathrm{rwd},1}
+\mathcal L^{(i)}_{\mathrm{rwd},2},
\\
\mathcal L^{(i)}_{\mathrm{reg}}
&=
W_{\mathrm{reg}}\norm{\xz^{(i)}-\xreg^{(i)}}^{2},
\\
\mathcal L^{(i)}_{\mathrm{rwd},k}
&=
W_{\mathrm{rwd}}\norm{\xz^{(i)}-\sg\big(\xz^{(i)}+g^{(i)}_{k}\big)}^{2},
\qquad k=1,2,
\end{aligned}
\label{eq:opsd-decoupled}
\end{equation}
where
$g^{(i)}_{k}=\omega_k\big(\tilde x^{(i)}_{\mathrm{rwd},k}-\tilde x^{(i)}_{\mathrm{base},k}\big)$,
i.e.\
$g^{(i)}_{1}=\omega_1\Delta_{\mathrm{rwd}}\tilde x^{(i)}_{+}$
and
$g^{(i)}_{2}=\omega_2\Delta_{\mathrm{rwd}}\tilde x^{(i)}_{-}$.
The branch magnitudes satisfy
$|\omega_1|+|\omega_2|=1/\beta$, so $\hat r^{(i)}$ redistributes that total
between the two branches. It can also change the common weight
$W=\beta^2(a_++a_-)$ whenever $\gamma_+\ne\gamma_-$. At
$\hat r^{(i)}\to1$ only the ascent displacement remains, at
$\hat r^{(i)}\to0$ only the descent displacement remains, and the
self-regression and two reward-augmentation contributions sum to zero at
$\xz^{(i)}=\xreg^{(i)}+g^{(i)}_{1}+g^{(i)}_{2}$.

\emph{(iii)}~For an average-velocity policy we take the v-prediction form
of Equations~(\ref{eq:vpred})--(\ref{eq:vpred-old}), as in \refex{ex:nft-vpred}. Suppressing
$(i)$ and writing $J=\Jterm{\uold}{t}{r}$, the corresponding quantities are
\begin{equation}
\begin{aligned}
\xz
&=\Xmap{\xt}{\uth+J}{0},
\\
\xreg=\tilde x_{\mathrm{base},1}=\tilde x_{\mathrm{base},2}
&=\xind=\Xmap{\xt}{\uold+J}{0},
\\
\tilde x_{\mathrm{rwd},1}&=\tilde x_{+},
\qquad
\tilde x_{\mathrm{rwd},2}=\tilde x_{-},
\end{aligned}
\label{eq:opsd-v-decomp}
\end{equation}
where the gradient steps \eqref{eq:opsd-target} are not started at $\xind$,
which contains $J$, but at the $r=0$ prediction $\xoldu$ of
\eqref{eq:upred-old}, and only the resulting displacement is transferred to
the anchor:
\begin{equation}
\Delta=\xoldu-\xind,
\qquad
\tilde x_{\pm}=\big[\text{apply \eqref{eq:opsd-target} from }\tilde x^{[0]}_\pm=\xoldu\big]-\Delta .
\label{eq:opsd-shift}
\end{equation}
The
weights are unchanged. We expand the three squares of
\eqref{eq:opsd-decoupled} in turn.

The self-regression is the self-regression square of \eqref{eq:vpred-nft-lreg}:
\begin{equation}
\mathcal L_{\mathrm{reg}}
=W_{\mathrm{reg}}\norm{\Xmap{\xt}{\uth(\xt,t,r)}{0}-\Xmap{\xt}{\uold(\xt,t,r)}{0}}^{2}.
\label{eq:opsd-v-lreg}
\end{equation}
Both copies of $J$ are the same detached estimate computed from $\uold$, so
they cancel.

The two reward squares are
\begin{equation}
\begin{aligned}
\mathcal L_{\mathrm{rwd},k}
&=W_{\mathrm{rwd}}\norm{
\Xmap{\xt}{\uth+J}{0}
-\sg\big(\Xmap{\xt}{\uth+J}{0}+g_k\big)}^{2}
\\
&=W_{\mathrm{rwd}}\norm{
\Xmap{\xt}{\uth(\xt,t,r)}{0}
-\sg\big(\Xmap{\xt}{\uth(\xt,t,r)}{0}+g_k\big)}^{2},
\qquad k=1,2,
\end{aligned}
\label{eq:opsd-v-lrwd}
\end{equation}
where, by \eqref{eq:opsd-v-decomp},
\begin{equation}
g_k=\omega_k\big(\tilde x_{\mathrm{rwd},k}-\tilde x_{\mathrm{base},k}\big),
\qquad\text{i.e.}\qquad
g_1=\omega_1\big(\tilde x_{+}-\xind\big),
\quad
g_2=\omega_2\big(\tilde x_{-}-\xind\big).
\label{eq:opsd-v-g}
\end{equation}
Because $\tilde x_\pm-\xind$ is constructed from $\xoldu$ by
\eqref{eq:opsd-shift}, $J$ also cancels from both reward squares
(\Cref{app:opsd}). Hence no square
of the loss contains $J$, \eqref{eq:opsd-decoupled} depends only on
$\xz=\Xmap{\xt}{\uth(\xt,t,r)}{0}$, $\xreg=\xoldur$ and the $J$-free $g_k$,
and by \eqref{eq:mf-jac} the Jacobian is that of
$\Xmap{\xt}{\uth(\xt,t,r)}{0}$. FewStepOPSD in v-prediction form thus reduces
to the $J$-free $(t,r)$ query of \S\ref{sec:fewstep} for every $(t,r)$, not
only at $r=0$.
\end{example}

\begin{example}[ReFL as Few-step DiffusionOPSD]\label{ex:refl}
Reward feedback learning (\textit{ReFL}) backpropagates the gradient of a
differentiable reward through a single late denoising step. Classic ReFL stops
a rollout of the old-policy sampler at a small noise level $t$ without gradient and
lets the policy predict the clean sample from that state with one network
evaluation; as in \refex{ex:opsd}, we instead evaluate the network at the
re-noised state $\xt^{(i)}=\AddNoise(\xh^{(i)},t)$ and maximise the reward of
the resulting prediction,
\begin{equation}
\mathcal L_{\mathrm{ReFL}}
=-\lambda\,\mathbb E_{\,i,\,t}\Big[R\big(\dec(\xz^{(i)}),c\big)\Big],
\qquad
\xz^{(i)}=\Xmap{\xt^{(i)}}{v_\theta}{0},
\label{eq:refl-loss}
\end{equation}
with a loss weight $\lambda>0$. Only this final evaluation is differentiated,
\begin{equation}
\nabla_\theta\mathcal L^{(i)}_{\mathrm{ReFL}}
=-\lambda\left(\frac{\partial\xz^{(i)}}{\partial\theta}\right)^{\!\!\top}
\nabla R\big(\dec(\xz^{(i)}),c\big),
\label{eq:refl-grad}
\end{equation}
where $\nabla R(\dec(\cdot),c)$ is the gradient with respect to the clean
sample, as in \eqref{eq:opsd-target}. The backward pass therefore propagates
through the decoder and the reward model.

Consider \eqref{eq:opsd-decoupled}
with a single reward-gradient step, $M_{\mathrm{tgt}}=1$, no self-regression,
$W_{\mathrm{reg}}=0$, and only the ascent term $\mathcal L_{\mathrm{rwd},1}$,
with any detached $W_{\mathrm{rwd}},\omega_1>0$. For $\eta\le1$ the single step
is shorter than $\kappa\norm{\tilde x_{\mathrm{old}}}$, the projection is
inactive, and \eqref{eq:opsd-increment} reads
\begin{equation}
\tilde x_{+}-\tilde x_{\mathrm{old}}
=h\,\frac{\nabla R\big(\dec(\tilde x_{\mathrm{old}}),c\big)}
{\norm{\nabla R\big(\dec(\tilde x_{\mathrm{old}}),c\big)}+\varepsilon_g}.
\label{eq:refl-increment}
\end{equation}
The loss is the single square
$\mathcal L_{\mathrm{rwd},1}
=W_{\mathrm{rwd}}\norm{\xz-\sg(\xz+g_1)}^{2}$ of \eqref{eq:opsd-decoupled},
using \eqref{eq:opsd-decomp},
$\tilde x_{\mathrm{rwd},1}=\tilde x_{+}$,
$\tilde x_{\mathrm{base},1}=\tilde x_{\mathrm{old}}$ and
$\omega_1>0$. Its gradient \eqref{eq:generalgrad} is
\begin{equation}
\begin{aligned}
\nabla_\theta\mathcal L_{\mathrm{rwd},1}
&=-2\left(\frac{\partial\xz}{\partial\theta}\right)^{\!\!\top}
W_{\mathrm{rwd}}\omega_1
\big(\tilde x_{\mathrm{rwd},1}-\tilde x_{\mathrm{base},1}\big)
=-2\,W_{\mathrm{rwd}}\,\omega_1
\left(\frac{\partial\xz}{\partial\theta}\right)^{\!\!\top}
\big(\tilde x_{+}-\tilde x_{\mathrm{old}}\big)
\\
&=-\frac{2W_{\mathrm{rwd}}\,\omega_1\,h}
{\norm{\nabla R\big(\dec(\tilde x_{\mathrm{old}}),c\big)}+\varepsilon_g}
\left(\frac{\partial\xz}{\partial\theta}\right)^{\!\!\top}
\nabla R\big(\dec(\tilde x_{\mathrm{old}}),c\big).
\end{aligned}
\label{eq:refl-proxy}
\end{equation}
Compared with \eqref{eq:refl-grad}, \eqref{eq:refl-proxy} replaces $\lambda$
by a detached, positive, sample-dependent factor and evaluates the reward
gradient at the old policy's prediction instead of at $\xz$ (at
$\tilde x_{\mathrm{old}}$ for an instantaneous-velocity policy, and at
$\xoldu$ in the few-step form below). It is therefore only
a proxy for ReFL. If $\theta_{\mathrm{old}}=\theta$ at the evaluation point,
which holds at the first inner iteration after each refresh when the EMA-decay
ratio is $0.0$, then $\xz=\tilde x_{\mathrm{old}}$ and the per-sample
directions are collinear; exact equality additionally requires the positive
factor to equal $\lambda$. Thus the DiffusionOPSD
configuration with a single ascent step and no self-regression is a detached
proxy of ReFL, by setting $M_{\mathrm{tgt}}=1$, $W_{\mathrm{reg}}=0$, $\mathcal L_{\mathrm{rwd},2}=0$ and the EMA-decay
ratio as $0.0$.

For the implemented few-step form, the current prediction and the old-policy
anchor follow \eqref{eq:opsd-v-decomp}, while the reward displacement is
computed at the $r=0$ prediction $\xoldu$ and transferred to $\xind$ by
\eqref{eq:opsd-shift}:
\begin{equation}
\begin{aligned}
\tilde x_{\mathrm{base},1}
&=\xind=\Xmap{\xt}{\uold+\Jterm{\uold}{t}{r}}{0},
\\
\Delta_{\mathrm{rwd}}\tilde x_+
&=h\,
\frac{\nabla R\big(\dec(\xoldu),c\big)}
{\norm{\nabla R\big(\dec(\xoldu),c\big)}+\varepsilon_g},
\qquad
\tilde x_{\mathrm{rwd},1}
=\tilde x_+=\xind+\Delta_{\mathrm{rwd}}\tilde x_+,
\\
g_1
&=\omega_1
\big(\tilde x_{\mathrm{rwd},1}-\tilde x_{\mathrm{base},1}\big)
=\omega_1\Delta_{\mathrm{rwd}}\tilde x_+ .
\end{aligned}
\label{eq:refl-anchor}
\end{equation}
The \emph{reward augmentation loss} is \eqref{eq:opsd-v-lrwd} with this $g_1$,
\begin{equation}
\mathcal L_{\mathrm{rwd},1}
=W_{\mathrm{rwd}}\norm{
\Xmap{\xt}{\uth(\xt,t,r)}{0}
-\sg\big(\Xmap{\xt}{\uth(\xt,t,r)}{0}+g_1\big)}^{2},
\label{eq:refl-fewstep}
\end{equation}
whose reward augmentation moves the prediction $\Xmap{\xt}{\uth(\xt,t,r)}{0}$ at each sampled $(t,r)$ along
$\nabla R(\dec(\xoldu),c)$. Apart from $W_{\mathrm{rwd}}$ and the velocity penalty, a step requires one detached rollout, two forward
passes of $\uold$ (at $r$ and at $r=0$), one reward
gradient, and one forward and backward pass of $\uth$; $J$ cancels, because it
shifts both ends of the displacement equally, and is therefore dropped. With
$J$ dropped and $r=0$
we have $\xind=\xoldu$, which is the u-form. At a generic $r$, only the
current prediction and its anchor use $(t,r)$; the decoder and the reward are
still evaluated at the clean prediction $\xoldu$, while the resulting
displacement trains $\uth(\cdot,t,r)$.
\end{example}

\section{Implementation}\label{sec:impl}

The framework of \S\ref{sec:frame} is implemented as one base trainer whose
method \mintinline{python}{step} runs one outer iteration. Each trainer of
\S\S\ref{sec:fewstep-grpo}--\ref{sec:FewStepOPSD} is a subclass that overrides
a few stage methods, and the instantaneous-velocity estimator of \S\ref{sec:fewstep},
\texttt{v-ind} or \texttt{v-aux}, is a separate object passed to the trainer.
Here \texttt{v-ind} uses the
MeanFlow identity with the shared old-policy correction $J$ to construct an
instantaneous velocity from an average velocity, whereas \texttt{v-aux} uses an auxiliary velocity to
approximate the instantaneous velocity of the old-policy rollout distribution
$p_{\mathrm{old}}$ (which DMD takes to be $p_\theta$), as in \eqref{eq:vphi-cond-old}. Stages 1 and 2 below operate on groups, since they
produce and score the $K$ members of a prompt, and keep the index $(i)$ as in
\S\S\ref{sec:prelim}--\ref{sec:fewstep}. From Stage 3 onward the framework acts
on one member at a time. The prediction, the
three anchors and the three weights are per-sample quantities; the index $(i)$
is omitted in the equations and in the loss terms $\mathcal L_{\mathrm{reg}}$,
$\mathcal L_{\mathrm{rwd}}$, and becomes the leading batch axis in code.

\subsection{A General Training Step}\label{sec:impl-base}

One outer iteration fixes an old policy $\uold$, rolls it out, scores the
rollouts, and then takes one gradient step on the mean over $n_{\mathrm{it}}$ queries of
\begin{equation}
\mathcal L(\theta)
=\mathbb E
\Big[W_{\mathrm{reg}}\,\msp{\big(\xz-\xreg\big)^2}
+W_{\mathrm{rwd}}\,\msp{\big(\xz-\sg(\xz+g)\big)^2}
+\lambda_{\mathrm{KL}}\,\mathcal L_{\mathrm{KL}}\Big],
\label{eq:impl-objective}
\end{equation}
where $\xh^{(i)}=\Rollout(\epsilon^{(i)},\uold,c)$ is a detached rollout of
prompt $c$. The per-sample mean $\msp{\cdot}$ rescales every weight $W$ by
$1/\Nel$ relative to the squared norms of \S\ref{sec:grad}; this constant is
absorbed into the learning rate. 
$\mathcal L_{\mathrm{KL}}$ is the velocity penalty of Stage~4 with weight
$\lambda_{\mathrm{KL}}$, evaluated on the same method-specific query.

\paragraph{The training workflow.} One outer iteration consists of five
operations, executed in order; the four method-specific stages of
\S\ref{sec:intro} are operations 1, 2, 4 and the loss inside 4, while
operations 3 and 5 are shared bookkeeping.
\begin{enumerate}
    \item \emph{Rollout}: the old policy $\uold$ samples $K$ members per
    prompt with the sampler of Stage~1, and the endpoints $\xh^{(i)}$ are detached.
    \item \emph{Reward and advantage}: each endpoint is decoded and scored,
    $R^{(i)}=R\big(\dec(\xh^{(i)}),c\big)$, and Stage~2 turns the rewards into the
    advantage $A^{(i)}$ and the mixing weight $\hat r^{(i)}$ (\eqref{eq:adv},
    \eqref{eq:rhat}). 
    \item \emph{Auxiliary fit}: a trainer that uses the auxiliary velocity
    $\vph$ first fits it to these rollouts with the flow-matching loss
    \eqref{eq:auxfit} for $n_{\mathrm{fm}}$ steps, before the policy is updated;
    the other trainers skip this step.
    \item \emph{Training loss}: $n_{\mathrm{it}}$ inner iterations each draw
    a query (Stage~3\,\emph{(i)}), compute the prediction, anchors and coefficients
    (Stage~3\,\emph{(ii)}--\emph{(iv)}), and accumulate the gradient of \eqref{eq:impl-objective}
    (Stage~4); one optimiser step is taken on their mean.
    \item \emph{Refresh}: $\uold$ becomes the EMA of $\uth$, and the next
    iteration rolls out with it.
\end{enumerate}

In code, this is the method \mintinline{python}{step} of the base trainer in
\Cref{ex:my-code}, one block per operation: \mintinline{python}{rollout},
\mintinline{python}{advantage}, \mintinline{python}{estimator.fit}, the loop over
\mintinline{python}{queries} with \mintinline{python}{anchors},
\mintinline{python}{weights} and \mintinline{python}{loss}, and
\mintinline{python}{refresh_old}. Each \mintinline{python}{Query} object is a
view of the batch for one method-specific training query; fields it does not
set, such as the advantage, are read from the batch.

A trainer of \S\S\ref{sec:fewstep-grpo}--\ref{sec:FewStepOPSD} is a subclass
that implements \mintinline{python}{anchors} and \mintinline{python}{weights}, and
overrides another hook only where it changes that stage (\Cref{tab:hooks}).
\mintinline{python}{anchors} returns the prediction and three anchors of
\eqref{eq:master}, with a tuple of rewarded anchors for FewStepOPSD;
\mintinline{python}{weights} returns the detached coefficients, with $\omega$
signed and a tuple for FewStepOPSD, and \mintinline{python}{loss} is shared by
every trainer except FewStepOPSD and ReFL, which override it. The named methods of
\S\ref{sec:exp} other than ReFL differ from their family only in the estimator they use.

\begin{pbox}[label={ex:my-code}]{The base trainer: one outer iteration}
    \begin{minted}[linenos, breaklines, mathescape, fontsize=\footnotesize, xleftmargin=2em]{python}
@dataclass
class Points:                          # prediction and anchors of Eq. ($\ref{eq:master}$)
    x_z: Tensor                        # the only tensor with a gradient
    x_reg: Tensor | None               # None disables self-regression
    x_base: Tensor | None              # None disables reward augmentation
    x_rwd: Tensor | tuple[Tensor, ...]  # a tuple for multi-branch FewStepOPSD

class FewStepRLTrainer:
    estimator = VInd()                 # Stage 3(ii)-(iii): \texttt{v-ind} or \texttt{v-aux}
    n_it = n_ts                        # default; FewStepGRPO uses $M_{\mathrm{step}}$, FewStepOPSD uses 1

    def step(self, prompts):                         # ONE outer iteration
        batch = self.rollout(prompts)                # Stage 1, detached
        R = reward(decode(batch.x_hat), batch.c)     # Stage 2
        batch.A, batch.r_hat = self.advantage(R, batch.c)
        self.estimator.fit(batch)                    # \texttt{v-aux} only, Eq. ($\ref{eq:auxfit}$)
        for q in self.queries(batch):                # Stage 3(i): $n_{\mathrm{it}}$ queries
            p = self.anchors(q)                      # Stage 3(ii)-(iii)
            W_reg, W_rwd, omega = self.weights(q, p) # Stage 3(iv)
            loss = self.loss(p, W_reg, W_rwd, omega) # Stage 4
            accumulate(mean_batch(loss + lam_kl * self.kl(q)) / self.n_it)
        self.update()                                # one step on $\theta$
        self.refresh_old()                           # $\uold\leftarrow$ EMA of $\uth$

    def rollout(self, prompts):                      # Eq. ($\ref{eq:rollout-indep}$)
        return ode_rollout(prompts, u_old)
    def advantage(self, R, c):                       # Eqs. ($\ref{eq:adv}$), ($\ref{eq:rhat}$)
        return adv_and_rhat(R, c)
    def queries(self, batch):                        # Stage 3(i)
        for _ in range(self.n_it):
            t, r = draw_tr(batch)        # $r\mid t\sim U(0,t)$; pinned $r=t$, $r=0$
            eps = randn_like(batch.x_hat)            # fresh noise at every query
            yield Query(batch, t=t, r=r, eps=eps, x_t=add_noise(batch.x_hat, t, eps))
    def anchors(self, q):     raise NotImplementedError   # -> Points
    def weights(self, q, p):  raise NotImplementedError   # -> W_reg, W_rwd, omega
    def loss(self, p, W_reg, W_rwd, omega):          # Code $\ref{ex:loss-ops}$; OPSD overrides
        return loss(p, W_reg, W_rwd, omega)
    \end{minted}
  \end{pbox}

\begin{table}[tbp]
\centering
\footnotesize
\setlength{\tabcolsep}{4pt}
\renewcommand{\arraystretch}{1.3}
\resizebox{\textwidth}{!}{%
\begin{tabular}{@{}llccccccc@{}}
\toprule
& & \multirow{2}{*}{FewStepGRPO}& \multicolumn{2}{c}{FewStepAWM} & \multicolumn{2}{c}{FewStepNFT} & \multicolumn{2}{c}{FewStepOPSD}\\
\cmidrule(l){4-5}\cmidrule(l){6-7} \cmidrule(l){8-9}
stage & quantity &  & MeanFlowAWM & DMDAWM & MeanFlowNFT & DMDNFT & -general$^{\dagger}$ & ReFL \\
\midrule
\texttt{rollout}  & & SDE & ODE & ODE & ODE & ODE & ODE with $\tilde x_\pm$ & ODE with $\tilde x_+$ \\
\texttt{advantage} & & --- & --- & --- & --- & --- & --- & $\hat r\equiv1$ \\
\texttt{queries}  & & $(t,s,r)$, $s$ drawn as $r$ & $(t,r)$ & $(t,r)$ & $(t,r)$ & $(t,r)$ & $(t,r)$ & $(t,r)$ \\
\multirow{3}{*}{\texttt{anchors} }
 & $\xreg$ & $\varnothing$ & $\varnothing$ & $\varnothing$ & $\xind$ & $\xoldur$$^{\ddagger}$ & $\xoldur$$^{\ddagger}$ & $\xoldur$$^{\ddagger}$ \\
 & $\xbase$ & $\sg(\mu_\theta)$ & $\sg(\xz)$ & $\xaux$ & $\xind$ & $\xaux$ & $\xoldur$$^{\ddagger}$ & $\xoldur$$^{\ddagger}$ \\
 & $\xrwd$ & $x_r$ & $\xh$ & $\xh$ & $\xh$ & $\xh$ & $\tilde x_\pm$ & $\tilde x_+$ \\
\multirow{3}{*}{\texttt{weights}}
 & $W_{\mathrm{reg}}$ & 0 & 0 & 0 & $c_{\mathrm{adv}}\beta^{2}(a_++a_-)$\textsuperscript{$\star$} & $c_{\mathrm{adv}}\beta^{2}(a_++a_-)$\textsuperscript{$\star$} & $c_{\mathrm{adv}}\beta^{2}(a_++a_-)$\textsuperscript{$\star$} & $0$ \\
 & $W_{\mathrm{rwd}}$ & $\bar\rho_{\mathrm{GRPO}}$ & $\rho_{\mathrm{AWM}}$  & $\rho_{\mathrm{AWM}}$ & $W_{\mathrm{reg}}$ & $W_{\mathrm{reg}}$ & $W_{\mathrm{reg}}$ & $c_{\mathrm{adv}}\beta^{2}/\gamma_+$\textsuperscript{$\star$} \\
 & $\omega$ & $\frac{\chi_{\mathrm{GRPO}}}{2\sigma_{\mathrm{k}}^{2}}A$ & $\frac{\chi_{\mathrm{AWM}}}{2\tau_{\mathrm{AWM}}^{2}}A$ & $\frac{\chi_{\mathrm{AWM}}}{2\tau_{\mathrm{AWM}}^{2}}A$
 & $\frac{a_+-a_-}{\beta(a_++a_-)}$ & $\frac{a_+-a_-}{\beta(a_++a_-)}$ & $\frac{(a_+,-a_-)}{\beta(a_++a_-)}$\textsuperscript{$\star$} & $+1/\beta$\textsuperscript{$\star$} \\
\texttt{loss} & & $\mathcal L_{\mathrm{rwd}}$ & $\mathcal L_{\mathrm{rwd}}$ & $\mathcal L_{\mathrm{rwd}}$ & $\mathcal L_{\mathrm{rwd}}$, $\mathcal L_{\mathrm{reg}}$ & $\mathcal L_{\mathrm{rwd}}$, $\mathcal L_{\mathrm{reg}}$ & $\mathcal L_{\mathrm{rwd},1}$,$\mathcal L_{\mathrm{rwd},2}$,  $\mathcal L_{\mathrm{reg}}$  & $\mathcal L_{\mathrm{rwd},1}$ \\
\midrule
estimator & & \texttt{v-ind}, replay traj. & \texttt{v-ind} & \texttt{v-aux} & \texttt{v-ind} & \texttt{v-aux} & \texttt{v-ind} & \texttt{v-ind} \\
\bottomrule
\end{tabular}}

\smallskip
{\raggedright\scriptsize 
$^{\dagger}$-general is FewStepOPSD in its general form, with both the ascent
and the descent points $\tilde x_\pm$. The ReFL proxy configuration sets
$\hat r\equiv1$, keeps only the ascent point $\tilde x_+$, and sets
$W_{\mathrm{reg}}=0$.
$^{\ddagger}$$\xoldur=\Xmap{\xt}{\uold(\xt,t,r)}{0}$, used
together with the $J$-free $(t,r)$ query $\Xmap{\xt}{\uth(\xt,t,r)}{0}$.
Dropping $J$ is exact, because $J$ cancels in every difference whose two terms
contain the same detached $J$. In DMDNFT it cancels in the self-regression square,
\eqref{eq:vpred-nft-phi}, while the baseline $\xaux$ contains no
$J$; in FewStepOPSD and ReFL it cancels in every square of the loss,
\eqref{eq:opsd-v-lreg}--\eqref{eq:opsd-v-lrwd}, so the loss and its gradient are the same for every detached
$J$ (\Cref{app:opsd-cancel}).
$^{\star}$$a_+=\hat r/\gamma_+$ and $a_-=(1-\hat r)/\gamma_-$, with the adaptive
normalisers $\gamma_\pm$ of \eqref{eq:nft-gamma}: measured against $\xh$ in
MeanFlowNFT, centred at $\xbase$ and displaced along $\sg(\xz)-\xreg$ in DMDNFT,
where $\xreg\neq\xbase$, and measured against the constructed points $\tilde x_\pm$ in
FewStepOPSD and ReFL (\eqref{eq:opsd-weights}). The factor
$c_{\mathrm{adv}}$ scales $W$ in NFT and general OPSD to undo the compression
of the advantage in \eqref{eq:rhat} (\S\ref{sec:fewstep-nft}), and is kept by the
ReFL override. In FewStepOPSD and ReFL the sign of $\omega$ is fixed by
construction, $+$ for the ascent term and $-$ for the descent term
(\eqref{eq:opsd-decomp}), and is not determined by $A$.\par}
\caption{Specific implementation of stages of \Cref{ex:my-code} for each trainer of
\S\ref{sec:exp}. The method-specific quantities $\mu_\theta$, $\bar\rho_{\mathrm{GRPO}}$,
$\rho_{\mathrm{AWM}}$, $\chi_{\mathrm{GRPO}}$, $\chi_{\mathrm{AWM}}$, $\sigma_{\mathrm{k}}$,
$\tau_{\mathrm{AWM}}$, $a_\pm$, $\gamma_\pm$ and $c_{\mathrm{adv}}$ are defined in
\S\S\ref{sec:impl-stages}--\ref{sec:fewstep-nft}.}
\label{tab:hooks}
\end{table}

\subsection{Introduction to the Key Stages}\label{sec:impl-stages}

This subsection describes, for each stage of \Cref{ex:my-code}, the default
computation, its cost, and the choices that the framework determines.

\paragraph{Stage 1: the rollout operator}\label{sec:rollout}

Let $\Rollout$ denote the sampler. We distinguish two cases.

\textit{(a) Sample-independent rollout.} For every trainer except the FewStepOPSD
family, the group members of a prompt are i.i.d.\ draws:
\begin{equation}
\xh^{(i)} \;=\; \Rollout\big(\epsilon^{(i)},\,\uold,\,c\big),
\qquad \epsilon^{(i)}\stackrel{\text{iid}}{\sim}\mathcal N(0,I),
\quad i=1,\dots,K,
\label{eq:rollout-indep}
\end{equation}
i.e.\ the only stochasticity is the initial noise (and, for the stochastic
sampler below, the per-step renoise). Members do not interact.

The sampler uses the old policy $\uold$ and not $\uth$, as indicated by
the argument of \eqref{eq:rollout-indep}. One old policy is fixed per
outer iteration, so that the rewarded endpoint $\xh^{(i)}$, the regression
anchor and the baseline anchor all refer to the same distribution, as required
in \S\ref{sec:grad}, where $\xreg^{(i)}$ and $\xbase^{(i)}$ must be evaluated
at the state from which $\xz^{(i)}$ is computed. If the samples were drawn from
$\uth$, the reward would be evaluated on a distribution different from the one
described by the anchors, and $A^{(i)}$ would not be the advantage of the
policy that defines the anchors.

\textit{(b) Reward-coupled rollout.} For reward-coupled rollout, FewStepOPSD extends
the rollout with the reward,
\begin{equation}
\big\{\xh^{(i)},\,\tilde{x}_{\mathrm{old}}^{\mathrm{u},(i)},\,\tilde x^{(i)}_{\pm}\big\}_{i=1}^{K}
\;=\;
\Rollout\big(\epsilon^{(i)},\,\uold,\,R,\,c\big).
\label{eq:rollout-dep}
\end{equation}
The members remain i.i.d.; the reward does not couple different members.
For each member, the sampler of \eqref{eq:rollout-indep} runs unchanged.
The reward-coupled construction then applies \eqref{eq:opsd-target} from the
$r=0$ prediction $\tilde{x}_{\mathrm{old}}^{\mathrm{u},(i)}$ and transfers
the resulting displacements to the anchor by \eqref{eq:opsd-shift}, giving
$\tilde x^{(i)}_{\pm}$ together with the endpoint.

In code, \mintinline{python}{rollout} defaults to (a) with the deterministic
sampler; FewStepGRPO overrides it with the stochastic sampler of \S\ref{sec:fewstep-grpo}, keeping the
transitions, and FewStepOPSD appends the reward-coupled construction of (b).

\paragraph{Stage 2: reward and group advantage}

The generic advantage \eqref{eq:adv-prelim} takes both the mean and the
standard deviation over the group; the implementation changes the population of the standard deviation.
With $R^{(i)}=R\big(\dec(\xh^{(i)}),c\big)$ the
endpoint reward ($\dec$ the VAE decoder), the advantage is
\begin{equation}
{\;
A^{(i)}
=\frac{R^{(i)}-\bar R_{c}}
{\operatorname{std}\big(\{R\}_{\mathcal B}\big)+10^{-4}},
\qquad
\bar R_{c}=\frac{1}{|H_c|}\sum_{R\in H_c} R ,
\;}
\label{eq:adv}
\end{equation}
where $H_c$ is the set of rewards of prompt $c$ in the current outer iteration and $\{R\}_{\mathcal B}$ is the \emph{current global batch}
gathered across all ranks. The numerator and denominator use different
populations. The per-prompt mean reduces the effect of prompt difficulty, and
the global standard deviation avoids division by a near-zero value when all
rewards in a group are equal.

The advantage is then compressed into a bounded mixing weight, generalising
\eqref{eq:rhat-prelim} by replacing the unit clip range with a tunable
$c_{\mathrm{adv}}$:
\begin{equation}
{\;
\hat r^{(i)} \;=\;
\frac{\clip\big(A^{(i)},\,-c_{\mathrm{adv}},\,c_{\mathrm{adv}}\big)}
{2\,c_{\mathrm{adv}}}+\frac12\;\in[0,1].
\;}
\label{eq:rhat}
\end{equation}
Here $\hat r\to1$ means ``far above the prompt's mean'', $\hat r\to0$ ``far
below'', $\hat r=\tfrac12$ ``average''. This $[0,1]$ form distinguishes the
NFT/OPSD families from FewStepGRPO and FewStepAWM, which use
$\clip(A,-c_{\mathrm{adv}},c_{\mathrm{adv}})$ as a
\emph{signed multiplier} (\S\ref{sec:grpo}); $A$ in \eqref{eq:grpo-decomp} and
\eqref{eq:awm-fewstep-decomp} denotes this clipped value.

In code, \mintinline{python}{advantage} returns $(A,\hat r)$; ReFL overrides it
with $\hat r\equiv1$.

\paragraph{Stage 3: anchors and weights}\label{sec:stage3}

The trainers differ mainly in this stage. It receives the detached rollout and
returns the current prediction, the anchors and the detached coefficients;
FewStepOPSD returns branch-indexed tuples for its rewarded anchors and $\omega$.
We describe it in four parts, namely \emph{(i)} the query levels, \emph{(ii)} the two ways to
compute the clean prediction, \emph{(iii)} the two estimators of the instantaneous
velocity that these predictions require, and \emph{(iv)} the weights.

\textit{(i) The training query $(t,r)$.}
By default, a fresh pair $(t,r)$ is drawn at every inner iteration;
``$r$ is resampled at every update'' in \S\ref{sec:fewstep} refers to this draw: $t$ is
drawn from a selectable law (the maximum of two uniforms or a logit-normal),
$r\mid t\sim U(0,t)$, and a fixed fraction of the batch is set to $r=t$ and to
$r=0$. The three cases play different roles. For $r=t$ the interval has zero
length and $J\equiv0$, so the network query is instantaneous and the loss
reduces to its instantaneous-velocity form; $r=0$ trains the one-step generator and gives $J$ its
largest prefactor $t-r$; the generic case lets $\partial\xz/\partial\theta$
reach $\uth(\cdot,t,r)$ for every $r$ in between, as required by the second
column of \Cref{tab:fewstep-forms}.

\textit{(ii) The u- and v-predictions.}
Except for the replayed transition score of FewStepGRPO, the implemented reward
losses evaluate \eqref{eq:xmap} at the clean level $r=0$, but the network
outputs the average velocity $\uth(\xt,t,r)$ instead of the instantaneous
velocity required by the map. Using $\uth$ directly gives the u-form:
\begin{equation}
{\;\xz=\xu=\Xmap{\xt}{\uth(\xt,t,0)}{0}\;}
\qquad\text{(u-form)}.
\label{eq:P-u}
\end{equation}
Adding the detached correction of \eqref{eq:vind} gives the
v-form; since the prediction carries the gradient, \texttt{v-ind} builds it with the self-induced velocity,
i.e.,
\begin{equation}
{\;\xz=\xv=\Xmap{\xt}{\uth(\xt,t,r)+\Jterm{\uold}{t}{r}}{0}\;}
\qquad\text{(v-form, \texttt{v-ind})}.
\label{eq:P-v}
\end{equation}
As in \eqref{eq:vind}, $J$ is computed from $\uold$,
so that a single estimate can be shared by all predictions, as required
in \emph{(iii)}. Since $J$ is detached, by \eqref{eq:mf-jac} the Jacobian of the
v-form $\xz$ is that of $\Xmap{\xt}{\uth(\xt,t,r)}{0}$.

For the anchors $\xbase$, $\xreg$ and $\xrwd$,
there are two alternatives in v-form, namely \texttt{v-aux} and \texttt{v-ind}.

\textit{(iii) Estimating the instantaneous velocity.} There are two estimators
of the instantaneous velocity, following MeanFlow (a.k.a. \texttt{v-ind}) and DMD (a.k.a. \texttt{v-aux}), respectively.

\emph{Self-induced $\vind$, with shared $J$} (\texttt{v-ind}). The v-form needs
$J$ only as a detached vector. The directional-difference estimate, for a step
size $\varepsilon_{\mathrm{cd}}>0$, is
\begin{equation}
\Jterm{\uold}{t}{r}
=(t-r)\,
\frac{
\uold\!\big(\xt+\varepsilon_{\mathrm{cd}}\vcond,t+\varepsilon_{\mathrm{cd}},r\big)
-\uold\!\big(\xt-\varepsilon_{\mathrm{cd}}\vcond,t-\varepsilon_{\mathrm{cd}},r\big)}
{2\varepsilon_{\mathrm{cd}}}.
\label{eq:cd}
\end{equation} For $r=t$ the prefactor $t-r$ makes it zero. It requires two additional
forward passes of $\uold$ per query, and one estimate is shared by the current
prediction and the anchors, so the correction cancels in the self-regression
difference $\xz-\xreg$ and across $\sg(\cdot)$, as discussed in Example~\ref{ex:nft-vpred}.

\emph{Auxiliary $\vph$, trained online} (\texttt{v-aux}). $\vph$ is instead trained at every update by plain
flow matching on the current rollout pool,
\begin{equation}
\mathcal L_{\mathrm{aux}}
=\msp{\big(\vph(\xt,t)-\vcond\big)^{2}},
\qquad
\xt=\AddNoise(\xh,t),
\label{eq:auxfit}
\end{equation}
running $n_{\mathrm{fm}}$ steps of \eqref{eq:auxfit} per outer iteration,
before the policy update. The auxiliary-to-policy optimiser-step ratio is
$n_{\mathrm{fm}}:1$ ($n_{\mathrm{fm}}=5$ in
\S\ref{sec:exp}). This requires no extra rollout and no $J$, but it requires a
second network and optimiser that must follow a target that changes with the
policy, as stated in \refex{ex:nft-vpred}.

\Cref{ex:estimators-code} implements the two instantaneous-velocity estimators
as estimator objects. Their method \mintinline{python}{fields} returns the
current and old-policy velocities of one query with one $J$, so the estimate of
\emph{(iii)} is shared by construction; \mintinline{python}{VInd} adds the correction
\eqref{eq:cd}, \mintinline{python}{ReplayVInd} uses \eqref{eq:cd-replay},
\mintinline{python}{VInd(cd=False)} drops $J$ where it cancels
(\Cref{app:opsd}), and \mintinline{python}{VAux} adds $\vph$ and fits it once
per outer iteration by \eqref{eq:auxfit}.

\begin{pbox}[label={ex:estimators-code}]{Instantaneous-velocity estimators}
    \begin{minted}[linenos, breaklines, mathescape, fontsize=\footnotesize, xleftmargin=2em]{python}
class VelocityEstimator:                        # Stage 3(ii)-(iii)
    def r(self, q):        return q.r           # lower endpoint of the flow map
    def J(self, q, r):     return 0.            # detached correction $\Jterm{\uold}{t}{r}$
    def v_base(self, q):   return None          # field of $\xbase$, \texttt{v-aux} only
    def fit(self, batch):  pass                 # once per outer iteration
    def fields(self, q, r=None):                # ONE $J$ for current and old fields
        r = self.r(q) if r is None else r
        J = self.J(q, r)
        v_th = u_theta(q.x_t, q.t, r) + J       # current field; differentiated
        v_old = sg(u_old(q.x_t, q.t, r)) + J    # old policy
        return v_th, v_old, self.v_base(q)

class VInd(VelocityEstimator):                  # Eq. ($\ref{eq:P-v}$)
    def __init__(self, cd=True): self.cd = cd   # cd=False: $J$ cancels in the loss
    def directional_J(self, q, lower, tangent):  # Eq. ($\ref{eq:cd}$), elementwise
        if not self.cd: return 0.
        up = u_old(q.x_t + eps_cd * tangent, q.t + eps_cd, lower)
        dn = u_old(q.x_t - eps_cd * tangent, q.t - eps_cd, lower)
        return (q.t - lower) * sg(up - dn) / (2 * eps_cd)
    def J(self, q, r):
        return self.directional_J(q, r, q.v_cond)  # tangent $\vcond=\epsilon-\xh$

class ReplayVInd(VInd):                         # Eq. ($\ref{eq:cd-replay}$), GRPO only
    def J(self, q, s):
        return self.directional_J(q, s, (q.x_t - q.x_hat) / q.t)  # two forwards

class VAux(VelocityEstimator):                  # Eq. ($\ref{eq:vpred-nft-phi}$): no $J$
    def v_base(self, q):   return sg(v_phi(q.x_t, q.t))
    def fit(self, batch):                       # Eq. ($\ref{eq:auxfit}$), before the update
        for _ in range(n_fm):
            t, eps = draw_t(batch), randn_like(batch.x_hat)
            x_t = add_noise(batch.x_hat, t, eps)
            v_cond = eps - batch.x_hat
            aux_step(mean_sp((v_phi(x_t, t) - v_cond) ** 2))
    \end{minted}
  \end{pbox}

\textit{(iv) The loss weights.} \eqref{eq:generalgrad} leaves three
per-sample, detached weights free,
\begin{equation}
\big(W_{\mathrm{reg}},\ W_{\mathrm{rwd}},\ \omega\big),
\label{eq:weights}
\end{equation}
namely the weight of the self-regression term, the weight of the
reward-augmentation term, and the signed reward coefficient of the reward step. The
four families differ mainly in these weights, which are given in
\eqref{eq:grpo-decomp}, \eqref{eq:awm-fewstep-decomp}, \eqref{eq:nft-weights}
and \eqref{eq:opsd-weights} and returned, with the sign of $\omega$, by the hook
\mintinline{python}{weights}.

\paragraph{Stage 4: the decoupled two-term loss}

This stage implements \eqref{eq:general} without the group index and with the
norms replaced by the per-sample mean $\msp{\cdot}$ of \Cref{app:notation},
which is the reduction used in the code. Given
$\big(\xz;\ \xreg,\xbase,\xrwd;\ W_{\mathrm{reg}},W_{\mathrm{rwd}},\omega\big)$,
every trainer except FewStepOPSD, which adds a second reward square
(\S\ref{sec:FewStepOPSD}), minimises
\begin{equation}
{\;
\mathcal L
\;=\;
\underbrace{W_{\mathrm{reg}}\,\msp{\big(\xz-\xreg\big)^2}}_{\textstyle \mathcal L_{\mathrm{reg}}
\ \text{(self-regression)}}
\;+\;
\underbrace{W_{\mathrm{rwd}}\,\msp{\big(\xz-\sg(\xz+g)\big)^2}}_{\textstyle \mathcal L_{\mathrm{rwd}}
\ \text{(reward augmentation)}},
\;}
\label{eq:master}
\end{equation}
with the \emph{reward displacement}
\begin{equation}
g \;=\; \omega\big(\xrwd-\xbase\big)
\qquad\text{(detached).}
\label{eq:g}
\end{equation}

With a single reward term, the base interface is specified by six entries:
\[
\big(\ \xreg,\ \xbase,\ \xrwd,\ W_{\mathrm{reg}},\ W_{\mathrm{rwd}},\ \omega\ \big).
\]
\S\S\ref{sec:fewstep-grpo}--\ref{sec:FewStepOPSD} specify these quantities, one family
at a time, from the derivations of Examples~\ref{ex:nft-upred}--\ref{ex:refl};
FewStepOPSD provides branch-indexed
$(\tilde x_{\mathrm{rwd},k},\omega_k)$ entries.
Setting $\xreg=\varnothing$ switches off $\mathcal L_{\mathrm{reg}}$;
setting $\xbase=\varnothing$ switches off $\mathcal L_{\mathrm{rwd}}$.

\begin{pbox}[label={ex:loss-ops}]{The decoupled loss}
    \begin{minted}[linenos, breaklines, mathescape, fontsize=\footnotesize, xleftmargin=2em]{python}
def self_regression(x_z, x_reg, W_reg):         # $\mathcal L_{\mathrm{reg}}$
    return W_reg * mean_sp((x_z - x_reg) ** 2)

def reward_augmentation(x_z, x_rwd, x_base, omega, W_rwd):   # $\mathcal L_{\mathrm{rwd}}$
    g = omega * (x_rwd - x_base)                # Eq. ($\ref{eq:g}$); omega is signed
    # value $-g$, Jacobian the identity: Eq. ($\ref{eq:sgid}$) holds exactly
    return W_rwd * mean_sp((x_z - sg(x_z + g)) ** 2)

def loss(p, W_reg, W_rwd, omega):               # Stage 4, Eq. ($\ref{eq:master}$)
    l = 0.
    if p.x_reg is not None:                     # $\xreg=\varnothing$: no self-regression
        l = l + self_regression(p.x_z, p.x_reg, W_reg)
    if p.x_base is not None:                    # $\xbase=\varnothing$: no reward augmentation
        l = l + reward_augmentation(p.x_z, p.x_rwd, p.x_base, omega, W_rwd)
    return l
    \end{minted}
  \end{pbox}

\Cref{ex:loss-ops} implements Stage 4 except the velocity penalty. The $\varnothing$ cases correspond to the two
\mintinline{python}{None} checks, so every trainer except FewStepOPSD and ReFL runs the same
\mintinline{python}{loss} regardless of the anchors it defines.

In addition to \eqref{eq:master}, an implemented loss may include a velocity
penalty against a frozen reference with its own weight, which
encourages the policy to stay close to the reference. This penalty is a separate term and
not an additional anchor. The average-velocity
reference $u_{\mathrm{ref}}$ is the initial pre-RL policy and remains fixed
throughout training.
For an endpoint-renoise query the KL-u penalty is
\begin{equation}
\mathcal L^{\mathrm u}_{\mathrm{KL}}
=\msp{\big(\uth(\xt,t,r)-u_{\mathrm{ref}}(\xt,t,r)\big)^{2}}.
\label{eq:kl-uref}
\end{equation}

When the reference is an instantaneous velocity, here the teacher
$v_{\mathrm{real}}$, it cannot be compared with $\uth$ directly for $r\neq t$. We therefore
convert it to an average-velocity target as in MeanFlow and regress $\uth$ onto
it (the KL-v penalty),
\begin{equation}
\mathcal L^{\mathrm v}_{\mathrm{KL}}
=\msp{\Big(\uth(\xt,t,r)-\big(v_{\mathrm{real}}(\xt,t)-\Jterm{\uold}{t}{r}\big)\Big)^{2}}.
\label{eq:kl-vref}
\end{equation}

\subsection{FewStepGRPO: Replaying Stored Trajectories}\label{sec:fewstep-grpo}

FewStepGRPO is our replay-based surrogate adaptation of FlowGRPO
\citep{liu2025flowgrpo}.

\paragraph{Stochastic sampler.}
The stochastic sampler generalises \eqref{eq:sde-step}. Each step first moves
from $t_j$ to a level $s_j\le t_{j+1}$, drawn uniformly from the remaining grid
levels, and then adds noise to return to level $t_{j+1}$ on the linear path,
\begin{equation}
\hat x_{s_j}=\Xmap{\hat x_{t_j}}{\uold(\hat x_{t_j},t_j,s_j)}{s_j},
\qquad
\hat x_{t_{j+1}}
=\frac{1-t_{j+1}}{1-s_j}\,\hat x_{s_j}+\tau_j\,\epsilon',
\qquad \epsilon'\sim\mathcal N(0,I),
\label{eq:rollout-sde}
\end{equation}
where the leading coefficient matches the clean component at level $t_{j+1}$
and $\tau_j$ matches the noise variance,
$\tau_j^{2}=t_{j+1}^{2}-\big(\tfrac{1-t_{j+1}}{1-s_j}\big)^{2}s_j^{2}$,
which is non-negative if and only if $s_j\le t_{j+1}$.
In \eqref{eq:rollout-sde}, $s_j$ is the level at which the flow map is
\emph{queried} and $t_{j+1}$ is the \emph{target} level of the step; in the
notation of \eqref{eq:rollout-op}, the flow map is thus queried with lower
endpoint $s_j$ rather than $r=t_{j+1}$, and $r=t_{j+1}$ denotes the target
level of the transition. \eqref{eq:sde-step} is the
case $s_j=0$, with $t'$ in place of $t_{j+1}$. 

\paragraph{Replay-trajectory query.}
Unlike endpoint-renoise trainers, it retains the
$M_{\mathrm{step}}$ transitions of the stochastic rollout, replays one stored $(x_t,x_r,t,r)$ per inner
iteration, and redraws only the flow-map query level $s$, with the law of $r$ in
Stage~3\,\emph{(i)}. The redrawn $s$ need not equal the grid level $s_j$ used
at rollout time; \Cref{app:grpo-kernel} shows that the scored kernel is
unaffected by this choice at zero drift.
Thus $n_{\mathrm{it}}=M_{\mathrm{step}}$, one endpoint advantage supervises
all transitions of a member, and the differentiated quantity is the mean
$\mu_\theta$ of the scoring kernel at level $r$ instead of a clean prediction.

\paragraph{Scoring kernel.}
FewStepGRPO scores $x_r$ with an Euler--Maruyama Gaussian, which reads
\begin{equation}
\begin{aligned}
\mu_\theta
&:=\alpha_t\,x_t-b_t\,\big(\uth(x_t,t,s)+J\big),\\
\mu_{\mathrm{old}}
&:=\alpha_t\,x_t-b_t\,\big(\uold(x_t,t,s)+J\big),\\
\alpha_t&:=1-\frac{\varsigma_t^{2}}{2t}\,\Delta t,\qquad
b_t:=\Big(1+\frac{\varsigma_t^{2}(1-t)}{2t}\Big)\Delta t,\qquad
\sigma_{\mathrm{k}}:=\varsigma_t\sqrt{\Delta t}.
\end{aligned}
\label{eq:grpo-v-mean}
\end{equation}
Here $\Delta t=t-r$, $\alpha_t$ and $b_t$ are the Euler--Maruyama
coefficients, $\sigma_{\mathrm{k}}$ is the kernel width, $\varsigma_t$ is the noise scale of the scoring kernel,
and both means use the same detached correction $J=\Jterm{\uold}{t}{s}$
evaluated on the replay trajectory with lower endpoint $s$. The
definition of $\varsigma_t$ that avoids the singularity at $t_0=1$, the trajectory correction and the SDE derivation are
deferred to \Cref{app:grpo-kernel}.

\paragraph{Anchors and weights.}
The tempered (per-element-averaged) scoring ratio and its detached clipping gate are,
explicitly,
\begin{equation}
\begin{aligned}
\bar\rho_{\mathrm{GRPO}}
&:=\exp\!\left[
\frac{\msp{(\mu_{\mathrm{old}}-x_r)^2}
-\msp{(\mu_\theta-x_r)^2}}{2\sigma_{\mathrm{k}}^{2}}
\right],\\
\chi_{\mathrm{GRPO}}
&:=\mathbf 1\!\left[
  \big(A>0\ \wedge\ \bar\rho_{\mathrm{GRPO}}\le1+\varepsilon_{\mathrm{clip}}\big)
  \ \vee\
  \big(A<0\ \wedge\ \bar\rho_{\mathrm{GRPO}}\ge1-\varepsilon_{\mathrm{clip}}\big)
\right].
\end{aligned}
\label{eq:grpo-score-ratio}
\end{equation}
Thus $\chi_{\mathrm{GRPO}}=0$ when $A=0$. The quantity
$\bar\rho_{\mathrm{GRPO}}$ is a surrogate scoring ratio rather than a rollout
importance weight. The anchors and weights are
\begin{equation}
\begin{aligned}
\xz&=\mu_\theta,\qquad
\xreg=\varnothing,\qquad
\xbase=\sg(\mu_\theta),\qquad
\xrwd=x_r,\\
W_{\mathrm{reg}}&=0,\qquad
W_{\mathrm{rwd}}=W=\bar\rho_{\mathrm{GRPO}},\qquad
\omega=\frac{\chi_{\mathrm{GRPO}}\,A}{2\sigma_{\mathrm{k}}^{2}},
\end{aligned}
\label{eq:grpo-decomp}
\end{equation}
so only $\mu_\theta$ is differentiated.
\Cref{app:grpo-kernel,app:grpo-loss} give the tempered interpretation and
show that \eqref{eq:master} reproduces its clipped gradient. An auxiliary
(\texttt{v-aux}) variant is not used because $\vph$ is fitted on endpoint-renoised
states rather than replayed sampler states
(\Cref{app:grpo-noaux}).

\begin{pbox}[label={ex:grpo-code}]{FewStepGRPO: the replayed transition}
    \begin{minted}[linenos, breaklines, mathescape, fontsize=\footnotesize, xleftmargin=2em]{python}
class FewStepGRPO(FewStepRLTrainer):            # replaces Stage 1 and Stage 3(i)
    estimator = ReplayVInd()                    # Eq. ($\ref{eq:cd-replay}$)
    n_it = M_step

    def rollout(self, prompts):                 # Eq. ($\ref{eq:rollout-sde}$)
        return sde_rollout(prompts, u_old, keep_transitions=True)
    def queries(self, batch):                   # replay; one $A$ per member
        for x_t, x_r, t, r in batch.transitions:    # $r$: target level of the step
            s = draw_s(t)                           # $s$ drawn as $r$ in Stage 3(i)
            yield Query(batch, t=t, r=r, s=s, x_t=x_t, x_r=x_r)
    def anchors(self, q):                       # Eqs. ($\ref{eq:grpo-v-mean}$), ($\ref{eq:grpo-decomp}$)
        v_th, v_old, _ = self.estimator.fields(q, r=q.s) # the velocity at $(t,s)$
        a, b, q.tau = em_coeffs(q.t, q.r)       # $\alpha_t,\,b_t,\,\sigma_{\mathrm{k}}$ of Eq. ($\ref{eq:grpo-v-mean}$)
        mu = a * q.x_t - b * v_th               # $\xz=\mu_\theta$, differentiated
        q.mu_old = a * q.x_t - b * v_old        # enters the ratio only
        return Points(x_z=mu, x_reg=None, x_base=sg(mu), x_rwd=q.x_r)
    def weights(self, q, p):
        l_new = mean_sp((p.x_base - p.x_rwd) ** 2)
        l_old = mean_sp((q.mu_old - p.x_rwd) ** 2)
        rho_grpo = exp((l_old - l_new) / (2 * q.tau ** 2)) # $\bar\rho_{\mathrm{GRPO}}$
        chi_grpo = clip_gate(rho_grpo, q.A, clip_range)    # $\chi_{\mathrm{GRPO}}$
        return 0., rho_grpo, chi_grpo * q.A / (2 * q.tau ** 2)
    def kl(self, q):                         # same $(x_t,t,s)$
        return mean_sp((u_theta(q.x_t, q.t, q.s)
                        - u_ref(q.x_t, q.t, q.s)) ** 2)
    \end{minted}
  \end{pbox}

\subsection{FewStepAWM: MeanFlowAWM and DMDAWM}\label{sec:fewstep-awm}

FewStepAWM adapts \eqref{eq:awmloss} as one clipped reward-augmentation term
on the matching residual of the model. It has no explicit self-regression term,
and in MeanFlowAWM the reward displacement is measured from the detached current prediction.

\paragraph{Rollout and prediction.}
Stage~1 uses the deterministic rollout, $n_{\mathrm{it}}=n_{\mathrm{ts}}$,
and every inner iteration draws a fresh $(t,r,\epsilon)$, re-noises the
rewarded endpoint and computes the v-prediction \eqref{eq:P-v},
\begin{equation}
\xt=\AddNoise(\xh,t),
\qquad
\xz=\xv=\Xmap{\xt}{\uth(\xt,t,r)+\Jterm{\uold}{t}{r}}{0},
\label{eq:awm-pred}
\end{equation}
so that, by \eqref{eq:xidentity},
\begin{equation}
\xz-\xh
=-t\Big(\uth-\big(\vcond-\Jterm{\uold}{t}{r}\big)\Big),
\qquad \vcond=\epsilon-\xh.
\label{eq:awm-residual}
\end{equation}
Thus $\uth$ regresses onto $\vcond-J$, the MeanFlow target
\eqref{eq:mfloss} with the conditional velocity in place of the teacher.

\paragraph{MeanFlowAWM anchors and weights.} The method uses the rollout endpoint as
the rewarded anchor, no explicit self-regression, and the current prediction as baseline:
$\xrwd=\xh$, $\xreg=\varnothing$, $\xbase=\sg(\xz)$. Besides the shared $J$ in $\xz$, the old policy
enters only through the weights, via the shared-$J$ prediction of
Stage~3\,\emph{(iii)},
\begin{equation}
\xind=\Xmap{\xt}{\uold(\xt,t,r)+\Jterm{\uold}{t}{r}}{0},
\qquad
\xz-\xind=-t\big(\uth-\uold\big).
\label{eq:awm-anchor}
\end{equation}
The weights reuse the quadratic-exponential form of
\S\ref{sec:fewstep-grpo}, but not its probabilistic interpretation:
\begin{equation}
\begin{aligned}
\rho_{\mathrm{AWM}}
&:=\exp\!\left[
\frac{\msp{(\xind-\xh)^2}-\msp{(\xz-\xh)^2}}
{2\tau_{\mathrm{AWM}}^{2}}\right],\\
\chi_{\mathrm{AWM}}
&:=\mathbf 1\!\left[
  \big(A>0\ \wedge\ \rho_{\mathrm{AWM}}\le1+\varepsilon_{\mathrm{clip}}\big)
  \ \vee\
  \big(A<0\ \wedge\ \rho_{\mathrm{AWM}}\ge1-\varepsilon_{\mathrm{clip}}\big)
\right].
\end{aligned}
\label{eq:awm-statistic}
\end{equation}
Thus $\chi_{\mathrm{AWM}}=0$ when $A=0$. The implemented residual scale is
$\tau_{\mathrm{AWM}}=\max\{t,\tau_{\min}\}$; unlike in FewStepGRPO, no sampler
kernel determines this scale. The choice $\tau_{\mathrm{AWM}}=t$ would cancel
the factor $t^2$ in both residuals; when the floor is active, the factor
$t^2/\tau_{\mathrm{AWM}}^2$ remains:
\begin{equation}
\frac{\partial\mathcal L}{\partial\uth}
=\frac{\rho_{\mathrm{AWM}}\,\chi_{\mathrm{AWM}}\,A}{\Nel}
\frac{t^2}{\tau_{\mathrm{AWM}}^2}
\Big(\uth-\big(\vcond-\Jterm{\uold}{t}{r}\big)\Big)
\label{eq:awm-grad}
\end{equation}
and its factor $t^2/\tau_{\mathrm{AWM}}^2$ equals one only above the floor. The resulting quantities are
\begin{equation}
\xreg=\varnothing,
\quad
\xbase=\sg(\xz),
\quad
\xrwd=\xh,
\quad
W_{\mathrm{reg}}=0,\quad
W_{\mathrm{rwd}}=W=\rho_{\mathrm{AWM}},
\quad
\omega=\frac{\chi_{\mathrm{AWM}}\,A}{2\tau_{\mathrm{AWM}}^{2}}.
\label{eq:awm-fewstep-decomp}
\end{equation}
Relative to \eqref{eq:awm-decomp} with $w(t)=1$, the step has an extra factor
$\tfrac12$, which comes from the Gaussian form of the statistic and is absorbed
by the learning rate. $\rho_{\mathrm{AWM}}$ is not an importance weight, since
$(\xt,\xh)$ is not a transition of the rollout; through the gate
$\chi_{\mathrm{AWM}}$, it limits one-sided changes of the matching residual
relative to $\uold$. We call this variant \emph{MeanFlowAWM}.

\paragraph{DMDAWM variant.}
The reported \texttt{v-aux} variant, \emph{DMDAWM}, drops $J$ in the current
and old-policy predictions (the $J$-free $(t,r)$ query and $\xoldur$ of \S\ref{sec:fewstep}) and replaces the reward baseline with the prediction of the auxiliary network:
\begin{equation}
\begin{gathered}
\xz=\Xmap{\xt}{\uth(\xt,t,r)}{0},\qquad
\xoldur
=\Xmap{\xt}{\uold(\xt,t,r)}{0},\qquad
\xbase=\xaux,\\
\rho_{\mathrm{AWM}}=\exp\!\left[
\frac{\msp{(\xoldur-\xh)^2}
-\msp{(\xz-\xh)^2}}{2\tau_{\mathrm{AWM}}^2}\right],
\\
\frac{\partial\mathcal L}{\partial\uth}
=\frac{\rho_{\mathrm{AWM}}\,\chi_{\mathrm{AWM}}\,A}{\Nel}
\frac{t^2}{\tau_{\mathrm{AWM}}^2}
\big(\vph-\vcond\big).
\end{gathered}
\label{eq:dmdawm}
\end{equation}
Above the floor the factor $t^2/\tau_{\mathrm{AWM}}^2$ equals one; descent moves $\uth$ along
$+A(\vcond-\vph)$, with $\vph$ as a learned per-sample baseline. If a detached
$J$ were kept in the first two predictions, it would change only the scalar
ratio $\rho_{\mathrm{AWM}}$ and the clipping gate, but not the gradient direction. The
reported method omits it, so it requires no directional-difference evaluation of $J$.

\paragraph{Decoupled loss.}
Both variants use \eqref{eq:master} with
$W_{\mathrm{reg}}=0$, so only $\mathcal L_{\mathrm{rwd}}$ remains.
MeanFlowAWM measures the reward displacement from $\sg(\xz)$; DMDAWM measures
it from the auxiliary prediction $\xaux$.
In code, FewStepAWM implements \mintinline{python}{anchors} and
\mintinline{python}{weights}, and MeanFlowAWM and DMDAWM differ only in their
estimator (\Cref{ex:awm-code}).

\begin{pbox}[label={ex:awm-code}]{FewStepAWM: anchors and weights}
    \begin{minted}[linenos, breaklines, mathescape, fontsize=\footnotesize, xleftmargin=2em]{python}
class FewStepAWM(FewStepRLTrainer):             # no self-regression; $\xrwd=\xh$
    def anchors(self, q):
        v_th, v_old, v_b = self.estimator.fields(q)
        x_z = X(q.x_t, v_th, 0)                 # Eq. ($\ref{eq:awm-pred}$)
        q.x_old = X(q.x_t, v_old, 0)            # Eq. ($\ref{eq:awm-anchor}$): statistic only
        # baseline: the current prediction, or the prediction of $\vph$, Eq. ($\ref{eq:dmdawm}$)
        x_base = sg(x_z) if v_b is None else X(q.x_t, v_b, 0)
        return Points(x_z=x_z, x_reg=None, x_base=x_base, x_rwd=q.x_hat)
    def weights(self, q, p):                    # Eq. ($\ref{eq:awm-fewstep-decomp}$)
        tau = clamp(q.t, min=TAU_MIN)           # $\tau_{\mathrm{AWM}}=\max(t,\tau_{\min})$
        l_new = mean_sp((sg(p.x_z) - p.x_rwd) ** 2)
        l_old = mean_sp((q.x_old - p.x_rwd) ** 2)
        rho_awm = exp((l_old - l_new) / (2 * tau ** 2)) # $\rho_{\mathrm{AWM}}$
        chi_awm = clip_gate(rho_awm, q.A, clip_range)   # $\chi_{\mathrm{AWM}}$
        return 0., rho_awm, chi_awm * q.A / (2 * tau ** 2)

class MeanFlowAWM(FewStepAWM): estimator = VInd()  # shared $J$
class DMDAWM(FewStepAWM):      estimator = VAux()  # $\vph$ baseline, no $J$
    \end{minted}
  \end{pbox}

\subsection{FewStepNFT: MeanFlowNFT and DMDNFT}\label{sec:fewstep-nft}

FewStepNFT maps the two implicit branches of DiffusionNFT to the two terms of
\eqref{eq:master}. It contains both a self-regression term and a signed
reward-augmentation term, with $\xrwd=\xh$ in both variants.

\paragraph{Anchors, predictions and weights.} For the two variants, the quantities of
\eqref{eq:nft-decomp} are listed in \Cref{tab:nft-anchors}.

\begin{table}[tbp]
\centering
\renewcommand{\arraystretch}{1.25}
\small\setlength{\tabcolsep}{4pt}
\begin{tabular}{@{}lll@{}}
\toprule
anchors   & $\xz$ & $\xreg$, $\xbase$  \\
\midrule
\texttt{v-ind}   & $\xv=\Xmap{\xt}{\uth(\xt,t,r)+J}{0}$ & $\xreg=\xbase=\xind$ \\
\texttt{v-aux}   & $\Xmap{\xt}{\uth(\xt,t,r)}{0}$  & $\xreg=\xoldur$, $\xbase=\xaux$  \\
\bottomrule
\end{tabular}
\caption{Prediction and anchors of the two FewStepNFT variants; $J=\Jterm{\uold}{t}{r}$.}
\label{tab:nft-anchors}
\end{table}
The theoretical weights of \eqref{eq:nft-decomp} are modified in the
implementation, following its official implementation in \citep{zheng2026diffusionnft}.
The derivation of \S\ref{sec:grad} gives
$W_{\mathrm{reg}}=W_{\mathrm{rwd}}=\beta^{2}/t^{2}$ and
$\omega=\clip(A,-1,1)/\beta$, i.e.\ one fixed per-$t$ scale for both branches,
with $\hat r$ from \eqref{eq:rhat-prelim} and $A$ normalised within the group.
The implementation keeps the two branches but changes the following: the per-$t$
scale is replaced by one adaptive normaliser per branch, $\gamma_\pm$, the mean
absolute residual of that branch; $\hat r$ is the compressed weight
\eqref{eq:rhat} with clip range $c_{\mathrm{adv}}$; and $A$ is the
advantage \eqref{eq:adv}. The weights are then\begin{equation}
\gamma_{\pm}=\max\Big\{\msp{\big|\xbase\pm\beta\big(\sg(\xz)-\xreg\big)-\xh\big|},
\ \varepsilon_\gamma\Big\},
\quad
a_{+}=\frac{\hat r}{\gamma_{+}},
\quad
a_{-}=\frac{1-\hat r}{\gamma_{-}},
\label{eq:nft-gamma}
\end{equation}
\begin{equation}
W_{\mathrm{reg}}=W_{\mathrm{rwd}}=W
=c_{\mathrm{adv}}\beta^{2}(a_++a_-),
\qquad
\omega=\frac{a_+-a_-}{\beta\,(a_++a_-)}.
\label{eq:nft-weights}
\end{equation}
If $\gamma_+=\gamma_-=t^{2}$, $c_{\mathrm{adv}}=1$ and $A$ is the
group-normalised advantage \eqref{eq:adv-prelim}, they reduce to
\eqref{eq:nft-decomp}, with $W=\beta^{2}/t^{2}$ and
$\omega=\clip(A,-1,1)/\beta$. In the implemented weighting, $\hat r$ enters
through $a_\pm$ as a mixing weight, $c_{\mathrm{adv}}$ scales both loss terms
to undo the compression of the advantage by $1/(2c_{\mathrm{adv}})$ in
\eqref{eq:rhat}, so that the effective step does not shrink when the clip
range is widened (\Cref{app:grpo-loss}),
and $\omega$ is defined directly. The \texttt{v-ind} estimator gives
\emph{MeanFlowNFT}, and the \texttt{v-aux} estimator gives \emph{DMDNFT}.

\paragraph{Decoupled loss.}
Both variants minimise
$\mathcal L_{\mathrm{reg}}+\mathcal L_{\mathrm{rwd}}$ with the common weight
$W$, the rollout endpoint as $\xrwd$, and \Cref{tab:nft-anchors} defining
$\xreg$ and $\xbase$.

\begin{pbox}[label={ex:nft-code}]{FewStepNFT: anchors and weights}
    \begin{minted}[linenos, breaklines, mathescape, fontsize=\footnotesize, xleftmargin=2em]{python}
class FewStepNFT(FewStepRLTrainer):             # both squares on; $\xrwd=\xh$
    def anchors(self, q):
        v_th, v_old, v_b = self.estimator.fields(q)
        x_z, x_reg = X(q.x_t, v_th, 0), X(q.x_t, v_old, 0)   # same $(t,r)$, same $J$
        # baseline: aliased to $\xreg$, or the prediction of $\vph$, Eq. ($\ref{eq:vpred-nft-phi}$)
        x_base = x_reg if v_b is None else X(q.x_t, v_b, 0)
        return Points(x_z=x_z, x_reg=x_reg, x_base=x_base, x_rwd=q.x_hat)
    def branch_weights(self, q, p, x_pos, x_neg):    # Eq. ($\ref{eq:nft-gamma}$)
        dev = sg(p.x_z) - p.x_reg               # direction from $\xreg$, centre $\xbase$
        g_pos = clamp(mean_sp(abs(p.x_base + beta * dev - x_pos)), min=eps_gamma)
        g_neg = clamp(mean_sp(abs(p.x_base - beta * dev - x_neg)), min=eps_gamma)
        return q.r_hat / g_pos, (1 - q.r_hat) / g_neg    # $\hat r$ enters ONLY here
    def weights(self, q, p):                    # Eq. ($\ref{eq:nft-weights}$), not ($\ref{eq:nft-decomp}$)
        a_pos, a_neg = self.branch_weights(q, p, p.x_rwd, p.x_rwd)
        W = c_adv * beta ** 2 * (a_pos + a_neg)
        return W, W, (a_pos - a_neg) / (beta * (a_pos + a_neg))

class MeanFlowNFT(FewStepNFT): estimator = VInd()  # $\xz=\xv$, $\xreg=\xbase=\xind$
class DMDNFT(FewStepNFT):      estimator = VAux()  # $\xbase=\xaux\neq\xreg$
    \end{minted}
  \end{pbox}

\subsection{FewStepOPSD: Extended Reward Augmentation and ReFL}\label{sec:FewStepOPSD}

FewStepOPSD belongs to the family of reward-gradient-based methods
\citep{xu2023imagereward,clark2024draft,wang2026rewardbackprop} and is related
to gradient-guided distribution matching \citep{dong2026guidingdmd}.
It implements \refex{ex:opsd}\,\emph{(iii)} as a FewStepNFT trainer whose rewarded
points are constructed from the reward gradient instead of sampled; it
therefore uses the normalisers of \eqref{eq:nft-gamma}, and its configuration
with a single ascent step is the ReFL proxy of \refex{ex:refl}. It modifies
more stages than FewStepNFT. Stage~1 depends on the reward, the rewarded
anchors are constructed points instead of rollout samples, and Stage~4
overrides \mintinline{python}{loss}. Stage~2 is unchanged, and $\hat r$ enters
only through the weights.

\paragraph{Rollout.} FewStepOPSD uses the reward-coupled rollout
\eqref{eq:rollout-dep}. It first draws the i.i.d.\ base samples with
\eqref{eq:rollout-indep}. The reward-coupled construction then applies
\eqref{eq:opsd-target}, taking $M_{\mathrm{tgt}}$ normalised reward-gradient steps in
each direction from the $r=0$ prediction $\xoldu$, with the displacement
transferred to the anchor by \eqref{eq:opsd-shift}, giving $\tilde x_\pm$.
The points $\tilde x_\pm$ are detached, so the computation graph of the reward
ends here and the policy update does not back-propagate through the decoder or
the reward model; the reward must be differentiable.

\paragraph{Anchors and weights.}
\emph{(i) Query.} Each member is queried at a single level $t$, drawn from the
law given in \S\ref{sec:exp-setup}; hence
$n_{\mathrm{it}}=1$. As in \refex{ex:opsd} and \refex{ex:refl}, the query
state is the re-noised endpoint $\xt=\AddNoise(\xh,t)$, so the anchor is
evaluated at the same state as the current prediction. The drawn $r$ is stored
and reused at fitting time, so the current prediction and the anchor are
queried at the same $(t,r)$.

\emph{(ii) Anchors.} The u-form uses
$\xreg=\xbase=\xoldu$ and the v-form uses
$\xreg=\xbase=\xind$, \eqref{eq:opsd-v-decomp}; in both, the rewarded anchors
are $\tilde x_\pm$. For the v-form, the gradient steps start from the $r=0$
prediction $\xoldu$, which does not depend on $J$, and only the resulting
displacement is added to the anchor $\xind$ by \eqref{eq:opsd-shift}.

\emph{(iii) Instantaneous velocity.} The v-form uses the shared $J$ of
Stage~3\,\emph{(iii)}. Because both rewarded points are constructed from the anchor,
$J$ cancels in every difference in the loss, as in \eqref{eq:opsd-v-lreg} and
\eqref{eq:opsd-v-lrwd}. 
The v-form reported in \S\ref{sec:exp} drops $J$ and evaluates no directional
difference (\eqref{eq:cd}); in code its estimator is
\mintinline{python}{VInd(cd=False)}.

\emph{(iv) Weights.} With \eqref{eq:nft-gamma} evaluated at $\tilde x_\pm$,
\begin{equation}
    \begin{aligned}
        W_{\mathrm{reg}}=W_{\mathrm{rwd}}=W
& =c_{\mathrm{adv}}\beta^{2}(a_++a_-),
\\
\omega_{1}=\frac{a_{+}}{\beta\,(a_++a_-)},
\quad
\omega_{2}&=-\frac{a_{-}}{\beta\,(a_++a_-)},
\qquad
|\omega_1|+|\omega_2|=\frac1\beta ,
    \end{aligned}
\label{eq:opsd-weights}
\end{equation}
so the sum of coefficient magnitudes is fixed at $1/\beta$ and $\hat r$
redistributes it between the two terms. It can still change the common weight
$W$ when $\gamma_+\ne\gamma_-$. The factor $c_{\mathrm{adv}}$ scales the common
weight but not the split between the branches.

\paragraph{Decoupled loss.} The loss \eqref{eq:opsd-decoupled} consists of one
self-regression term and two reward-augmentation terms, which share the
baseline and differ in the rewarded point $\tilde x_{\mathrm{rwd},k}$ and the
coefficient $\omega_k$; the signs of $\omega_k$ are fixed by construction and
do not depend on $A$.

\begin{pbox}[label={ex:opsd-code}]{FewStepOPSD: reward-coupled construction and loss}
    \begin{minted}[linenos, breaklines, mathescape, fontsize=\footnotesize, xleftmargin=2em]{python}
def tr_steps(x0, c, direction, M_tgt):          # Eq. ($\ref{eq:opsd-target}$)
    radius = kappa * norm(x0); h = eta * radius / M_tgt
    x = x0
    for _ in range(M_tgt):
        grad_R = grad(reward(decode(x), c), x)  # through reward model AND decoder
        x = x + direction * h * grad_R / (norm(grad_R) + eps_g)
        d = x - x0                              # trust region on the displacement
        x = x0 + d * clamp(radius / (norm(d) + eps_g), max=1.)
    return sg(x)

class FewStepOPSD(FewStepNFT):                  # NFT onto constructed points
    estimator = VInd(cd=False)                  # $J$ cancels in the loss
    M_tgt, directions = 2, (+1, -1)
    def rollout(self, prompts):                 # Eq. ($\ref{eq:rollout-dep}$)
        b = super().rollout(prompts)
        b.t, b.r = draw_t_logitnormal(b), draw_r(b)          # stored, one per member
        b.x_t = add_noise(b.x_hat, b.t)
        x_old = X(b.x_t, u_old(b.x_t, b.t, b.r), 0)          # the anchor
        shift = X(b.x_t, u_old(b.x_t, b.t, 0.), 0) - x_old   # Eq. ($\ref{eq:opsd-shift}$)
        b.x_pm = [tr_steps(x_old + shift, b.c, d, self.M_tgt) - shift
                  for d in self.directions]                  # detached
        return b
    def queries(self, batch):                   # $n_{\mathrm{it}}=1$, stored $(t,r)$
        yield Query(batch, t=batch.t, r=batch.r, x_t=batch.x_t)
    def anchors(self, q):                       # Eq. ($\ref{eq:opsd-v-decomp}$)
        p = super().anchors(q)                  # $\xreg=\xbase$: old-policy prediction
        return Points(x_z=p.x_z, x_reg=p.x_reg, x_base=p.x_base, x_rwd=q.x_pm)
    def weights(self, q, p):                    # Eq. ($\ref{eq:opsd-weights}$)
        a_pos, a_neg = self.branch_weights(q, p, *p.x_rwd)   # $\gamma_\pm$ at $\tilde x_\pm$
        W = c_adv * beta ** 2 * (a_pos + a_neg)
        s = beta * (a_pos + a_neg)
        return W, W, (a_pos / s, -a_neg / s)    # $|\omega_1|+|\omega_2|=1/\beta$
    def loss(self, p, W_reg, W_rwd, omega):     # override, Eq. ($\ref{eq:opsd-decoupled}$)
        (x_pos, x_neg), (om_1, om_2) = p.x_rwd, omega
        l_reg = self_regression(p.x_z, p.x_reg, W_reg)
        l_pos = reward_augmentation(p.x_z, x_pos, p.x_base, om_1, W_rwd)  # positive branch
        l_neg = reward_augmentation(p.x_z, x_neg, p.x_base, om_2, W_rwd)  # negative branch
        return l_reg + l_pos + l_neg
    \end{minted}
  \end{pbox}

\paragraph{ReFL.} The ReFL configuration of
\eqref{eq:opsd-decoupled}, identified in \refex{ex:refl}, reuses the stages
above, with the changes below. Stage~1 takes a single ascent step, $M_{\mathrm{tgt}}=1$, without a
descent point. As for FewStepOPSD, the step starts at the $r=0$ prediction
and its displacement is added back by \eqref{eq:opsd-shift}, so $\nabla R$ is
evaluated at $\xoldu$ as specified by \eqref{eq:refl-anchor}.
Stage~2 is not used, because ReFL has no advantage; $\omega_1>0$ is fixed by
construction and the implementation sets $\hat r\equiv1$ in
\eqref{eq:opsd-weights}, giving $\omega_1=1/\beta$ and
$W_{\mathrm{rwd}}=c_{\mathrm{adv}}\beta^{2}/\gamma_+$, detached and positive as \refex{ex:refl}
allows. The ReFL override keeps the factor $c_{\mathrm{adv}}$ of
\eqref{eq:opsd-weights}. Stage~3 keeps the same
anchors. Stage~4 keeps only $\mathcal L_{\mathrm{rwd},1}$ with
$W_{\mathrm{reg}}=0$, i.e.\ the loss of \Cref{ex:opsd-code} reduces to its
ascent term (\Cref{ex:refl-code}). The resulting gradient is that of
\eqref{eq:refl-fewstep}, the few-step form of \eqref{eq:refl-proxy}.

\begin{pbox}[label={ex:refl-code}]{ReFL: the proxy configuration of FewStepOPSD}
    \begin{minted}[linenos, breaklines, mathescape, fontsize=\footnotesize, xleftmargin=2em]{python}
class ReFL(FewStepOPSD):                        # proxy configuration of Eq. ($\ref{eq:opsd-decoupled}$)
    M_tgt, directions = 1, (+1,)                # one ascent step, no descent point
    def advantage(self, R, c):                  # $\hat r\equiv1$: no Stage 2
        return None, ones_like(R)
    def weights(self, q, p):                    # $\omega_1=1/\beta$, $W_{\mathrm{rwd}}=c_{\mathrm{adv}}\beta^2/\gamma_+$
        a_pos, _ = self.branch_weights(q, p, p.x_rwd[0], p.x_rwd[0])
        return 0., c_adv * beta ** 2 * a_pos, (1. / beta,)
    def loss(self, p, W_reg, W_rwd, omega):     # $W_{\mathrm{reg}}=0$: ascent term only
        return reward_augmentation(p.x_z, p.x_rwd[0], p.x_base, omega[0], W_rwd)
    \end{minted}
  \end{pbox}

\subsection{Cost of an Outer Iteration}\label{sec:impl-cost}

\Cref{tab:cost} lists the network evaluations per inner iteration for each
network. The rollout of Stage~1 is amortised over one outer iteration and is
not counted here. Each entry is determined by the family and the estimator of
\S\ref{sec:impl-base}: $n_{\mathrm{it}}$ by \mintinline{python}{queries},
$c_J$ by the estimator ($2$ for the directional-difference estimate of \texttt{v-ind}, and
$0$ otherwise), and $c_\phi$ and the auxiliary part of $c_{\mathrm{extra}}$ by
\texttt{v-aux}; the reward-coupled construction of FewStepOPSD and ReFL is
counted in $c_{\mathrm{extra}}$. The counters $c_\theta$, $c_{\mathrm{old}}$, $c_J$, $c_\phi$,
$c_{\mathrm{ref}}$, $c_{\mathrm{extra}}$, $C$ and $C_{\mathrm{rollout}}$ are
defined in the caption and listed in \Cref{app:notation}.

\begin{table}[ht]
    \centering
    \renewcommand{\arraystretch}{1.2}
    \resizebox{0.85\linewidth}{!}{%
    \begin{tabular}{@{}lccccccl@{}}
    \toprule
    Trainers & $n_{\mathrm{it}}$ & $c_\theta$ & $c_{\mathrm{old}}$
    & $c_J$ & $c_\phi$ & $c_{\mathrm{ref}}$ &  $c_{\mathrm{extra}}$ \\
    \midrule
    \S\ref{sec:fewstep-grpo}, FewStepGRPO         & $M_{\mathrm{step}}$ & 1 & 1 & 2 & --- & 1 & ---\\
    \S\ref{sec:fewstep-awm}, MeanFlowAWM    & $n_{\mathrm{ts}}$ & 1 & 1 & 2 & --- & 1 & ---\\
    \S\ref{sec:fewstep-awm}, DMDAWM   & $n_{\mathrm{ts}}$ & 1 & 1 & 0 & 1 & 1 & $n_{\mathrm{fm}}$ forwards and backwards of $\vph$ \\
    \S\ref{sec:fewstep-nft}, MeanFlowNFT   & $n_{\mathrm{ts}}$ & 1 & 1 & 2 & --- & 1 & ---\\
    \S\ref{sec:fewstep-nft}, DMDNFT    & $n_{\mathrm{ts}}$ & 1 & 1 & 0 & 1 & 1 & $n_{\mathrm{fm}}$ forwards and backwards of $\vph$ \\
    \S\ref{sec:FewStepOPSD}, FewStepOPSD    & 1 & 1 & 1 & $0^{\dagger}$ & --- & 1 & $2M_{\mathrm{tgt}}$ reward backwards, 2 no-grad forwards of $\uold$ \\
    \refex{ex:refl}, ReFL                   & 1 & 1 & 1 & $0^{\dagger}$ & --- & 1 & 1 reward backward, 2 no-grad forwards of $\uold$\\
    \bottomrule
    \end{tabular}}
    \caption{Network evaluations per inner iteration.
    $n_{\mathrm{it}}$: number of inner iterations per outer iteration, each
    querying every member once;
    $c_\theta$: forward--backward passes of $\uth$;
    $c_{\mathrm{old}}$: no-grad forwards of $\uold$ for the old-policy anchor;
    $c_J$: maximum extra no-grad forwards of $\uold$ for the directional-difference
    estimate of $J$ (\eqref{eq:cd} or \eqref{eq:cd-replay}; $0$ where the
    estimator uses no $J$;
    $^{\dagger}$with $J$ dropped because it cancels, which leaves the loss and its gradient unchanged by \Cref{app:opsd});
    $c_\phi$: forwards with no gradient of the auxiliary network $\vph$;
    $c_{\mathrm{ref}}$: forwards of the frozen reference for the velocity penalty;
    $c_{\mathrm{extra}}$: extra cost per outer iteration, where $n_{\mathrm{fm}}$
    is the number of flow-matching steps that fit $\vph$.
    The cost of one outer iteration is thus
    $C=C_{\mathrm{rollout}}+c_{\mathrm{extra}}
    +n_{\mathrm{it}}\,(c_\theta+c_{\mathrm{old}}+c_J+c_\phi+c_{\mathrm{ref}})$,
    where $C_{\mathrm{rollout}}$ is the plain rollout of Stage~1.}
    \label{tab:cost}
\end{table}

\section{Experiment}\label{sec:exp}

Having established a general view of these methods, we use experiments to
answer three questions. Every trainer of
\S\S\ref{sec:fewstep-grpo}--\ref{sec:FewStepOPSD} implements one of these
methods, minimises the same loss form, and differs mainly in its anchors and
weights. \emph{(i)}~Which of these choices, in particular the reward signal and the
estimator of the instantaneous velocity, determine how effectively and how
efficiently a few-step generator is fine-tuned, and how strongly it exploits the
reward? \emph{(ii)}~Do these
conclusions carry over to a different modality, or do they show limitations in
other scenarios? \emph{(iii)}~Does a generator fine-tuned at a few steps improve when
it is sampled with more steps, without retraining?

\subsection{Setup on Image Generation}\label{sec:exp-setup}

\begin{table}[tbp]
    \centering
    \footnotesize
    \setlength{\tabcolsep}{3.5pt}
    \renewcommand{\arraystretch}{1.2}
    \resizebox{0.9\textwidth}{!}{%
    \begin{tabular}{@{}clllll@{}}
    \toprule
    class & method & reward signal & query of $\uth$ & $\xbase$ & instantaneous field \\
    \midrule
    \multirow{5}{*}{Group-adv.}
     & FewStepGRPO, \S\ref{sec:fewstep-grpo} & $A$ & $(t,s)$, replayed & $\sg(\mu_\theta)$ & Eq.~(\ref{eq:cd-replay}) \\
     & MeanFlowAWM, \S\ref{sec:fewstep-awm} & $A$ & $(t,r)$ & $\sg(\xz)$ & Eq.~(\ref{eq:cd}) \\
     & DMDAWM, \S\ref{sec:fewstep-awm} & $A$ & $(t,r)$ & $\xaux$ & $\vph$ baseline; $J=0$\\
     & MeanFlowNFT, \S\ref{sec:fewstep-nft}  & $\hat r$ & $(t,r)$ & $\xind$ & Eq.~(\ref{eq:cd}) \\
     & DMDNFT, \S\ref{sec:fewstep-nft} & $\hat r$ & $(t,r)$ & $\xaux$ & $\vph$ \\
    \midrule
    \multirow{2}{*}{Reward-grad.}
     & FewStepOPSD, \S \ref{sec:FewStepOPSD} & $\nabla R$, $\hat r$ & $(t,r)$ & $\xind$ & $J=0$ (App.~\ref{app:opsd}) \\
     & ReFL, \S \ref{sec:FewStepOPSD}& $\nabla R$ & $(t,r)$ & $\xind$ & $J=0$ \\
    \bottomrule
    \end{tabular}}
    \caption{The methods compared; ReFL is \refex{ex:refl}. Class: the first five
    methods compute scalar rewards for the members of each prompt group and form
    per-sample advantages using the per-prompt mean and current
    global-batch standard deviation of \eqref{eq:adv}, so they are
    group-advantage-based. The final two obtain
    their clean-sample update direction from $\nabla R$ and are
    reward-gradient-based; FewStepOPSD still uses $\hat r$ only to split the
    ascent/descent weights. Reward signal: the advantage as a signed multiplier
    ($A$), as a mixing weight ($\hat r$, Stage~2), or the reward gradient
    ($\nabla R$, \eqref{eq:opsd-target}). Every
    method except FewStepGRPO uses the deterministic sampler of
    \eqref{eq:rollout-op} in Stage~1. \Cref{app:opsd} establishes when dropping
    $J$ (last column) is admissible.}
    \label{tab:arms}
    \end{table}
        
\paragraph{Generator and sampler.} The policy is the average-velocity network
$\uth(\xt,t,r)$ of \S\ref{sec:prelim}, initialised from the pretrained
SD3.5-Medium backbone \citep{stabilityai2024sd35,esser2024sd3} and distilled
with AnyFlow \citep{gu2026anyflow}, which first trains with the MeanFlow
identity \eqref{eq:mfid} and then applies on-policy distribution matching
\eqref{eq:dmdloss} against the frozen teacher.
Reinforcement learning trains a LoRA \citep{hu2022lora} of rank $32$ (scaling $64$) on the
attention projections, at resolution $512\times512$. We set
$M_{\mathrm{step}}=4$ steps on the grid of levels
$\{1,0.9,0.75,0.5,0\}$, without classifier-free guidance
\citep{ho2021classifierfree}, with the deterministic
sampler of \eqref{eq:rollout-op} for every trainer except FewStepGRPO, which
requires the stochastic sampler of \eqref{eq:rollout-sde}. Evaluation always
uses the deterministic sampler on the same grid.

\paragraph{Optimisation.} In Stage~1 and Stage~2, each outer iteration draws $48$ prompts with $K=24$
members each, i.e.\ $1152$ rollouts, on $8$ NVIDIA H20 GPUs. All methods use
AdamW \citep{loshchilov2019adamw} with learning rate $3\times10^{-6}$, weight
decay $10^{-4}$ and gradient-norm clipping at $1$; an old policy $\uold$
given by an EMA of $\uth$, refreshed once per outer iteration after the
policy update with the decay schedule of the official DiffusionNFT
implementation (\Cref{app:hparams}); for the NFT/OPSD families, $\beta=0.1$ and $c_{\mathrm{adv}}=5$
in \eqref{eq:rhat}; and the KL-u penalty against the fixed AnyFlow initial
policy, using \eqref{eq:kl-uref} with
weight $10^{-4}$. An ablation with the penalty \eqref{eq:kl-vref}, labelled
KL-v, on MeanFlowNFT is given in \Cref{app:klablation}.
Stage~3\,\emph{(i)} draws $t$ and $r$ as the maximum and the minimum of two
uniforms before the time
shift of the SD3 scheduler, $t\mapsto 3t/(1+2t)$ (shift $3.0$, the same
shifted grid as in sampling),
sets half of each batch to $r=t$ and a quarter to $r=0$, and
uses $n_{\mathrm{ts}}=4$ queries per member and outer iteration. The floors
are $\tau_{\min}=10^{-2}$, $\varepsilon_\gamma=10^{-5}$,
$\varepsilon_g=10^{-12}$ and $\varepsilon_{\mathrm{std}}=10^{-4}$, and
$\varepsilon_{\mathrm{cd}}=5\times10^{-3}$ (\Cref{app:hparams}).

\paragraph{Rewards.} We use two reward settings to test the methods.
The first is the OCR setting, where the reward is the OCR accuracy of the
text rendered in the image with the OCR prompts. It has no gradient with respect to the clean
sample, so FewStepOPSD and ReFL, which require $\nabla R$, do not apply. In the
mixed-reward setting the reward is the unweighted sum of CLIPScore, HPSv2 and
PickScore on the PickScore prompt set
\citep{hessel2021clipscore,wu2023hpsv2,kirstain2023pickapic}.

\paragraph{Specific settings.} For the group-advantage-based methods,
$n_{\mathrm{it}}=4$, with $n_{\mathrm{it}}=M_{\mathrm{step}}$ for FewStepGRPO
and $n_{\mathrm{it}}=n_{\mathrm{ts}}$ for the others.
 FewStepGRPO retains the states of
\eqref{eq:rollout-sde} and uses the base noise scale $\varsigma=0.25$ of
$\varsigma_t$ in \eqref{eq:grpo-v-mean} (\eqref{eq:grpo-varsigma}),
with clipping radius $\varepsilon_{\mathrm{clip}}=4\times10^{-3}$ (OCR) or
$3\times10^{-3}$ (mixed reward) in $\chi_{\mathrm{GRPO}}$;
MeanFlowAWM and DMDAWM use radius $10^{-2}$ (OCR) or $2\times10^{-3}$
(mixed reward) in $\chi_{\mathrm{AWM}}$. For DMDAWM and DMDNFT, the auxiliary
velocity $\vph$ is fitted by \eqref{eq:auxfit} with learning rate $10^{-5}$ and
$n_{\mathrm{fm}}=5$ steps per outer iteration. FewStepOPSD uses $\kappa=0.1$, $M_{\mathrm{tgt}}=2$
and $\eta=1$ in \eqref{eq:opsd-target}, with $\nabla R$ of the full training
reward, drawing
$t$ from a logit-normal law with median $0.278$ (5--95\% range
$[0.125,0.508]$), i.e., the small noise levels of \refex{ex:opsd}, and $r$ as in Stage~3\,\emph{(i)}. ReFL is \refex{ex:refl} with $M_{\mathrm{tgt}}=1$,
$\hat r\equiv1$ and $\nabla R$ of the full training reward; for a fair
comparison, $\uold$ is also the EMA of $\uth$.
A summary of the methods is given in \Cref{tab:arms}.

\paragraph{Baselines.} In the mixed-reward setting we additionally
run the original objectives on the pre-RL few-step model without the
adaptation of \S\ref{sec:fewstep}, i.e., with the average velocity
$\uth(\xt,t,r)$, $r$ drawn as in Stage~3\,\emph{(i)} ($s$ for FlowGRPO), substituted for the instantaneous velocity wherever the
original objective requires one. For DiffusionNFT and AWM, it is
\eqref{eq:upred} with $\uth(\xt,t,r)$ in place of $\uth(\xt,t,0)$. For FlowGRPO, this substitution is not theoretically solid. We denote these baselines by the names
of the original methods, FlowGRPO, AWM and DiffusionNFT, and compare them with
FewStepGRPO, MeanFlowAWM and MeanFlowNFT, respectively, while keeping the other setting the same. In addition, by \Cref{app:opsd}, the detached
correction $J$ cancels from its v-form loss, so the u-form in the FewStepOPSD family is the special case
of the v-form in which $r$ is fixed to $0$ instead of being drawn at every
query, so we did not compare them.

\paragraph{Evaluation.} Three sources are used, with the conventions of the
figures. \emph{(i)}~Online, every $50$ iterations, within training: the aesthetic
score, CLIPScore and OCR accuracy, and in the mixed-reward setting HPSv2 and
PickScore; on DrawBench prompts \citep{saharia2022imagen} in the mixed-reward setting, and on the OCR
prompt set for OCR accuracy and for every dimension of the OCR setting.
\emph{(ii)}~Offline, at the checkpoints saved every $100$ iterations, on $256$ images
with seed $12345$ and the same prompts: HPSv3 and ImageReward
\citep{ma2025hpsv3,xu2023imagereward}. \emph{(iii)}~Offline,
GenEval2 (Soft-TIFA) \citep{kamath2025geneval2}, the successor to GenEval
\citep{ghosh2023geneval}, in its official evaluation setting, following
MeanFlowNFT \citep{huang2026meanflownft} (\Cref{app:hparams}). The LAION aesthetic score
\citep{schuhmann2022aesthetics}, HPSv3,
ImageReward and GenEval2 serve as held-out
evaluators, as do OCR accuracy in the mixed-reward setting and CLIPScore in
the OCR setting. Iteration $0$ is
the shared AnyFlow Init. checkpoint.

\subsection{Setup on Video Generation}\label{sec:exp-setup-video}
\paragraph{Generator and sampler.} The policy is the few-step generator of
\S\ref{sec:prelim} on the Wan2.1-T2V-1.3B backbone \citep{wan2025wan},
distilled on-policy with AnyFlow \citep{gu2026anyflow}
against its frozen teacher as in the image setting. Reinforcement
learning trains a LoRA \citep{hu2022lora} of rank $32$ (scaling $64$) on the attention
projections, at $480\times832$ and $81$ frames, i.e.\ latents of shape
$21\times16\times60\times104$. We set $M_{\mathrm{step}}=4$ steps without
classifier-free guidance \citep{ho2021classifierfree}, with the deterministic sampler of
\eqref{eq:rollout-op} for every method except FewStepGRPO, which requires
\eqref{eq:rollout-sde}. Evaluation always uses the deterministic sampler.

\paragraph{Optimisation.} Each outer iteration draws two batches, each with
$2$ prompts per rank and $K=16$ members, on $32$ NVIDIA H20 GPUs, i.e.\ $2048$
rollouts per outer iteration. We use AdamW \citep{loshchilov2019adamw} with
learning rate $3\times10^{-6}$ and gradient-norm clipping at $1$; $\uold$ is an
EMA of $\uth$ refreshed once per outer iteration; for the NFT/OPSD families,
$\beta=0.1$ and $c_{\mathrm{adv}}=5$ in \eqref{eq:rhat}; the KL-u penalty
against the fixed AnyFlow initial policy uses \eqref{eq:kl-uref} with weight $10^{-4}$; and each member is
queried $n_{\mathrm{ts}}=4$ times, with half of each batch set to $r=t$ and a
quarter to $r=0$.
For FewStepOPSD and ReFL, the world size is $48$ instead, because
computing the gradient with respect to the video latents requires much memory,
while the per-rank batch is unchanged. FewStepOPSD and ReFL therefore use
$3072$ rollouts per outer iteration, $1.5$ times as many as the 32-GPU runs.
The runs are thus matched in the number of outer iterations but not in the
number of samples, and the quality comparison should be interpreted with this
caveat. GPU-hours remain comparable across runs, whereas s/iter is reported but
is not comparable across runs with different world sizes.

\paragraph{Rewards.} Using the same four reward components as the
LongCat/MeanFlowNFT video setup
\citep{meituan2025longcatvideo,huang2026meanflownft}, we define our training
reward as an unweighted sum. The HPSv3 scorer \citep{ma2025hpsv3} enters through
HPSv3-general and HPSv3-percentile (per-frame, the latter the mean of the top
$30\%$ of frames under the caption); VideoAlign's motion quality (MQ) and text
alignment (TA) are drawn from video-feedback scoring
\citep{liu2025videofeedback}. MQ and TA are
$z$-normalised before
entering the sum; tables report their raw scorer outputs.
Only HPSv3 is differentiable here, and its two reductions account for
$96\%$ of the aggregate reward under the $z$-normalised weights
(\Cref{app:exp-video-reward}). The VideoAlign reward takes file paths as
input and decodes the videos in its own pipeline, so there is no differentiable
path from the latents to this reward. FewStepOPSD and ReFL therefore use only
the gradients of the HPSv3 terms, while they are \emph{evaluated} on all four
terms; \Cref{app:exp-video-reward} quantifies this asymmetry, which does not
occur in the image setting.

\paragraph{Specific settings.} As in the image setting, $n_{\mathrm{it}}=4$ for
 the group-advantage-based methods ($n_{\mathrm{it}}=M_{\mathrm{step}}$ for
 FewStepGRPO, $n_{\mathrm{it}}=n_{\mathrm{ts}}$ for the others), and the
 auxiliary velocity $\vph$ is fitted by \eqref{eq:auxfit} with learning rate
 $10^{-5}$ and $n_{\mathrm{fm}}=5$ steps per outer iteration for DMDAWM and
 DMDNFT. Because there is one reward setting rather than two, the gate radius
 $\varepsilon_{\mathrm{clip}}=10^{-3}$ is used for both
 $\chi_{\mathrm{GRPO}}$ in FewStepGRPO and $\chi_{\mathrm{AWM}}$ in
 MeanFlowAWM/DMDAWM, and FewStepGRPO uses $\varsigma=0.7$ in
 \eqref{eq:grpo-v-mean}.
 FewStepOPSD and ReFL require further settings, because the video backbone
 imposes three constraints on $\nabla R$ that do not occur in the image
 setting. First, as stated above, $\nabla R$ is computed from
 the two HPSv3 reductions instead of the full training reward; the VideoAlign terms still
 enter the reward and the advantage but do not contribute to the gradient.
 Second, the reward model is cast to fp32 with the \texttt{sdpa} attention
 backend; with the released \texttt{bfloat16} weights, the reward
 \emph{gradient} changes direction substantially while the reward \emph{value}
 changes by only ${\sim}0.003$, and since \eqref{eq:opsd-target} normalises
 $\nabla R$ to unit length and thus uses only its direction, this error
 directly affects the update. Third, the video is decoded one latent frame at
 a time, so $\nabla R$ is computed from $21$ of the $81$ pixel frames of a
 clip, each decoded without its causal prefix; \Cref{app:exp-video-reward}
 reports the memory measurements that make this necessary.

\paragraph{Evaluation.} We use two evaluations that measure different
properties. \emph{(i)}~Every $10$ iterations, we compute the four training rewards on
$256$ held-out prompts with seed $12345$. Since this is the training objective,
it cannot distinguish a quality gain from reward hacking. \emph{(ii)}~At iteration
$250$, we evaluate VBench \citep{huang2024vbench} on $946$ GPT-augmented
prompts $\times\,5$ samples $=4730$ videos per method, with all $16$
dimensions aggregated by the quality/semantic/overall weights of AnyFlow
\citep{gu2026anyflow}. No method optimises any part of it. Iteration $0$ is the
shared pre-RL checkpoint trained with AnyFlow.

\subsection{Main Results on Image Generation}\label{sec:exp-main}
\begin{table}[tbp]
\centering
\footnotesize
\setlength{\tabcolsep}{3.5pt}
\renewcommand{\arraystretch}{1.15}
\textbf{(a) OCR setting}\par\smallskip
\resizebox{\textwidth}{!}{%
\begin{tabular}{@{}lrccccrcccccc@{}}
\toprule
& & \multicolumn{4}{c}{OCR accuracy at iteration} & first it.\ with
& \multicolumn{6}{c}{at the first saved checkpoint with OCR accuracy $>0.82$} \\
\cmidrule(lr){3-6}\cmidrule(l){8-13}
trainer & s/iter & 50 & 100 & 150 & 200 & OCR $\ge0.85$ (h)
& it. & OCR & aesthetic & HPSv3 & ImageReward & GenEval2 \\
\midrule
AnyFlow Init. & --- & --- & --- & --- & --- & ----- & 0 & 0.498 & 5.24 & 12.86 & 1.233 & 0.363 \\
FewStepGRPO & 451 & 0.599 & 0.645 & 0.708 & 0.776 & ---$^{a}$ & --- & --- & --- & --- & --- & --- \\
MeanFlowAWM & 496 & 0.575 & 0.661 & 0.755 & 0.802 & ---$^{a}$ & --- & --- & --- & --- & --- & --- \\
DMDAWM & \textbf{342} & 0.698 & 0.820 & 0.902 & \textbf{0.936} & 150 (14.3) & 200 & 0.936 & 4.49 & 5.05 & 0.092 & 0.244 \\
MeanFlowNFT & 483 & \textbf{0.785} & \textbf{0.850} & \textbf{0.945} & {0.934} & \textbf{100 (13.4)} & 100 & 0.850 & \textbf{4.99} & \textbf{10.64} & \textbf{0.887} & \textbf{0.306} \\
DMDNFT & 346 & 0.765 & 0.829 & 0.879 & 0.897 & 150 (14.4) & 100 & 0.829 & 4.69 & 7.58 & 0.556 & 0.277 \\
\midrule
FewStepOPSD & \multicolumn{12}{c}{\multirow{2}{*}{not applicable: OCR accuracy has no gradient with respect to the sample decoded from rollout latents}} \\
ReFL & \multicolumn{12}{c}{} \\
\bottomrule
\end{tabular}}

\smallskip
{\raggedright\scriptsize $^{a}$Not reached within the $200$-iteration horizon of
\Cref{fig:ocr}; OCR accuracy at iteration $200$ is $0.776$ (FewStepGRPO) and
$0.802$ (MeanFlowAWM).\par}

\bigskip
\textbf{(b) Mixed-reward setting, iteration $400$}\par\smallskip
\resizebox{\textwidth}{!}{%
\begin{tabular}{@{}lrrccccccccccc@{}}
\toprule
& & & \multicolumn{4}{c}{training reward} & \multicolumn{4}{c}{held-out reward}
& \multicolumn{3}{c}{GenEval2 (held-out)} \\
\cmidrule(lr){4-7}\cmidrule(lr){8-11}\cmidrule(l){12-14}
method & s/iter & h & sum & HPSv2 & PickScore & CLIPScore & aesthetic & OCR
& HPSv3 & ImageReward & it.\ 400 & mean & best (it.) \\
\midrule
AnyFlow Init. & --- & --- & 1.450 & 0.297 & 0.864 & 0.289 & 5.36 & 0.507 & 10.30 & 1.113 & 0.363 & --- & --- \\
FewStepGRPO & 423 & 47.0 & 1.461 & 0.311 & 0.867 & 0.283 & 5.80 & 0.382 & 11.58 & 1.193 & 0.379 & 0.375 & 0.379 (400) \\
MeanFlowAWM & 425 & 47.2 & 1.477 & 0.308 & 0.890 & 0.279 & 5.92 & 0.413 & 12.59 & 1.265 & 0.363 & 0.357 & 0.397 (100) \\
DMDAWM & 304 & 33.8 & 1.498 & 0.326 & 0.891 & 0.281 & 5.86 & 0.424 & 13.10 & 1.272 & 0.353 & 0.346 & 0.377 (100) \\
MeanFlowNFT & 435 & 48.3 & 1.509 & 0.329 & 0.899 & 0.281 & \textbf{6.04} & 0.461 & 13.23 & 1.385 & 0.407 & 0.393 & 0.411 (300) \\
DMDNFT & 326 & 36.2 & 1.519 & 0.332 & 0.902 & 0.285 & 5.98 & 0.448 & \textbf{13.49} & 1.343 & 0.412 & 0.375 & 0.412 (400) \\
\midrule
FewStepOPSD& 268 & 29.8 & \textbf{1.641} & \textbf{0.373} & \textbf{0.946} & \textbf{0.322} & 5.91 & 0.493 & 13.06 & \textbf{1.476} & \textbf{0.463} & 0.442 & 0.463 (400) \\
ReFL & \textbf{219} & \textbf{24.3} & 1.623 & 0.368 & 0.938 & 0.317 & 5.89 & \textbf{0.546} & 13.24 & 1.469 & 0.452 & \textbf{0.451} & \textbf{0.465} (200) \\
\bottomrule
\end{tabular}}

\caption{Main results. (a)~OCR setting, restricted to the $200$-iteration
horizon of \Cref{fig:ocr}.  (b)~Mixed-reward setting at
iteration~$400$: s/iter and wall-clock hours to iteration $400$; the training
reward and its three summands; the held-out evaluators; GenEval2 at iteration
$400$, averaged over iterations $100$--$400$, and at the best checkpoint.
Bold: best among the trained methods per comparable column.}
\label{tab:main}
\end{table}

\begin{figure}[p]
\centering
\includegraphics[width=0.85\textwidth]{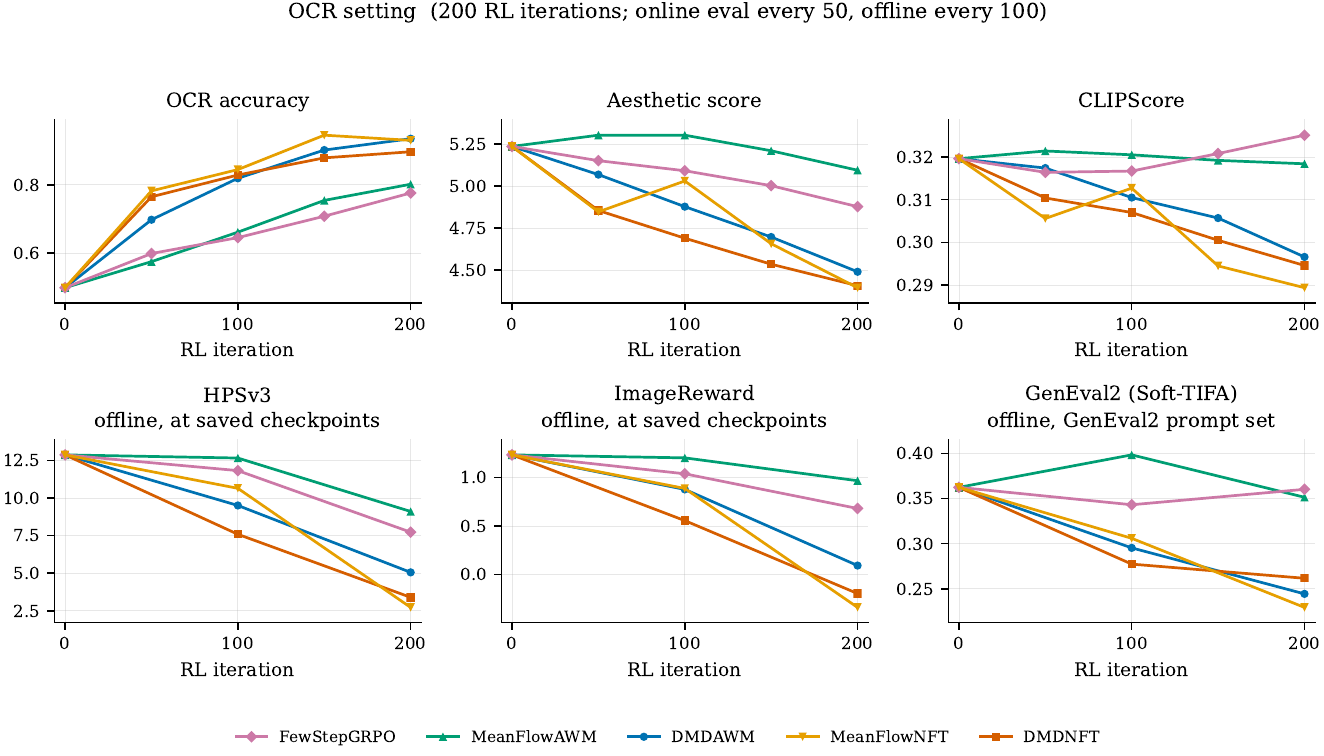}\vspace{-1em}
\caption{OCR setting: the evaluated dimensions against the iteration, over the
first $200$ iterations. Top row: online evaluation every $50$ iterations.
Bottom row: offline evaluation at saved checkpoints, every $100$ iterations;
HPSv3 and ImageReward on the OCR prompts, GenEval2 on its own prompts. }
\label{fig:ocr}\vspace{0.6em}
\includegraphics[width=0.85\textwidth]{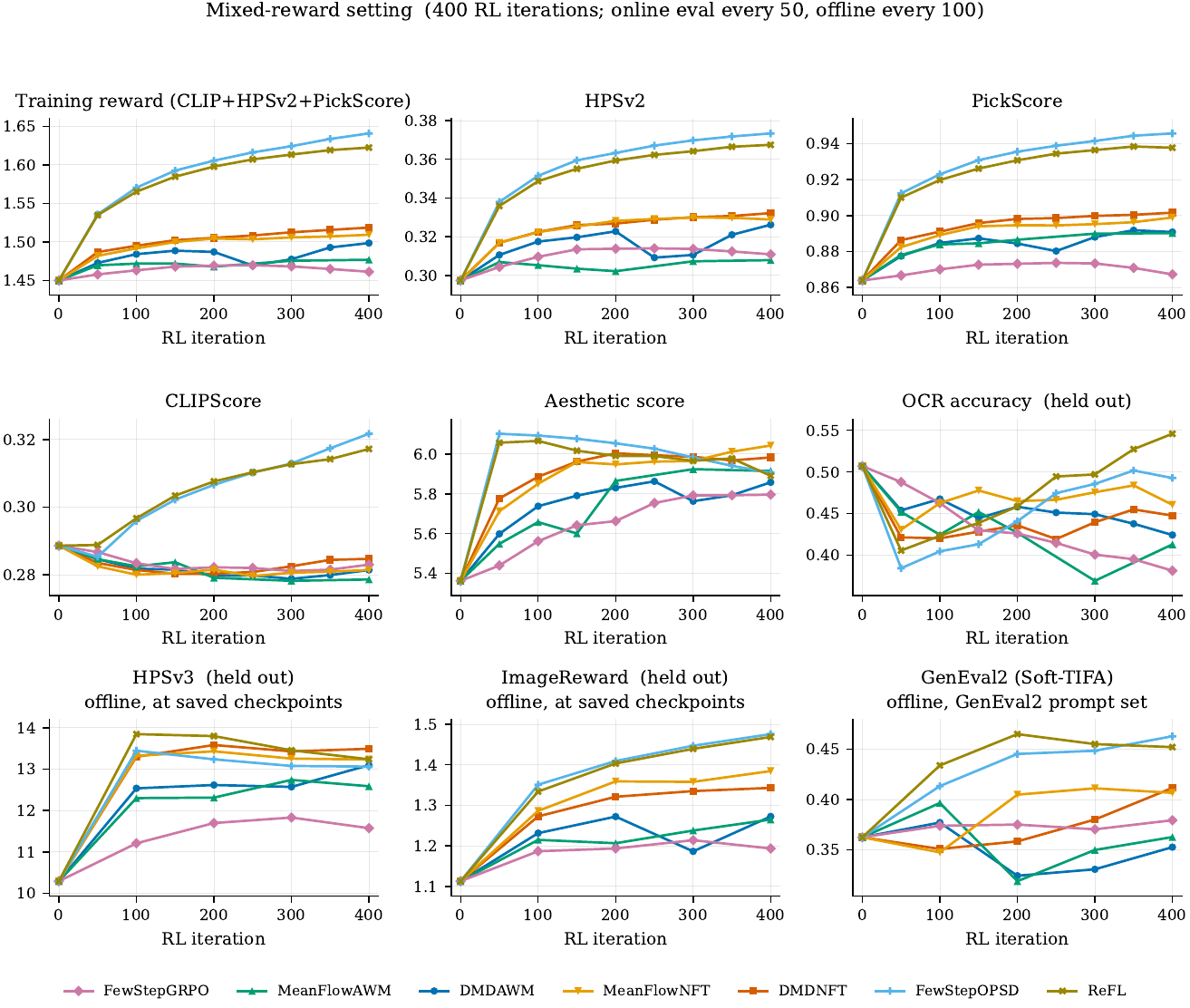}\vspace{-1em}
\caption{Mixed-reward setting: the evaluated dimensions against the iteration,
over the first $400$ iterations. The first panel is the training reward; OCR
accuracy, HPSv3, ImageReward and GenEval2 are held out. Online evaluation every
$50$ iterations, offline evaluation at saved checkpoints every $100$. }
\label{fig:mix}
\end{figure}
\begin{figure}[tbp]
    \centering
    \includegraphics[width=0.975\textwidth]{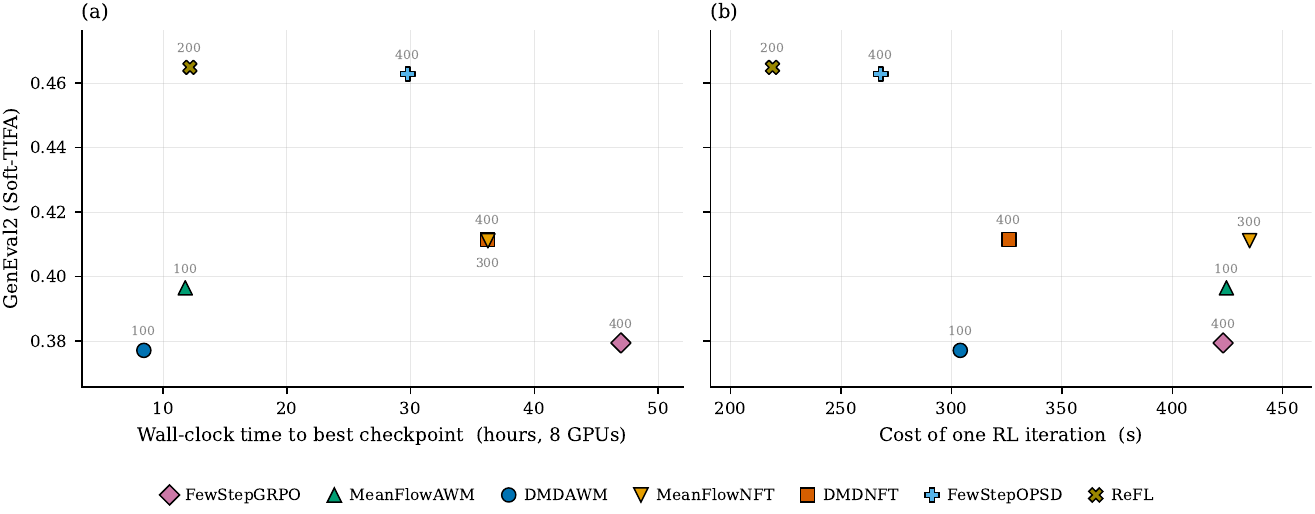}
    \caption{Mixed-reward setting: GenEval2 of each trainer's best checkpoint within
    $400$ iterations. (a)~Left: against the wall-clock time to reach it, s/iter
    times the iteration, on $8$ GPUs. (b)~Right: against s/iter. Both panels show
    the same checkpoints, so they share the vertical axis and the legend, and a
    trainer moves only horizontally from (a) to (b). The number beside each marker
    is the iteration of that checkpoint.}
    \label{fig:eff}
\end{figure}

\begin{figure}[p]
\centering
\begin{subfigure}{0.92\textwidth}
\centering
\includegraphics[width=\linewidth]{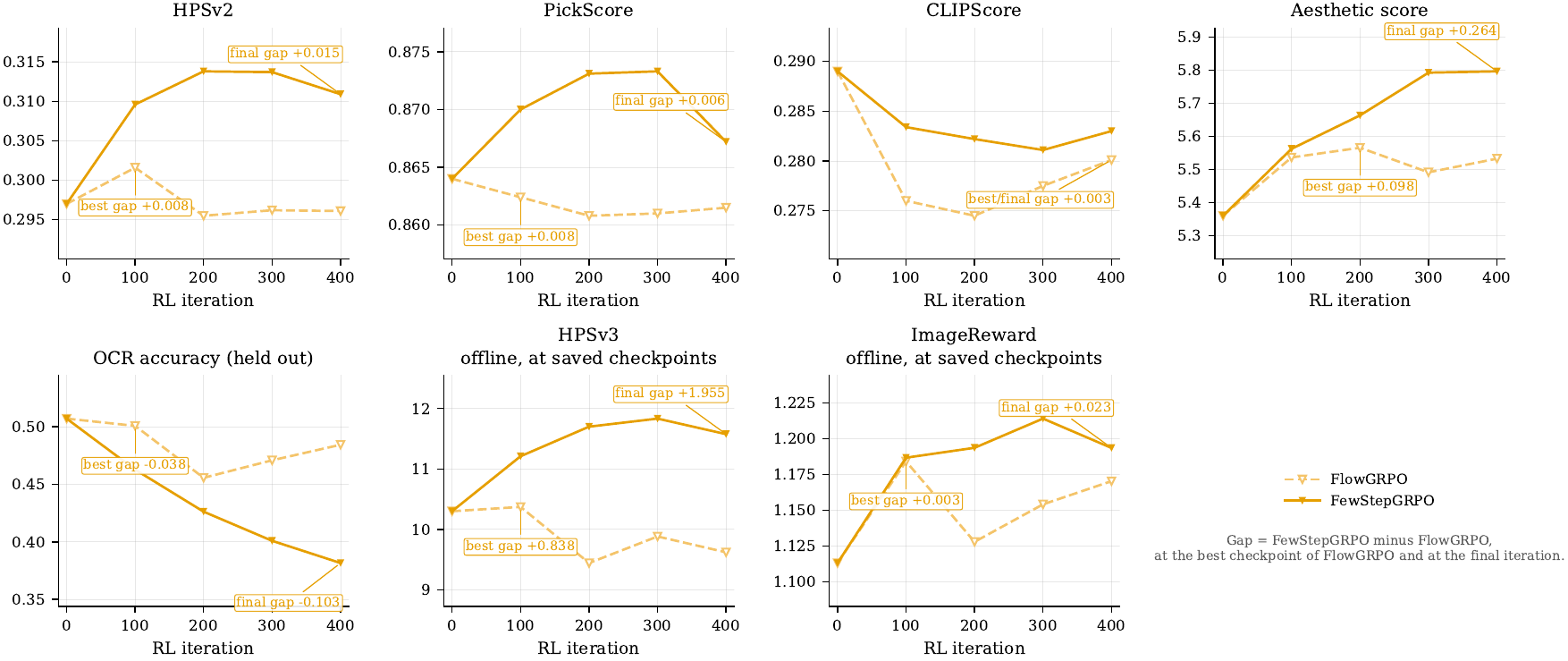}
\caption{FewStepGRPO (\S\ref{sec:fewstep-grpo}) vs.\ FlowGRPO}
\end{subfigure}\par\vspace{0.5em}
\begin{subfigure}{0.92\textwidth}
\centering
\includegraphics[width=\linewidth]{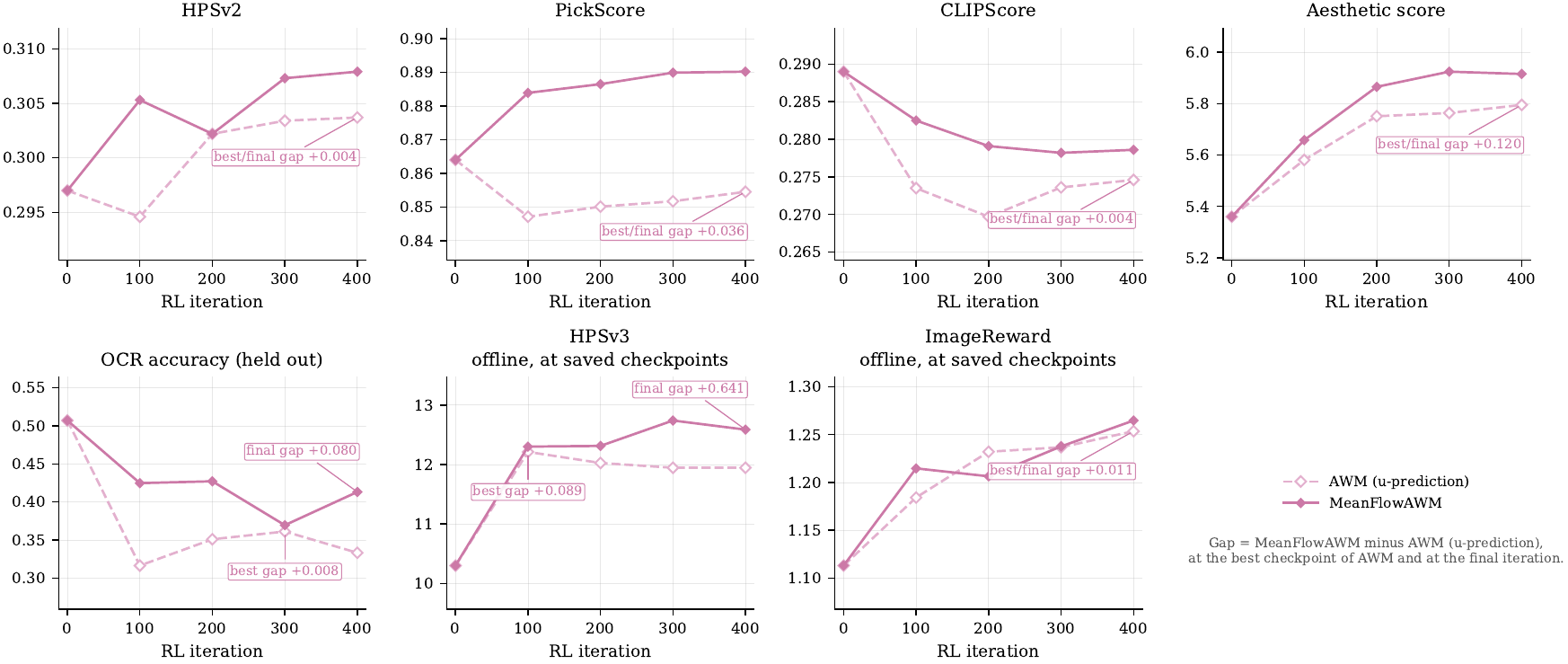}
\caption{MeanFlowAWM (\S\ref{sec:fewstep-awm}) vs.\ AWM}
\end{subfigure}\par\vspace{0.5em}
\begin{subfigure}{0.92\textwidth}
\centering
\includegraphics[width=\linewidth]{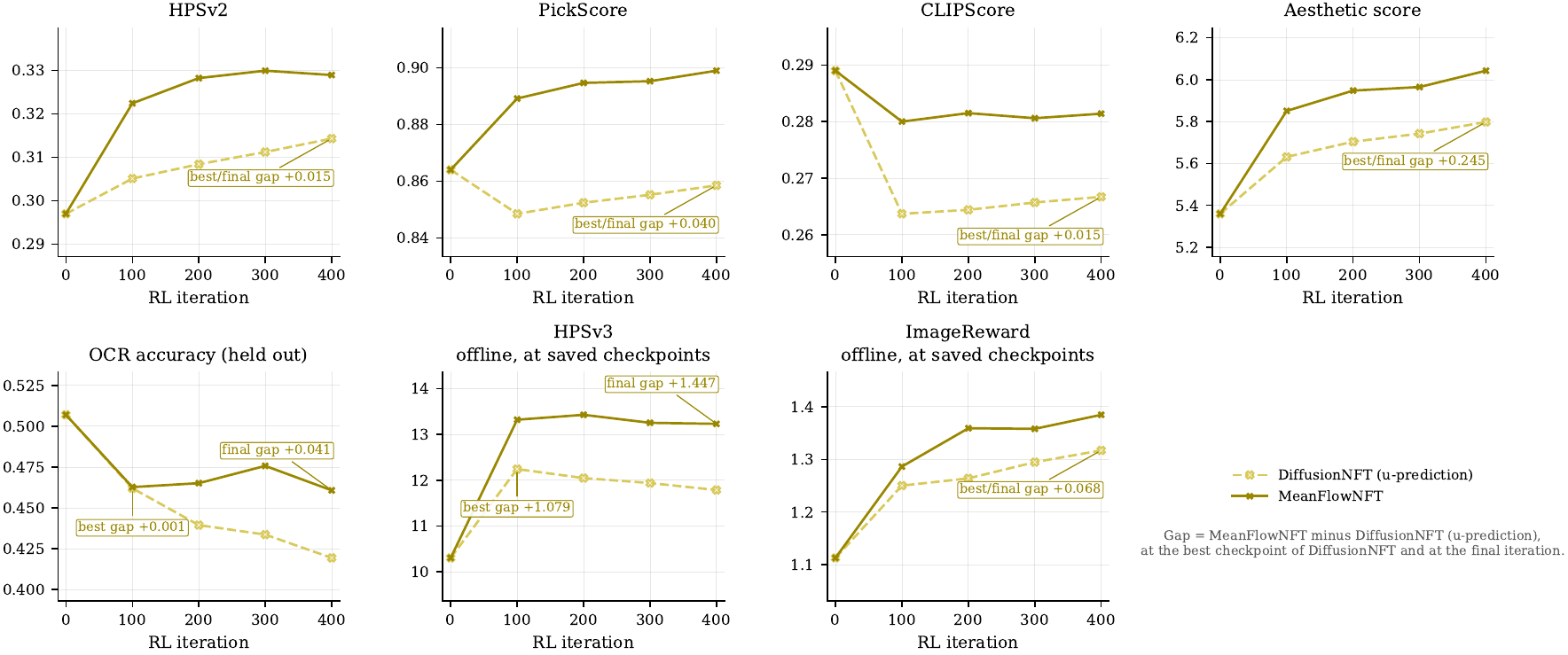}
\caption{MeanFlowNFT (\S\ref{sec:fewstep-nft}) vs.\ DiffusionNFT}
\end{subfigure}
\caption{Mixed-reward setting: the baselines of
\S\ref{sec:exp-setup} against the
few-step adaptations of this paper (solid), from the shared pre-RL checkpoint.
Panels as in \Cref{fig:mix}. Each annotated gap is the
adaptation minus the baseline, at the best checkpoint of the
baseline and at iteration $400$; the two coincide when the baseline is best
at iteration $400$. GenEval2 for the same
runs is shown in \Cref{fig:motivation}.}
\label{fig:upred-comparison}
\end{figure}

\paragraph{Effectiveness.} \Cref{tab:main} and \Cref{fig:ocr,fig:mix} support
the following observations.
\begin{enumerate}
\item The adaptation of \S\ref{sec:fewstep} is beneficial in these runs, and the
v-prediction form gives better performance in most of these comparisons
(\Cref{fig:upred-comparison,fig:motivation}). FewStepGRPO raises the
training reward from $1.450$ to $1.461$ and the held-out HPSv3 from $10.30$
to $11.58$, whereas FlowGRPO, its counterpart with direct employment of average-velocity, leaves the
training reward at $1.438$ and lowers HPSv3 to $9.62$.  Meanwhile, MeanFlowNFT as 
the v-prediction counterpart of DiffusionNFT, which is the direct employment of 
average-velocity, reaches a training
reward of $1.509$ against $1.440$, and is higher on the
held-out HPSv3 ($13.23$ against $11.79$), ImageReward ($1.385$ against
$1.317$) and best GenEval2 ($0.411$ against $0.399$), the last within the
${\sim}0.03$ resolution.
\item In the OCR setting, DMDAWM, MeanFlowNFT and DMDNFT raise OCR accuracy
from $0.498$ to at least $0.85$ within $100$--$150$ iterations, whereas
MeanFlowAWM and FewStepGRPO reach only $0.802$ and $0.776$ at iteration $200$.
Furthermore, it can be observed that within the $200$-iteration horizon of \Cref{fig:ocr}, the highest logged OCR
accuracy is reached by MeanFlowNFT, about $0.945$ at iteration $150$, and it
is also the first to reach $0.85$, at iteration $100$. DMDAWM starts lower at
iteration $50$ but rises steeply thereafter and
reaches $0.936$ at iteration $200$, where the two curves coincide within the
resolution of the plot (\Cref{tab:main}a). Its rapid rise is consistent with a combination of two properties.
\emph{(i)}~It has no self-regression
($\xreg=\varnothing$), so only reward augmentation contributes to its
decoupled update. \emph{(ii)}~It is DMD-based and models the instantaneous velocity
with the auxiliary network $\vph$, which is refitted to the current rollouts at
every iteration, so the baseline $\xbase=\xaux$ follows the
policy; since this baseline is fixed within an update, reward augmentation uses
the direction $A\,(\vcond-\vph)$ instead of one based on the directional-difference
estimate
\eqref{eq:cd}. This fast increase coincides with the large held-out
degradation; at its first checkpoint with OCR accuracy above $0.82$, HPSv3
drops from $12.86$ to $5.05$ and ImageReward from $1.233$ to $0.092$. 
\item In the mixed-reward setting, FewStepOPSD and ReFL achieve the two
highest training rewards, with FewStepOPSD highest at the final iteration.
Using the gradient of the full reward, the two raise the training reward from
$1.450$ to $1.64$ and $1.62$, against $1.46$--$1.52$ for the five
group-advantage methods, and they also lead on ImageReward and GenEval2. Their
curves nearly coincide (\Cref{fig:mix}); the logged coefficient magnitudes of
FewStepOPSD differ by $0.24$ between the two branches, i.e.\ $2.4\%$ of their
fixed sum $1/\beta=10$, although the two
reward-augmentation vectors also depend on their different increments
(\Cref{app:opsd-refl}). Among the five group-advantage methods, DMDNFT and
MeanFlowNFT come next and perform similarly (training reward $1.51$--$1.52$,
GenEval2 $0.41$ at iteration $400$).
\item In the default mixed-reward configuration, none of the five
group-advantage runs increases CLIPScore, which decreases slightly from $0.289$
to $0.279$--$0.285$, whereas FewStepOPSD and ReFL raise it to
$0.317$--$0.322$. Dedicated CLIP-focused group-advantage ablations do increase
it (\Cref{app:clipscore}).
\end{enumerate}

\begin{table}[tbp]
    \centering
    \footnotesize
    \setlength{\tabcolsep}{5pt}
    \renewcommand{\arraystretch}{1.15}
    \begin{tabular}{@{}lrrrcccc@{}}
    \toprule
    trainer & s/iter & sampling (s) & update (s) & $n_{\mathrm{it}}$ & forwards for $J$ & $ c_\phi$ & $n_{\mathrm{fm}}$ \\
    \midrule
    FewStepGRPO & 423 & 78.7 & 344.2 & 4 & 8 & 0 & 0 \\
    MeanFlowAWM & 425 & 78.3 & 344.5 & 4 & 8 & 0 & 0 \\
    DMDAWM & 304 & 78.2 & 223.4 & 4 & 0 & 1 & 5 \\
    MeanFlowNFT & 435 & 78.8 & 354.6 & 4 & 8 & 0 & 0 \\
    DMDNFT & 326 & 78.4 & 244.7 & 4 & 0 & 1 & 5 \\
    \midrule
    FewStepOPSD& 268 & 224.7 & 43.3 & 1 & 0 & 0 & 0 \\
    ReFL & 219 & 174.6 & 44.2 & 1 & 0 & 0 & 0 \\
    \bottomrule
    \end{tabular}
    \caption{Cost of one outer iteration in the mixed-reward setting, split into
    the sampling time (rollout, reward and, for FewStepOPSD and ReFL, the reward-coupled
    construction of \eqref{eq:rollout-dep}) and the update time; medians over the
    timers in the training logs, on $8$ H20 GPUs. The small residual between the
    sum of the two times and s/iter covers the refresh of $\uold$ and logging.}
    \label{tab:time}
    \end{table}

\paragraph{Efficiency.} We measure efficiency by the time per iteration
(\Cref{tab:time}) and the time to reach the best checkpoint
(\Cref{fig:eff}a). It can be observed that
\begin{enumerate}
\item The two reward-gradient-based methods are the cheapest per iteration,
$219$\,s for ReFL and $268$\,s for FewStepOPSD, because they compute one update
direction per member in the clean-sample space with one query of $\uth$
instead of four ($n_{\mathrm{it}}=1$), and because $J$ cancels in their loss.
However, their sampling time, which includes the reward-coupled rollout of
\eqref{eq:rollout-dep}, is $175$--$225$\,s per iteration, against about $78$\,s
for the plain rollout of the other five methods (\Cref{tab:time}). Since the
reward is defined in pixel space, its gradient with respect to the clean latent
requires back-propagation through both the reward model and the VAE decoder.
\item The directional-difference evaluations \eqref{eq:cd} and
\eqref{eq:cd-replay} for the shared-$J$ v-form plausibly account for most of the cost difference.
FewStepGRPO, MeanFlowAWM and MeanFlowNFT require two additional forward
passes of $\uold$ per query. These methods take
$423$--$435$\,s, against $304$--$326$\,s for DMDAWM and DMDNFT, which compute
the baseline from $\vph$ and compute no $J$. The difference arises in the
update time ($344$--$355$ against $223$--$245$\,s), and the OCR setting gives
the same ordering ($451$--$496$ against $342$--$346$\,s).
\item ReFL has a marginally higher best GenEval2 point estimate than
FewStepOPSD ($0.465$ against $0.463$), but the $0.002$ gap is below the stated
$0.03$ resolution and does not separate the two methods. ReFL reaches its
selected checkpoint after $12.2$\,h (iteration $200$), whereas FewStepOPSD
continues to iteration $400$ ($29.8$\,h);
both peak on the held-out aesthetic score and HPSv3 by iteration $100$ (the
aesthetic trajectory of FewStepOPSD is tabulated in \Cref{tab:cd}; HPSv3 is
read from \Cref{fig:mix}), after
which these scores decline while the training reward still rises, which
is consistent with the onset of reward hacking. Among the group-advantage-based methods,
MeanFlowAWM and DMDAWM reach their best GenEval2 early, at iteration $100$
($11.8$ and $8.4$\,h), but at lower values ($0.397$ and $0.377$).
\end{enumerate} 

\subsection{Main Results on Video Generation}\label{sec:exp-main-video}

\begin{table}[tbp]
\centering
\footnotesize
\setlength{\tabcolsep}{3.5pt}
\renewcommand{\arraystretch}{1.15}
\resizebox{0.95\textwidth}{!}{%
\begin{tabular}{@{}lrrrcccccccc@{}}
\toprule
& & & & \multicolumn{5}{c}{training reward} & \multicolumn{3}{c}{VBench (held-out)} \\
\cmidrule(lr){5-9}\cmidrule(l){10-12}
method & GPUs & s/iter & GPU-h & sum & HPSv3-G & HPSv3-P & MQ & TA
& overall & quality & semantic \\
\midrule
AnyFlow Init. & --- & --- & --- & 16.54 & 6.42 & 9.43 & 2.64 & 1.85 & 83.68 & 85.36 & 76.98 \\
FewStepGRPO & 32 & 735 & 1633 & 15.89 & 6.24 & 9.38 & 2.40 & 1.23 & 83.62 & 85.27 & 76.98 \\
MeanFlowAWM & 32 & 737 & 1637 & 16.66 & 6.49 & 9.46 & 2.68 & 1.84 & \textbf{83.70} & 85.36 & \textbf{77.04} \\
DMDAWM & 32 & 821 & 1824 & 16.78 & 6.51 & 9.53 & 2.74 & 1.82 & 83.64 & 85.35 & 76.76 \\
MeanFlowNFT & 32 & \textbf{734} & \textbf{1631} & 16.80 & 6.53 & 9.51 & 2.76 & 1.80 & 83.62 & 85.34 & 76.75 \\
DMDNFT & 32 & 821 & 1824 & \textbf{17.93} & \textbf{7.01} & \textbf{10.01} & \textbf{3.06} & 1.68 & 83.38 & 85.07 & 76.63 \\
\midrule
FewStepOPSD & 48 & 787 & 2623 & 16.26 & 6.30 & 9.32 & 2.54 & \textbf{1.91} & 83.60 & 85.37 & 76.53 \\
ReFL & 48 & 728 & 2426 & 16.46 & 6.39 & 9.38 & 2.61 & 1.89 & 83.69 & \textbf{85.40} & 76.85 \\
\bottomrule
\end{tabular}}

\smallskip
{\raggedright\scriptsize The four component columns show raw scorer outputs;
the sum uses the logged $z$-scores for MQ and TA. GPU-h $={}$GPUs
$\times$ s/iter $\times 250/3600$, truncated to integer hours, is the
comparable cost because the methods ran on different world sizes with a fixed
per-rank batch. Consequently, the 48-GPU rows use $1.5\times$ as many
rollouts per iteration and are not sample-budget-matched to the 32-GPU rows.
The dimension-level VBench results underlying the three
aggregates are reported in \Cref{tab:video-dims} of
\Cref{app:experiments}. For FewStepOPSD and ReFL, only part of the video reward
is differentiable, unlike in the image setting (\Cref{app:exp-video-reward}).\par}
\caption{Main results on video generation at iteration $250$. Left: cost.
Middle: the aggregate training reward and its four components on $256$
held-out prompts. Right: VBench, which no method optimises. Bold: best per
comparable column; s/iter is compared only among the 32-GPU rows.}
\label{tab:video-main}
\end{table}

\paragraph{Effectiveness.} \Cref{tab:video-main} and \Cref{tab:video-dims}
support the following observations.
\begin{enumerate}
\item FewStepGRPO is the least effective at increasing the training
reward. It is the only group-advantage-based method that ends below the shared
initialisation ($15.89$ against $16.54$), and its TA drops markedly, from
$1.85$ to $1.23$. Among the group-advantage-based methods DMDNFT is the most
effective, raising the training reward to $17.93$ ($+1.39$) and leading on
three of its four terms. In this run the larger training-reward gain coincides
with larger held-out degradation. Against
MeanFlowAWM, the highest on VBench overall, DMDNFT is $0.32$ lower in VBench overall
($83.38$ against $83.70$, last of the seven), $5.56$ lower in dynamic degree
($53.33$ against $58.89$, the lowest of any method) and $0.16$ lower in TA
($1.68$ against $1.84$). 
\item Among the remaining methods, FewStepOPSD and ReFL have the lowest training reward
after FewStepGRPO,
ending at $16.26$ and $16.46$, slightly below the initialisation. For video, the
reward gradient must be computed for every latent token of a long sequence,
through the decoder and the reward model, which requires several numerical
simplifications in the reward-gradient-based runs. Their VBench point
estimates degrade less than that of DMDNFT in these runs, which does not
establish that they protect against reward exploitation. TA rises above
the initialisation ($1.91$ and $1.89$ against $1.85$),
ReFL is second on VBench overall ($83.69$) and first on quality ($85.40$), and
FewStepOPSD has the highest aesthetic quality, background consistency and
appearance style of any method ($70.87$, $95.68$ and $74.53$); FewStepOPSD
and ReFL have the second- and third-highest dynamic degree ($59.44$ and $59.17$).
\end{enumerate}

\paragraph{Efficiency.} The GPU-hours in \Cref{tab:video-main} lead to the
following observations.
\begin{enumerate}
\item Unlike in the image setting, the DMD-based methods are slower than their
MeanFlow counterparts, in both s/iter ($821$ against $734$--$737$\,s) and
GPU-hours. With $n_{\mathrm{fm}}=5$, omitting the directional difference saves up to
$8$ forward passes of $\uold$ without gradient per member ($n_{\mathrm{ts}}=4$
queries, two each) but adds $5$ forward and backward passes of $\vph$ per outer
iteration. The fit costs about $144$\,s per iteration and the net s/iter
difference is $+84$--$87$\,s, so the saved $J$ passes account for about
$60$\,s by difference; the fit dominates, plausibly because backward passes through
the larger video model are expensive.
\item The reward-gradient-based methods lose their cost advantage from the
image setting and are the two most expensive in GPU-hours. FewStepOPSD and ReFL
have a sampling time of up to $674$\,s per iteration (FewStepOPSD), which includes the
reward-gradient computation, against ${\sim}291$\,s for the other methods, and
an update time of only $112$\,s. The gradient of a video reward with respect to the
latent tokens is expensive, plausibly because the latent sequence is much longer than for
images and because both the reward model and the decoder are larger than their
image counterparts. In addition, MQ and TA are not differentiable
(\S\ref{sec:exp-setup-video}), which further limits these two methods.
\end{enumerate}

\subsection{Test-Time Scaling Analysis}
\label{sec:tts}
In deployment, a few-step model should ideally support several operating
points on the quality--latency trade-off, using its training step count for
most requests and more sampling steps when higher quality is required. This
motivates training at a small $M_{\mathrm{step}}$ while allowing more solver
steps at inference. If quality improves without retraining, the model supports
test-time scaling and can trade compute for output quality.
Flexible-step generation and inference-time scaling have also been studied for
consistency-type models, improved MeanFlow and scalable generators
\citep{kim2024ctm,wang2024phasedconsistency,geng2026improvedmeanflow,xie2025sana15};
our experiment is restricted to the checkpoints and solver grid below.

\paragraph{Setup.} We therefore consider the image generation task, and take the checkpoints trained and evaluated
at $M_{\mathrm{step}}=4$ in \S\ref{sec:exp-setup} and sample each at
$M_{\mathrm{step}}\in\{4,8,16,32\}$. We use the checkpoints listed in
\Cref{tab:tts}. We
keep the $256$
DrawBench prompts, seed $12345$, resolution, deterministic sampler
\eqref{eq:rollout-op}, absence of classifier-free guidance, and the time shift
$3.0$ of \S\ref{sec:exp-setup} fixed. Increasing $M_{\mathrm{step}}$ only refines the same shifted
solver grid; no model is retrained. Thus the experiment isolates a
$1{\times}$, $2{\times}$, $4{\times}$, or $8{\times}$ inference-step budget.
As references, we also evaluate the AnyFlow initialisation, from which all
seven trainers start, and the pretrained SD3.5-Medium, with the same prompts,
seed, resolution and shifted grid. SD3.5-Medium models the instantaneous
velocity and is not distilled for few-step sampling, so it uses the Euler step
\eqref{eq:rollout-euler} with classifier-free guidance of scale $4.5$, i.e.,
two network evaluations per step.

\begin{figure}[tbp]
\centering
\includegraphics[width=\textwidth]{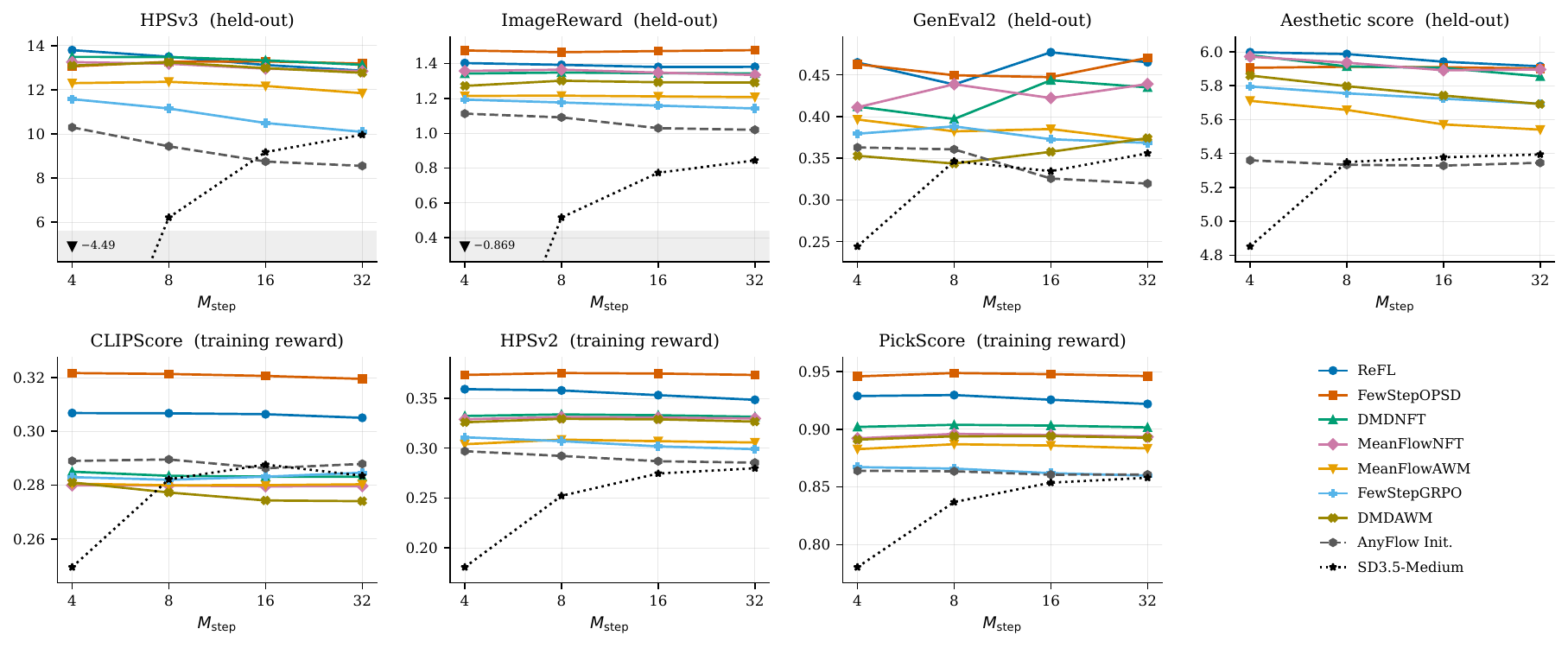}
\caption{Absolute value of each evaluated dimension against
$M_{\mathrm{step}}$, one line per fine-tuned model, with the AnyFlow
initialisation (dashed) and the pretrained SD3.5-Medium sampled with classifier-free guidance (dotted) as references. For the fine-tuned
models, the three training-reward summands (CLIPScore, HPSv2 and PickScore)
show little systematic gain from additional steps, while the held-out HPSv3 and
aesthetic score more often decline. Trainer labels are as in \Cref{tab:arms};
the checkpoint is given in \Cref{tab:tts}. \Cref{fig:tts-trend-rel} and
\Cref{fig:tts-trend-arm} in \Cref{app:experiments} show the same results as
percentage changes from the four-step value, grouped by evaluator and by model,
respectively.}
\label{fig:tts-trend}
\end{figure}

\begin{figure}[!htb]
    \centering
    \includegraphics[width=0.78\textwidth]{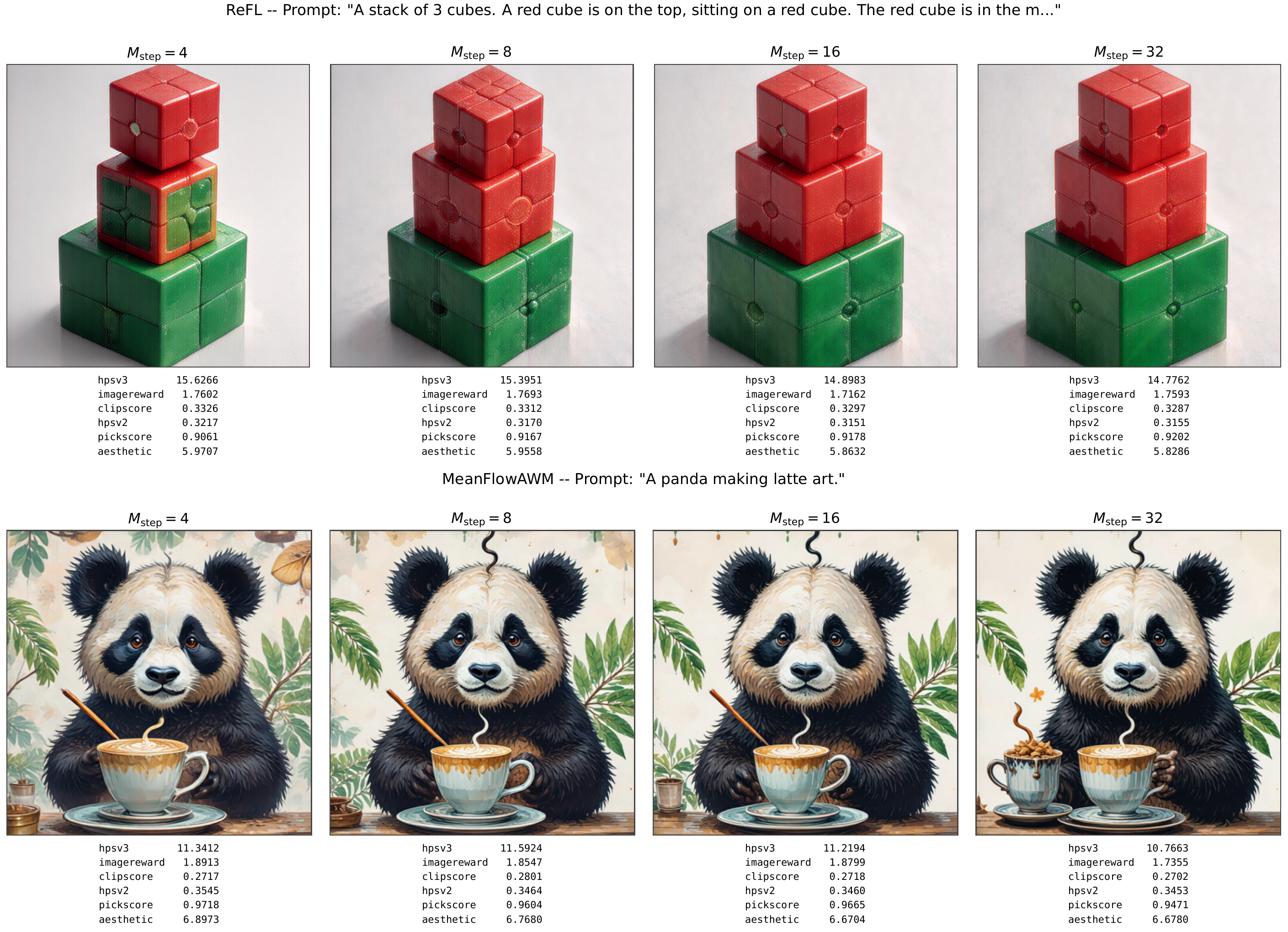}
    \caption{Two selected mild-degradation cases; each panel shows the
    six evaluated scores of that image (CLIPScore, HPSv2, PickScore, HPSv3,
    aesthetic score and ImageReward). Top, ReFL on the three-cube prompt, where
    HPSv3 gradually falls from $15.63$ to $14.78$. Bottom, MeanFlowAWM on ``A
    panda making latte art'', where HPSv3 first rises from $11.34$ to $11.59$ at
    $M_{\mathrm{step}}=8$, then falls to $10.77$ at $32$ as an extra cup appears.
    }
    \label{fig:tts-qual}
    \end{figure}

\begin{figure}[!tbp]
\centering
\includegraphics[width=0.78\textwidth]{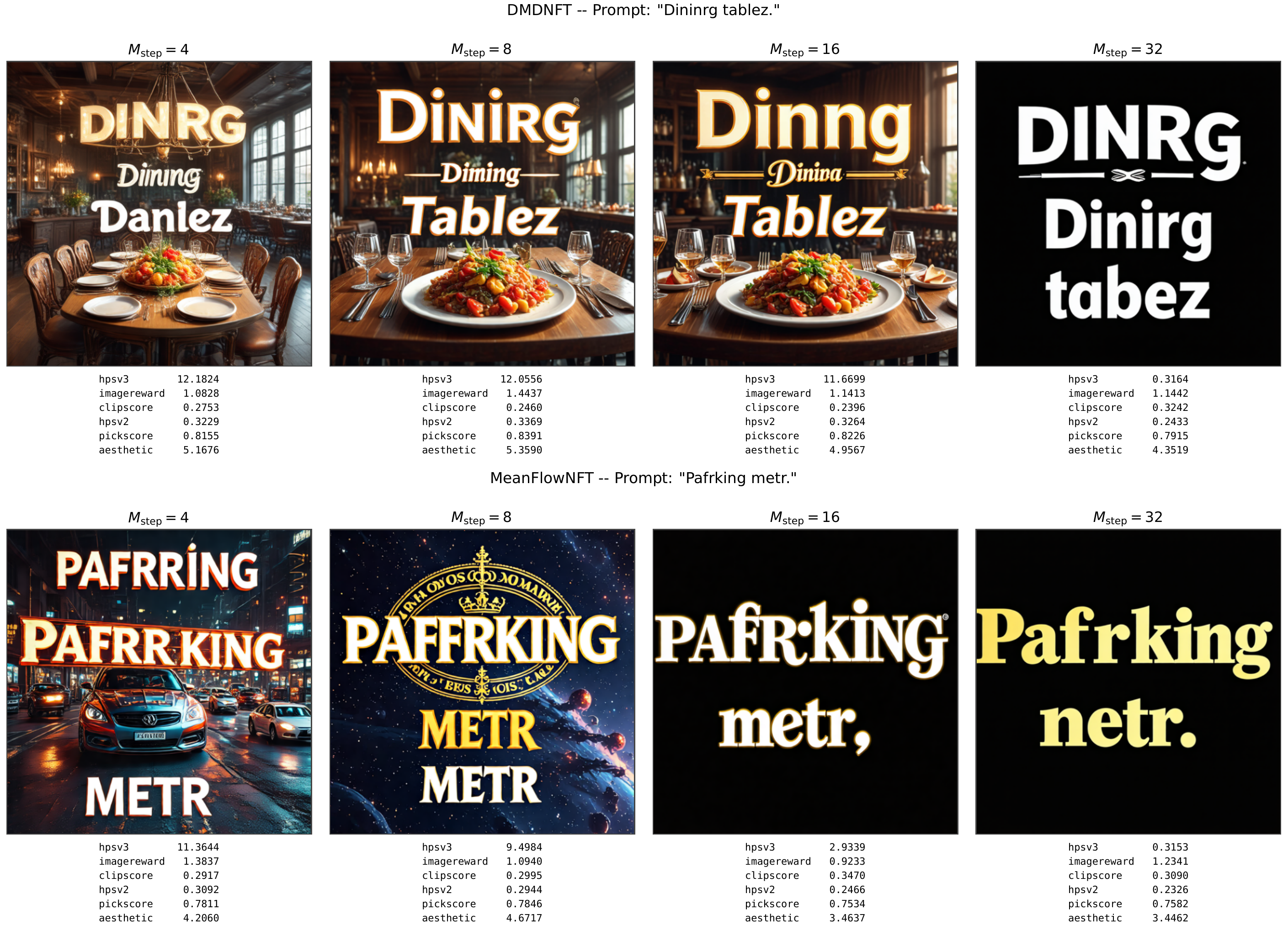}
\caption{Two selected OCR-sensitive text-dominance cases under test-time scaling: DMDNFT
(top) and MeanFlowNFT (bottom). Within each row, the prompt and seed are fixed
and only $M_{\mathrm{step}}$ changes; the values below each image are that
image's rewards. These are selected qualitative failure cases rather than
population averages.}
\label{fig:tts-reward-hacking}

\medskip
\includegraphics[width=0.78\textwidth]{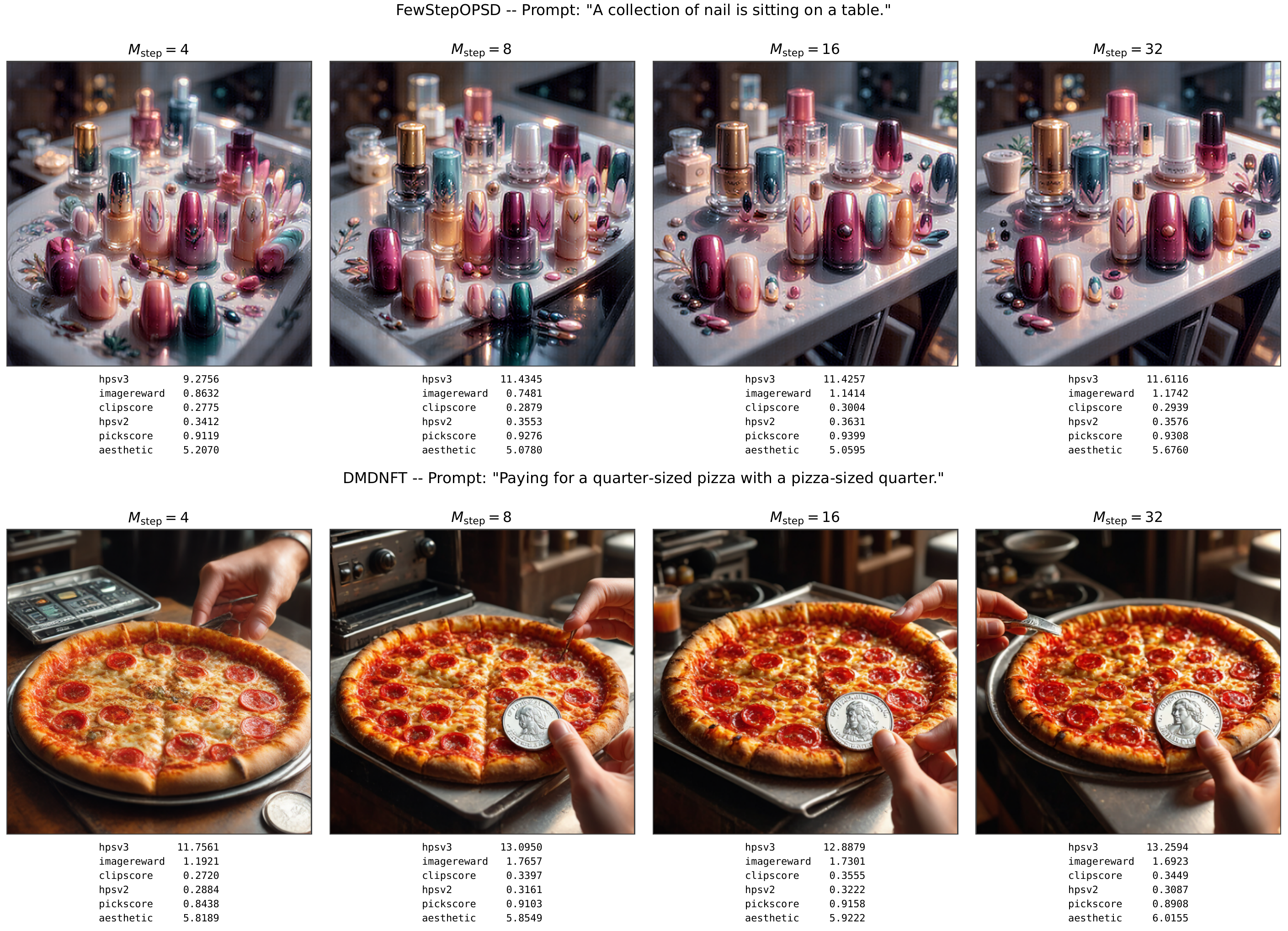}
\caption{Selected cases where a modest inference-step increase helps. Top,
FewStepOPSD on the nail-products prompt: the $4$-step image is cluttered and
soft, whereas the $8$- and $16$-step outputs have clearer objects and a more
stable arrangement.}
\label{fig:tts-gain}
\end{figure}

\paragraph{Main results.} \Cref{fig:tts-trend} shows the per-step trends, and
\Cref{tab:tts} in \Cref{app:experiments} reports the percentage changes. We
draw four conclusions.
\begin{enumerate}
\item \textbf{Models at different training stages show different
test-time scaling behaviour.}
The pretrained SD3.5-Medium models the instantaneous velocity, so four Euler
steps are far too coarse; its scores rise sharply from four to eight steps and
keep improving more slowly up to $32$ steps (HPSv3 from $-4.49$ to $9.96$,
\Cref{fig:tts-trend}).
Even at $32$ steps, i.e., $64$ network evaluations, it remains below every
fine-tuned model at four steps in HPSv3, ImageReward, the aesthetic score,
HPSv2 and PickScore. In contrast, the AnyFlow initialisation, distilled from
SD3.5-Medium for few-step sampling, appears to have already fixed its output patterns at
the step count used in training, and its HPSv3, ImageReward and GenEval2 drop
markedly with more steps; for example, from $4$ to $32$ steps, ImageReward
falls from $1.11$ to $1.02$ and GenEval2 from $0.363$ to $0.320$
(\Cref{fig:tts-trend}), beyond the
${\sim}0.03$ resolution. By comparison, the RL fine-tuned models, which all
start from this initialisation, fluctuate less in these three evaluators.

\item \textbf{There is no systematic gain on the training reward.}
Across the fine-tuned models, the three reward summands remain nearly flat
(\Cref{fig:tts-trend} and \Cref{tab:tts}). From $4$ to $32$ steps, the held-out
HPSv3 changes by between $+1.0\%$ and $-6.8\%$ for six of the seven trainers,
compared with $-17\%$ for the AnyFlow initialisation (\Cref{fig:tts-trend}). The stronger degradation
of the initialisation itself suggests that the decline in rewards of the fine-tuned
models is inherited from the initialisation rather than introduced by RL
fine-tuning. Further, these small changes are consistent with the
$4$-step outputs already having settled into patterns that score well under the
training reward, including sampler-specific reward exploitation.
Doubling to $M_{\mathrm{step}}=8$ often preserves or slightly improves a
score, but the gain generally saturates or reverses as the grid is refined
further. \Cref{fig:tts-qual} shows both behaviours; MeanFlowAWM improves at
eight steps and then declines, whereas ReFL declines gradually.

\item \textbf{OCR-sensitive prompts can expose a much sharper failure mode.}
The selected failures occur on malformed or OCR-heavy
prompts. A clear example is shown in
\Cref{fig:tts-reward-hacking}, where the DMDNFT and MeanFlowNFT
samples collapse
at $M_{\mathrm{step}}=32$ to little more than foreground text. HPSv3 falls
from $12.18$ to $0.32$ and from $11.36$ to $0.32$, respectively, even as
CLIPScore rises from $0.275$ to $0.324$ and from $0.292$ to $0.309$.
In these selected cases, additional solver steps produce increasingly
text-dominated outputs rather than a visual-quality improvement.

\item \textbf{Visible sample-level scaling exists.}
On inspection, many samples improve with $2$--$4$ times as many
sampling steps, even though the aggregate rewards barely change; on the
selected samples, HPSv3 does rise (\Cref{fig:tts-gain}). At four steps, they
have weak clarity or unstable details, while eight or sixteen steps yield
cleaner and more stable generations; \Cref{fig:tts-gain} shows two such
cases. The aggregate rewards thus appear less sensitive to these improvements than visual
inspection, and test-time scaling may be useful beyond what the aggregate scores
suggest, although more steps do not improve quality monotonically.
\end{enumerate}

\subsection{Practical Guidance}
The empirical results, each from a single run, suggest the following guidance:
\begin{enumerate}
\item \textbf{Where it is affordable, the reward gradient is our recommended
approach.} Driving the update directly by $\nabla R$ was the most effective choice
in our mixed-reward image experiments; FewStepOPSD and ReFL achieve the two highest
training rewards in the mixed-reward setting and improve all three summands
(\S\ref{sec:exp-main}). However, the applicability of these methods is
restricted in practice. \emph{(i)}~The gradient can become too expensive when the
reward model and the decoder are large or the latent token sequence is long, as
on video (\S\ref{sec:exp-main-video}). \emph{(ii)}~The reward must be differentiable
with respect to the clean sample, which excludes OCR accuracy
and VideoAlign's MQ and TA (\S\ref{sec:exp-setup},
\Cref{app:exp-video-reward}). Where both conditions hold, ReFL alone is
already a strong baseline in our runs.

\item \textbf{Otherwise, group-advantage-based methods apply.} When the reward
gradient is too expensive or unavailable, the reward enters only through the
group advantage, and these methods must convert the average velocity into the
instantaneous one (\S\ref{sec:fewstep}). Among them, the DMD-based methods,
which estimate the instantaneous velocity with the auxiliary network $\vph$,
have a lower per-iteration cost in the image setting but a higher one on
video, depending on the relative cost of the saved forward passes for $J$ and
the added forward and backward passes of $\vph$ (\S\ref{sec:exp-main},
\S\ref{sec:exp-main-video}). They need extra optimisation
steps and memory for
$\vph$, and in some of the reported runs their larger reward gains coincide
with larger held-out degradation.

\item \textbf{Among the group-advantage-based methods, the DiffusionNFT family
reached high reward values in relatively few iterations.} The
FlowGRPO-style replay of FewStepGRPO optimises slowly and has the weakest
observed reward endpoints within the common iteration budget: it barely improves
the mixed reward on images and loses the most reward on video. The DiffusionNFT family
reached high reward values within the common iteration budget in both settings, although the cost
per iteration varies; MeanFlowNFT is expensive in the image setting and DMDNFT
on video. AWM mostly lies between the two in the reported reward outcomes and uses a
PPO-style clip to limit each update.

\item \textbf{Test-time scaling of the fine-tuned few-step models saturates quickly.}
Aggregate scores show no systematic gain; three of the seven HPSv3 curves
peak at the training step count of four and one is tied between four and
eight steps (\Cref{tab:tts}). By visual inspection, however, many
individual samples look cleaner with $2$--$4$ times as many steps before the
effect saturates or reverses. Additional steps therefore improve the visual
quality of some samples but do not improve aggregate scores monotonically
(\S\ref{sec:tts}), and can degrade sample quality on OCR-sensitive prompts.
\end{enumerate}

\section{Conclusion}\label{sec:conclusion}

Few-step generators reduce the rollout cost that typically dominates RL post-training of
diffusion models, but they model an average velocity over an interval, whereas
existing RL objectives assume an instantaneous velocity and a many-step
sampler. This paper adapts these objectives to few-step models systematically
in three phases and compares the resulting methods within one codebase. The
transformation to $x$-space expresses each objective we consider through a
common policy prediction. The decomposition writes the gradient of each
objective as the sum of a reward-free self-regression and a reward
augmentation, so that a method reduces to a choice of three anchors and three
weights. The adaptation computes these anchors from a few-step model, at the
end of its transport or along its tangent, with the instantaneous velocity
recovered from the network itself through the MeanFlow identity (the
self-induced velocity) or from an auxiliary network. The seven resulting trainers share one training step and are
compared under shared rollout, reward and evaluation code, each in a single
run. Where the reward
is differentiable and its gradient affordable, FewStepOPSD and ReFL, which use
the reward gradient, were the most effective. Otherwise, the DiffusionNFT family
reached high rewards in relatively few iterations, and the DMD-based
variants were faster per iteration on image but slower on video, depending on
the relative cost of forward and backward passes, and in some runs coincided
with larger held-out degradation.
For the fine-tuned models, additional sampling steps gave no systematic
aggregate gain, although many individual samples improve with two to four
times as many steps as in training.

\paragraph{Open problems.} Several questions remain open.
\begin{enumerate}
\item The DMD-based variants estimate the instantaneous velocity with an
auxiliary network, which requires
additional forward and backward passes to fit this network and GPU memory to
hold it. How to reduce this overhead remains open.
\item Reward-gradient-based methods require a reward that is differentiable
with respect to the clean sample and a gradient that fits in memory. On video,
the motion-quality and text-alignment terms of VideoAlign provide no gradient,
and the per-frame decoding required by GPU memory makes the gradient deviate
strongly from that of a joint decode. Cheaper and more faithful reward
gradients for long latent sequences are needed.
\item With a scalar group advantage in the mixed-reward setting, whether every reward summand takes effect
depends on the weighting of the summands, on the rescaling by the
global-batch standard deviation in \eqref{eq:adv}, and on the small
within-group spread of some summands (\Cref{app:clipscore}). A general and more
reliable construction of the group advantage remains open.
\item In our experiments, a few-step model fine-tuned at one step count does not improve
systematically with more steps. Fine-tuning at more step counts may give
test-time scaling but makes the rollouts more expensive. Whether it can be
introduced cheaply during training, or without training, so that a model
fine-tuned at a small step count still improves with more steps at test time,
remains open.
\end{enumerate}

\clearpage
\bibliography{./materials/references_verified}

\appendix
\appendixtitle
\crefalias{section}{appendix}
\etocdepthtag.toc{appendix}
\setcounter{figure}{0}
\setcounter{table}{0}
\renewcommand{\thefigure}{S\arabic{figure}}
\renewcommand{\thetable}{S\arabic{table}}
\renewcommand{\theHfigure}{S\arabic{figure}}
\renewcommand{\theHtable}{S\arabic{table}}

\begingroup
\etocsettagdepth{mainmatter}{none}
\etocsettagdepth{appendix}{subsection}
\etocsettocstyle{\section*{Appendix Contents}}{}
\tableofcontents
\endgroup
\clearpage

\section{Notation}\label{app:notation}

This paper connects several families of methods and therefore uses a large
number of symbols. We collect them in the following table; each symbol is also
defined where it first appears.

\begin{longtable}[c]{>{\raggedright\arraybackslash\breakablecommas$}p{0.32\textwidth}<{$} p{0.60\textwidth}}
\toprule
\textbf{\text{Symbol}} & \textbf{Meaning} \\
\midrule
\endfirsthead
\toprule
\textbf{\text{Symbol}} & \textbf{Meaning} \\
\midrule
\endhead
\midrule
\multicolumn{2}{r}{\footnotesize\itshape continued on the next page}\\
\endfoot
\bottomrule
\endlastfoot
c & text prompt (conditioning); omitted from an argument list when it is fixed
or irrelevant. A bare $c$ always has this meaning; subscripted quantities such
as $c_{\mathrm{adv}}$ and $c_\theta$ are distinct \\
\epsilon,\ \epsilon' & standard Gaussian noise,
$\epsilon,\epsilon'\sim\mathcal N(0,I)$; a prime marks a fresh rollout-time
renoise draw \\
t, r & flow-map endpoints, $0\le r\le t\le 1$; generation runs from noise at
$t=1$ to data at $t=0$. The unhatted letter $r$ always names a noise level and
never a reward; the hatted $\hat r$ below is the mixing weight. Individual
objectives state any stricter query domain required by divisions by $t$ or
$t-r$ \\
\sigma & a generic level on the same scale, $\sigma\in[0,1]$, used as the time of the SDEs of
\Cref{app:grpo} \\
K & positive-integer \emph{group size}: the number of rollouts sharing one
prompt, $K\in\mathbb N_{\ge1}$; members carry the index $(i)$ \\
x & a \emph{clean sample}. Without a noise-level subscript, an $x$-symbol
denotes a quantity at $\sigma=0$; a subscript $t$, $r$ or $t_j$ indicates its
level, while $\theta$, $\mathrm{old}$, $\mathrm{real}$ or a role label indicates
its source. Method-specific quantities have the level of the object they
represent; in particular, the MeanFlow quantities of \eqref{eq:mf-decomp} and
the FewStepGRPO prediction and anchors of \eqref{eq:grpo-decomp} are defined at
level $r$ \\
x_t,\ x_r,\ x_\nu & a \emph{state on a trajectory} of either the
probability-flow ODE \eqref{eq:avgvel} or the sampler, as opposed to the
reconstructed state $\xt$ below. Among the reward methods, only those that replay a transition need
the states of the sampler, which \S\ref{sec:prelim} and Stage~1 write as
$\hat x_{\theta,t_j}$ and $\hat x_{t_j}$ \\
M_{\mathrm{step}},\ t_j & positive-integer number of sampler transitions and
their $M_{\mathrm{step}}+1$ grid levels,
$1=t_0>t_1>\dots>t_{M_{\mathrm{step}}}=0$ \\
\ell & step count drawn per update by the stochastic DMD sampler of
\S\ref{sec:prelim}, $\ell\in\{1,\dots,M_{\mathrm{step}}\}$ \\
\nu & integration variable along a trajectory in \eqref{eq:avgvel} \\
s,\ s_j & two distinct flow-map query levels: $s\in[0,t]$ is redrawn by
FewStepGRPO at training time, whereas $s_j$ is the rollout-time level drawn at
step $j$, with $s_j\le t_{j+1}$ (\eqref{eq:rollout-sde}) \\
\tau_j & non-negative renoise scale of the stochastic rollout
(\eqref{eq:rollout-sde}) \\
\midrule
\uth(x_t,t,r) & trainable \emph{average velocity} over $[r,t]$, which defines a flow map over $[r,t]$ for few-step generation \\
\uold & frozen old policy, with parameters $\theta_{\mathrm{old}}$; EMA of
$\uth$, refreshed once per outer iteration \\
v_\theta(x_t,t) & \emph{instantaneous velocity induced by the current policy},
i.e., the field whose flow map $\uth$ averages (\eqref{eq:avgvel}). As stated
in \S\ref{sec:prelim}, we identify it with the flow-matching marginal velocity under the distribution $p_{\theta}$ of
the current policy's samples instead of the data distribution,
\[
v_\theta(x_t,t)=\mathbb E_{\,\hat x\sim p_{\theta},\ \epsilon\sim\mathcal N(0,I)}
\!\left[\,\epsilon-\hat x \;\middle|\; x_t\,\right],
\]
the two coinciding when the flow map is exact; it is
recovered from $\uth$ through the MeanFlow identity, and a network that
parameterises the instantaneous field (\S\ref{sec:rl}) outputs it directly \\
v_{\mathrm{old}},\ p_{\mathrm{old}} & the instantaneous field of the old
policy, counterpart of $\uold$, and the distribution of its rollouts \\
v_{\mathrm{real}}(x_t,t) & the pretrained \emph{teacher} field, frozen; the distillation target \\
\vcond & \emph{conditional velocity} of a sample: $\vcond=\epsilon-x$ for a clean
$x$, and $\vcond^{(i)}=\epsilon-\xh^{(i)}$ for a rollout. It is the velocity
of the straight path from the sample to the noise, requires no network
evaluation, and is the single-sample regression target of every flow-matching
loss in this paper \\
\vph(x_t,t) & \emph{auxiliary velocity} with its own parameters $\phi$,
fitted online by flow matching to the student's samples so that it tracks
$p_\theta$ rather than the data distribution $p_{\mathrm{data}}$; in RL those
samples are the old-policy rollouts, so it tracks $p_{\mathrm{old}}$
(\eqref{eq:vphi-cond-old}) \\
\partial_t u & total derivative $\mathrm d u/\mathrm d t$ along the trajectory of
the instantaneous field that $u$ averages (\eqref{eq:avgvel}); the
implementation estimates it by the directional-difference estimate of $\uold$
along $\vcond$ (\eqref{eq:cd}), except on a replayed GRPO state, where the
tangent is $(x_t-\xh)/t$
(\eqref{eq:cd-replay}) \\
\Jterm{u}{t}{r} & \emph{detached Jacobian-vector term}
$(t-r)\,\sg(\partial_t u)$, i.e., the detached difference between the
instantaneous velocity at $t$ and the average velocity over $[r,t]$. Apart
from the MeanFlow loss \eqref{eq:mfloss}, it is evaluated on $\uold$ and
shared between the current and old policies; a bare $J$ denotes
$J[\uold,t,r]$ at the current query \\
\vind & \emph{self-induced instantaneous velocity}
$\vind=\uold+\Jterm{\uold}{t}{r}$ (\eqref{eq:vind}), computed from the average velocity and its
total derivative in $t$ without a second network; detached. It equals the
instantaneous velocity of $\uold$ wherever the flow map is exact \\
\midrule
\xhth & \emph{rollout sample} of the current policy, $\xhth=\Rollout(\epsilon,\uth,c)$,
run on the grid of $M_{\mathrm{step}}$ steps (\eqref{eq:rollout-def}).
The subscript $\theta$ indicates that $\xhth$ depends differentiably on
$\theta$, which requires a differentiable sampler \\
\xh & the rollout sample \emph{detached}, $\partial\xh/\partial\theta=0$;
every reward method uses this detached sample. In RL it is drawn by
the old policy, $\xh=\Rollout(\epsilon,\uold,c)$ (Stage~1), so its
distribution is $p_{\mathrm{old}}$ \\
\xt & \emph{renoised state} of the rollout sample, $\xt=\AddNoise(\xh,t)=(1-t)\xh+t\epsilon$,
with $\epsilon$ a fresh draw independent of the rollout's own noise \\
\xr=\Xmap{\xt}{v}{r} & \emph{transported state}, i.e., the state obtained by
moving $\xt$ along a velocity $v$ to level $r$, $\Xmap{\xt}{v}{r}=\xt-(t-r)v$;
normalised so that $\Xmap{\xt}{\epsilon-x}{0}=x$ \\
\xz & the \emph{policy prediction} differentiated by the loss, with clean-end
default $\xz=\Xmap{\xt}{v_\theta}{0}$ (\eqref{eq:xz-inst}), also called the
renoise-and-denoise sample in \Cref{fig:overview}. Its clean-end
instances include $\xu$ and $\xv$. For FewStepGRPO, the differentiated
prediction is instead the mean $\mu_\theta$ of the scoring kernel at level $r$
(\eqref{eq:grpo-decomp}). Every anchor is detached \\
\xzr & the policy's \emph{one-step transport} to level $r$, $\xzr=\Xmap{\xt}{\uth}{r}$\\
\xu & \emph{u-prediction}, $\xu=\Xmap{\xt}{\uth(\xt,t,0)}{0}$ \\
\xv & \emph{v-prediction},
$\xv=\Xmap{\xt}{\uth(\xt,t,r)+\Jterm{\uold}{t}{r}}{0}$ (\eqref{eq:vpred}), the
current-policy counterpart of $\xind$. Since $J$ is detached, its Jacobian is
that of $\Xmap{\xt}{\uth(\xt,t,r)}{0}$ \\
\Xmap{\xt}{\uth(\xt,t,r)}{0} & the \emph{$J$-free $(t,r)$ query}, i.e., $\xv$
with $J$ dropped; it has the same Jacobian as $\xv$ and is the current
prediction of DMDAWM, DMDNFT, FewStepOPSD and ReFL \\
\xind,\ \xaux & the two detached v-form anchors, named after the velocity
they use: the \emph{self-induced} prediction $\xind=\Xmap{\xt}{\vind}{0}$
(\eqref{eq:xz-vind}), anchor of the \texttt{v-ind} variants, and the
\emph{auxiliary} prediction $\xaux=\Xmap{\xt}{\vph}{0}$, baseline of the
\texttt{v-aux} variants (in DMD evaluated at $\xt'$, \eqref{eq:dmd-xdiff}) \\
\tilde x_{\mathrm{old}},\ \xoldu,\ \xoldur & the old policy's other
predictions at the same state: for an instantaneous-velocity policy
$\tilde x_{\mathrm{old}}=\Xmap{\xt}{v_{\mathrm{old}}}{0}$ (\eqref{eq:xold-def});
for an average-velocity policy the u-prediction
$\xoldu=\Xmap{\xt}{\uold(\xt,t,0)}{0}$ and the $J$-free
$(t,r)$ query $\xoldur=\Xmap{\xt}{\uold(\xt,t,r)}{0}$, which serves as
anchor in DMDAWM, DMDNFT, FewStepOPSD and ReFL
(\Cref{tab:hooks}) \\
\xreal & the teacher's prediction at the same state, $\Xmap{\xt}{v_{\mathrm{real}}}{0}$ \\
\xrealr,\ \xindr & teacher and self-induced predictions at level
$r$ (\eqref{eq:x-step}) \\
v_\theta^\pm,\ \tilde x_\theta^\pm & positive and negative implicit-policy
branches and their clean predictions
(\eqref{eq:nft-implicit}, \eqref{eq:nft-xbranch}) \\
\xreg & anchor 1, the \emph{regression anchor}, constructed for the
self-regression loss that regularises the tuning of $\uth$ \\
\xbase & anchor 2, the \emph{baseline anchor}, from which the reward
displacement is measured \\
\xrwd & anchor 3, the \emph{rewarded anchor}. It is the scored rollout $\xh$
in FewStepAWM and FewStepNFT, and a teacher prediction, a replayed action or a
constructed rewarded point in the other methods. Together with $\xbase$ it defines the
reward-augmentation displacement, whose direction is set by the sign of
$\omega$ \\
\midrule
R(\cdot,c),\ \dec & reward model and decoder; $R^{(i)}=R\big(\dec(\xh^{(i)}),c\big)$
is the endpoint reward of member $i$ \\
A & signed, detached advantage of the endpoint reward. Under
\eqref{eq:adv-prelim} it is normalised and zero-mean within the prompt group;
under the implemented \eqref{eq:adv} it is centred on the prompt's mean in the
current outer iteration and scaled by the current global-batch standard deviation \\
Z_c & positive normaliser in \eqref{eq:rhat-prelim},
$Z_c=\operatorname{std}(\{R\})+\varepsilon_{\mathrm{std}}$, so that the bracketed quotient
equals the $A$ of \eqref{eq:adv-prelim} \\
\hat r & bounded mixing weight, or \emph{optimality probability}, taking
values in $[0,1]$; monotone non-decreasing in $A$ and equal to $\tfrac12$ when
$A=0$. ReFL overrides it with $\hat r\equiv1$ \\
\beta & positive branch coefficient, $\beta>0$, that directly scales the
implicit-policy branch displacement; the induced policy-space reward step
later contains $1/\beta$ \\
\omega,\ \omega_k & detached signed reward coefficients (scalars) used in the
general loss. Their magnitudes are $|\omega|$ and $|\omega_k|$.
In FewStepGRPO and FewStepAWM, a non-zero $\omega$ has the sign of $A$, and the
clipping gate may set it to zero. FewStepNFT defines
$\omega$ directly, and FewStepOPSD uses a positive ascent coefficient
$\omega_1$ and a negative descent coefficient $\omega_2$ \\
W_{\mathrm{reg}},\ W_{\mathrm{rwd}} & non-negative, detached per-sample loss
weights. A bare $W$ is local shorthand for $W_{\mathrm{rwd}}$ when
$W_{\mathrm{reg}}=0$, or for their common value where they coincide \\
g,\ g_k & detached \emph{reward displacement},
$g=\omega(\xrwd-\xbase)$, with the branch-indexed counterpart used by
FewStepOPSD \\
\rho_{\mathrm{tr}} & true transition-density ratio of the background
FlowGRPO objective (\eqref{eq:grpoloss}) \\
\bar\rho_{\mathrm{GRPO}} & tempered (per-element-averaged) scoring-kernel ratio of
FewStepGRPO, a per-sample scalar equal to the $1/\Nel$ power of the corresponding Gaussian
density ratio (\eqref{eq:grpo-score-ratio}) \\
\rho_{\mathrm{AWM}} & detached residual-change statistic of FewStepAWM; it is
not an importance ratio (\eqref{eq:awm-statistic}) \\
\chi(z,A) & detached unclipped-gradient gate of
\eqref{eq:clip-gate}; its method-specific evaluations
$\chi_{\mathrm{GRPO}}$ and $\chi_{\mathrm{AWM}}$ are written explicitly in
\eqref{eq:grpo-score-ratio} and \eqref{eq:awm-statistic} \\
\mu_\theta,\ \mu_{\mathrm{old}},\ \sigma_{\mathrm{k}},\ \Delta t & means and width of the
Gaussian scoring kernel used by FewStepGRPO over a step of length
$\Delta t=t-r$ (\eqref{eq:grpo-v-mean}). The width $\sigma_{\mathrm{k}}$ is distinct from
the level of $x_\nu$ and the rollout renoise scale $\tau_j$ \\
\alpha_t,\ b_t & Euler--Maruyama coefficients of the scoring kernel,
$\mu=\alpha_t x_t-b_t v$ (\eqref{eq:grpo-v-mean}); distinct from the NFT
branch weights $a_\pm$ \\
\tau_{\mathrm{AWM}},\ \tau_{\min} & implemented FewStepAWM residual scale
$\tau_{\mathrm{AWM}}=\max\{t,\tau_{\min}\}$ with floor $\tau_{\min}>0$; it is
not a transition-kernel width \\
\varsigma,\ \varsigma_t & base noise scale and per-level noise scale of the scoring kernel of FewStepGRPO,
using the definition of \eqref{eq:grpo-varsigma}, which avoids the singularity at the source
endpoint $t_0=1$ by evaluating only the denominator at $t_1$ \\
a_\pm,\ \gamma_\pm & non-negative branch weights
$a_+=\hat r/\gamma_+$ and $a_-=(1-\hat r)/\gamma_-$, and strictly positive
adaptive normalisers $\gamma_\pm\ge\varepsilon_\gamma>0$, used by
FewStepNFT and FewStepOPSD (\eqref{eq:nft-gamma}) \\
\tilde x_\pm,\ \tilde x^{[m]}_\pm,\ h,\ \eta,\ \kappa,\ M_{\mathrm{tgt}} & ascent and descent
points, their iterates indexed by $[m]$, step size $h$, total step length $\eta$ in units of the trust radius,
relative trust radius $\kappa$, and positive-integer number of steps
$M_{\mathrm{tgt}}\in\mathbb N_{\ge1}$; $h,\eta,\kappa\ge0$
(\eqref{eq:opsd-target}) \\
\Delta_{\mathrm{rwd}}\tilde x_\pm,\ \Delta & FewStepOPSD branch increments
(\eqref{eq:opsd-increment}, \eqref{eq:opsd-decomp}) and the detached shift
$\Delta=\xoldu-\xind$ that transfers them to the v-form anchor
(\eqref{eq:opsd-shift}) \\
\lambda & loss weight of the ReFL objective (\eqref{eq:refl-loss}) \\
\xi,\ B,\ \bar B & standard Gaussian increment and forward/backward Brownian
motions of the SDEs in \Cref{app:grpo-kernel} \\
c_\theta,\ c_{\mathrm{old}},\ c_J,\ c_\phi,\ c_{\mathrm{ref}},\ c_{\mathrm{extra}},\ C,\ C_{\mathrm{rollout}} &
per-inner-iteration network-evaluation counters, the extra per-outer-iteration
cost, and the total and rollout cost of one outer iteration (\Cref{tab:cost}) \\
c_{\mathrm{adv}},\ n_{\mathrm{it}},\ n_{\mathrm{ts}},\ n_{\mathrm{fm}} &
positive clip half-range $c_{\mathrm{adv}}>0$ in \eqref{eq:rhat}, also used
as the explicit overall $W$ scale of the implemented NFT, general OPSD and
ReFL trainers; inner iterations per outer iteration, training
queries per member, and fitting steps of $\vph$ per outer iteration \\
H_c,\ \{R\}_{\mathcal B} & rewards of prompt $c$ in the current outer iteration and the current
global batch of rewards, the two populations of \eqref{eq:adv} \\
u_{\mathrm{ref}} & initial pre-RL average-velocity policy used as the KL
reference and kept fixed throughout training (\eqref{eq:kl-uref}); distinct
from the per-outer-iteration old policy $\uold$ \\
\lambda_{\mathrm{KL}},\ \mathcal L_{\mathrm{KL}}^{\mathrm u},\
\mathcal L_{\mathrm{KL}}^{\mathrm v} & weight, endpoint-renoise and replay
forms of the average-velocity penalty, and the instantaneous-reference form
(\eqref{eq:kl-uref}, \eqref{eq:kl-vref}) \\
w(t) & AWM's time weight  \\
\varepsilon_{\mathrm{std}},\ \varepsilon_{\mathrm{clip}},\ \varepsilon_{\mathrm{cd}},\
\varepsilon_\gamma,\ \varepsilon_g
& advantage standard-deviation floor (set to $10^{-4}$ in \eqref{eq:adv}), clipping radius, directional-difference step,
adaptive-normaliser floor and reward-gradient-norm floor;
$\varepsilon_{\mathrm{std}},\varepsilon_{\mathrm{cd}},\varepsilon_\gamma,\varepsilon_g>0$
and $\varepsilon_{\mathrm{clip}}\ge0$ \\
\sg(\cdot) & stop-gradient: identity forward, zero backward \\
\msp{\cdot} & \emph{per-sample mean}: the average over the $\Nel$ elements of a
single sample, $\msp{v}=\tfrac1{\Nel}\sum_{k=1}^{\Nel}v_k$, taken before any
batch reduction; squares inside it are elementwise, so
$\norm{v}^2=\Nel\,\msp{v^2}$. It is the reduction the implemented losses use in
place of $\norm{\cdot}^2$; the resulting per-sample losses are subsequently
averaged over the batch \\
\end{longtable}
\addtocounter{table}{-1}

\section{FewStepGRPO}\label{app:grpo}

Throughout this appendix, $(x_t,x_r)$ denotes a transition of the stochastic
sampler \eqref{eq:rollout-sde} replayed as in \S\ref{sec:fewstep-grpo}, where
$t>r$ are consecutive levels of the rollout grid, $\Delta t=t-r$, and
$s\in[0,t]$ is the lower endpoint of the flow map, redrawn at every inner
iteration. The
same estimate $J=\Jterm{\uold}{t}{s}$ is added to both means of
\eqref{eq:grpo-v-mean},
\begin{equation}
\mu_\theta=\alpha_t\,x_t-b_t\,\big(\uth(x_t,t,s)+J\big),\qquad
\mu_{\mathrm{old}}=\alpha_t\,x_t-b_t\,\big(\uold(x_t,t,s)+J\big),
\label{eq:app-grpo-means}
\end{equation}
where $J$ is evaluated by \eqref{eq:cd-replay}, with tangent
$(x_t-\xh)/t$. Squared norms are per-sample means $\msp{\cdot}$, as in Stage~4.

\subsection{Why there is no auxiliary-velocity variant}\label{app:grpo-noaux}

The auxiliary velocity cannot be used for the current prediction, because
substituting $\vph$ would give $\partial\xz/\partial\theta=0$
(\S\ref{sec:fewstep}); the trainers that use $\vph$ therefore use it only for
the baseline, $\xbase=\xaux$ (\eqref{eq:vpred-nft-phi},
\eqref{eq:dmdawm}). In FewStepGRPO, both anchors that an auxiliary field could
replace are already fixed, $\xreg=\varnothing$ and
$\xbase=\sg(\xz)=\sg(\mu_\theta)$ (\eqref{eq:grpo-decomp}), and
\S\ref{app:grpo-loss} shows that this choice reproduces the gradient of the
tempered Gaussian log-score of the action. Apart from the detached $J$
shared by both means, the old policy enters through
$\mu_{\mathrm{old}}$ in $\bar\rho_{\mathrm{GRPO}}$. An auxiliary variant would
therefore have to place $\vph$ inside the scoring kernel.

\emph{(a) $\vph$ as the old-policy field.} If
$\mu_{\mathrm{old}}$ were replaced by
$\mu_\phi=\alpha_t\,x_t-b_t\,\vph(x_t,t)$, then at
$\theta=\theta_{\mathrm{old}}$ the scoring ratio would compare the distances
of $\mu_\phi$ and $\mu_{\mathrm{old}}$ to $x_r$. The ratio would then generally differ
from one at zero policy drift, so the PPO clip would respond to the
approximation error of $\vph$ as well as to the change of the policy.

\emph{(b) $\vph$ as the source of the shared correction.} Replacing $J$ by
$\vph(x_t,t)-\uold(x_t,t,s)$ in both means of \eqref{eq:app-grpo-means} keeps
$\mu_\theta-\mu_{\mathrm{old}}=-b_t\,(\uth-\uold)$, hence
$\bar\rho_{\mathrm{GRPO}}=1$ at
$\theta=\theta_{\mathrm{old}}$, but it requires $\vph$ at the replayed state
$x_t$. By \eqref{eq:auxfit}, $\vph$ is fitted on re-noised states
$\AddNoise(\xh,t)=(1-t)\xh+t\epsilon$, with $\epsilon$ independent of
$\xh$, against the conditional velocity $\vcond=\epsilon-\xh$. A replayed state
is not of this form, because \eqref{eq:rollout-sde} adds fresh noise at every
step; hence $x_t$ and the endpoint $\xh$ of the same trajectory are not
connected by a straight path with independent noise $\epsilon$, and $\vcond$ is not defined at $x_t$. Exact transport at every
step would be sufficient for the two state marginals to agree, but the
finite-step map used here is not exact; the replay query can therefore differ
from the distribution on which $\vph$ is fitted. In an earlier run of
construction (b), not among
those of \S\ref{sec:exp}, which queried $\vph$ at the replayed states, the
evaluated OCR accuracy fell from $0.498$ to $0.016$ within $40$ iterations
while the loss \eqref{eq:auxfit} decreased from $0.354$ to $0.152$. This is
consistent with $\vph$ improving on its training distribution while being evaluated on a different
one. Querying $\vph$ at a re-noised state instead removes the mismatch, but
the correction is then evaluated at a state other than the one at which the
kernel is centred, and the scored density is no longer that of the transition
that was taken.

The trainers of \S\S\ref{sec:fewstep-awm}--\ref{sec:fewstep-nft} avoid both
problems. Their query state $\AddNoise(\xh,t)$ follows the training
distribution of $\vph$ (Stage~3\,\emph{(i)}), and $\vph$ enters only through their
baseline.
FewStepGRPO, whose kernel has to be evaluated at the states the sampler
visited, therefore uses only the shared-$J$ construction.

\subsection{The transition kernel}\label{app:grpo-kernel}

\paragraph{Replay trajectory.}
A state stored on the replay trajectory is not an endpoint-renoise state
$\AddNoise(\xh,t)$, so $\vcond$ is not its tangent. We instead use
$(x_t-\xh)/t$:
\begin{equation}
\Jterm{\uold}{t}{s}
=(t-s)\,
\frac{
\uold\!\big(x_t+\varepsilon_{\mathrm{cd}}\tfrac{x_t-\xh}{t},t+\varepsilon_{\mathrm{cd}},s\big)
-\uold\!\big(x_t-\varepsilon_{\mathrm{cd}}\tfrac{x_t-\xh}{t},t-\varepsilon_{\mathrm{cd}},s\big)}
{2\varepsilon_{\mathrm{cd}}}.
\label{eq:cd-replay}
\end{equation}
The result is detached and shared by both means. It costs two
no-grad forwards of $\uold$.

\paragraph{Boundary condition.}
The coefficients of \eqref{eq:grpo-v-mean} use the following implemented
noise scale on the rollout grid:
\begin{equation}
\varsigma_{t_j}
=
\begin{cases}
\displaystyle \varsigma\sqrt{\frac{t_0}{1-t_1}}, & j=0,\\[8pt]
\displaystyle \varsigma\sqrt{\frac{t_j}{1-t_j}}, &
j=1,\dots,M_{\mathrm{step}}-1 .
\end{cases}
\label{eq:grpo-varsigma}
\end{equation}
The coefficients $\alpha_t$ and $b_t$ retain the source level $t=t_j$; only the
singular denominator of $\varsigma_{t_0}$ is evaluated at $t_1$.

\paragraph{Derivation of \eqref{eq:grpo-v-mean}.} Write $\sigma$ for the
level along the linear path, on which the probability-flow ODE reads
$\mathrm dx_\sigma=v(x_\sigma,\sigma)\,\mathrm d\sigma$, with $v$ the marginal
velocity, and the score is
$\nabla_{x_\sigma}\log p_\sigma(x_\sigma)
=-\big(x_\sigma+(1-\sigma)\,v(x_\sigma,\sigma)\big)/\sigma$
(\S\ref{sec:prelim}, the paragraph on DMD). For any diffusion coefficient
$\varsigma_\sigma\ge0$, the forward-time SDE
\begin{equation}
\mathrm dx_\sigma=\Big[v(x_\sigma,\sigma)+\tfrac12\varsigma_\sigma^{2}
\nabla_{x_\sigma}\log p_\sigma(x_\sigma)\Big]\mathrm d\sigma
+\varsigma_\sigma\,\mathrm dB
\label{eq:app-fsde}
\end{equation}
has the same marginals as the probability-flow ODE: in its Fokker--Planck
equation, the term
$-\tfrac12\varsigma_\sigma^{2}\nabla\!\cdot(p_\sigma\nabla\log p_\sigma)$
contributed by the added drift and the diffusion term
$\tfrac12\varsigma_\sigma^{2}\nabla^{2}p_\sigma$ cancel, since
$p_\sigma\nabla\log p_\sigma=\nabla p_\sigma$. The time reversal of
\eqref{eq:app-fsde}, run from $\sigma=1$ to $\sigma=0$, is
\begin{equation}
\mathrm dx_\sigma=\Big[v(x_\sigma,\sigma)-\tfrac12\varsigma_\sigma^{2}
\nabla_{x_\sigma}\log p_\sigma(x_\sigma)\Big]\mathrm d\sigma
+\varsigma_\sigma\,\mathrm d\bar B,\qquad \mathrm d\sigma<0 .
\label{eq:app-rsde}
\end{equation}
One Euler--Maruyama step from $t$ to $r$, i.e.\ $\mathrm d\sigma=-\Delta t$, followed by
the substitution of the score, gives
\begin{equation}
\begin{split}
x_r&=x_t-\Delta t\,v-\frac{\varsigma_t^{2}\Delta t}{2t}\big(x_t+(1-t)v\big)+\varsigma_t\sqrt{\Delta t}\,\xi\\
&={\Big(1-\frac{\varsigma_t^{2}\Delta t}{2t}\Big)}x_t
-{\Big(1+\frac{\varsigma_t^{2}(1-t)}{2t}\Big)\Delta t}\,v
+\underbrace{\varsigma_t\sqrt{\Delta t}}_{\sigma_{\mathrm{k}}}\,\xi ,
\end{split}
\label{eq:app-em}
\end{equation}
with $\xi\sim\mathcal N(0,I)$, which is \eqref{eq:grpo-v-mean}. The
choice \eqref{eq:grpo-varsigma} of $\varsigma_t$, which avoids the singularity at the source endpoint, fixes the coefficients; at $t_0=1$
only the denominator of $\varsigma_{t_0}$ is evaluated at $t_1$, while $\alpha_t$
and $b_t$ retain the source level $t_0$. The transition density is
$\mathcal N(\alpha_t\,x_t-b_t\,v,\ \sigma_{\mathrm{k}}^{2}I)$, whose mean is affine in $v$; moreover,
$\sigma_{\mathrm{k}}>0$ on every transition of the grid.

\paragraph{Equivalence with the kernel of an instantaneous-velocity policy.}
For an average-velocity policy the MeanFlow identity, \eqref{eq:mfid}, gives
$\uth(x_t,t,s)+(t-s)\,\partial_t\uth(x_t,t,s)=v_\theta(x_t,t)$ for every
$s\in[0,t]$. If $J$ were computed from $\uth$, the velocity used in
\eqref{eq:app-em} would therefore be the instantaneous velocity of the policy,
the same for every $s$, and the transition density of FewStepGRPO would be that
of FlowGRPO for the policy $v_\theta$. Computing $J$ from $\uold$, as every
v-form trainer does
(Stage~3\,\emph{(ii)}), makes this exact at $\theta=\theta_{\mathrm{old}}$, up to the
error of \eqref{eq:cd-replay}; for $\theta\neq\theta_{\mathrm{old}}$ the field is
$v_\theta-(t-s)\,\partial_t(\uth-\uold)$, which depends on $s$ only through a
term of the order of the drift of the policy from the old policy. Two
consequences follow.
\begin{itemize}
\item $s$ enters the gradient through the entry of the flow map that is
differentiated, $\partial\mu_\theta/\partial\theta=-b_t\,\partial\uth(x_t,t,s)/\partial\theta$,
and the scored density only through that drift term. Redrawing $s\mid t$ at
every inner iteration therefore distributes the gradient of one transition over
the entries $\uth(\cdot,t,s)$, $s\in[0,t]$, as Stage~3\,\emph{(i)} does for the
trainers that re-noise, without changing what is scored at
$\theta=\theta_{\mathrm{old}}$.
\item For any objective $\mathcal L(\mu_\theta)$,
$\nabla_\theta\mathcal L=-b_t\,(\partial\uth/\partial\theta)^{\top}\partial\mathcal L/\partial\mu_\theta$,
since $J$ is detached (\eqref{eq:mf-jac}). This equals the FlowGRPO gradient
with $\partial v_\theta/\partial\theta$ replaced by
$\partial\uth(\cdot,t,s)/\partial\theta$.
\end{itemize}

\paragraph{The tempered scoring ratio.}
The ratio \eqref{eq:grpo-score-ratio} is the $1/\Nel$ power of the Gaussian
density ratio, equivalently the ratio at temperature $\Nel\sigma_{\mathrm{k}}^2$.

\paragraph{The tempered ratio at zero drift.} Since the same $J$ is added to both means,
$\mu_\theta-\mu_{\mathrm{old}}=-b_t\,(\uth-\uold)(x_t,t,s)$, and expanding the
two squares of $\log\bar\rho_{\mathrm{GRPO}}$ in
\eqref{eq:grpo-score-ratio} gives
\begin{equation}
\log\bar\rho_{\mathrm{GRPO}}
=-\frac{b_t}{\sigma_{\mathrm{k}}^{2}}\,\msp{(\uth-\uold)\odot(x_r-\mu_{\mathrm{old}})}
-\frac{b_t^{2}}{2\sigma_{\mathrm{k}}^{2}}\,\msp{(\uth-\uold)^{2}} .
\label{eq:app-logratio}
\end{equation}
Hence $\bar\rho_{\mathrm{GRPO}}=1$ at $\theta=\theta_{\mathrm{old}}$,
regardless of the error of $J$ and of the kernel that generated $x_r$, because
the same $J$ enters both means and the same $x_r$ enters both squares. The
ratio differs from one only through $\uth-\uold$.

\paragraph{Relation to the kernel that produced the action.} In the notation of
\eqref{eq:rollout-sde}, $x_t=\hat x_{t_j}$ and $x_r=\hat x_{t_{j+1}}$, and the
action was drawn by the flow-map step to the level $s_j$ drawn by the sampler,
$\hat x_{s_j}=\hat x_{t_j}-(t_j-s_j)\,\uold(\hat x_{t_j},t_j,s_j)$, followed by
$\hat x_{t_{j+1}}=\frac{1-t_{j+1}}{1-s_j}\,\hat x_{s_j}+\tau_j\epsilon'$. Given
an exact flow-map jump to $s_j$, this rollout kernel preserves the linear-path
marginal exactly: if
$\hat x_{s_j}=(1-s_j)x+s_j\epsilon$, then
$\hat x_{t_{j+1}}=(1-t_{j+1})x+\big(\tfrac{1-t_{j+1}}{1-s_j}s_j\epsilon+\tau_j\epsilon'\big)$,
a noise of variance $t_{j+1}^{2}$ by the definition of $\tau_j$. This kernel,
however, differs from \eqref{eq:app-em}. Its mean is
$\frac{1-t_{j+1}}{1-s_j}\big(\hat x_{t_j}-(t_j-s_j)\uold\big)$ and its width
is $\tau_j$, with $\tau_j=0$ whenever $s_j=t_{j+1}$, in which case the kernel is
deterministic and has no likelihood ratio. On the four-step grid, with $s_j$
drawn uniformly from the remaining grid levels, this happens for $52\%$ of the
transitions, including every final transition. FewStepGRPO therefore scores the
action under \eqref{eq:app-em}, for three reasons: $\sigma_{\mathrm{k}}>0$ on every
transition; the mean is affine in the instantaneous velocity, so that the
equivalence above holds; and the tempered ratio is exact at zero drift by
\eqref{eq:app-logratio}. The approximation lies in the scored transition.
Besides the noise $\tau_j\epsilon'$, both $x_r-\mu_{\mathrm{old}}$ in
\eqref{eq:app-logratio} and $x_r-\mu_\theta$ in the gradient of
\S\ref{app:grpo-loss} contain the deterministic difference between the means of
the two kernels. Consequently, $\bar\rho_{\mathrm{GRPO}}$ is a tempered ratio
of the surrogate scoring kernels and not an exact importance ratio for the
rollout kernel.

\subsection{The decoupled loss}\label{app:grpo-loss}

\paragraph{The decoupled loss reproduces the gradient of the tempered clipped
surrogate.} Per transition, define
\[
\mathcal L_{\mathrm{GRPO}}^{\mathrm{temp}}
:=-\min\{A\bar\rho_{\mathrm{GRPO}},\
A\clip(\bar\rho_{\mathrm{GRPO}},1\pm\varepsilon_{\mathrm{clip}})\}.
\]
Where the first branch attains the minimum, its gradient is
$-A\bar\rho_{\mathrm{GRPO}}\nabla_\theta\log\bar\rho_{\mathrm{GRPO}}$;
where the second does, it is constant in $\theta$ and contributes no gradient.
With the explicit gate $\chi_{\mathrm{GRPO}}$ of
\eqref{eq:grpo-score-ratio} and the per-element Gaussian log-score
$s_\theta^{\mathrm{GRPO}}
:=-\msp{(\mu_\theta-x_r)^{2}}/(2\sigma_{\mathrm{k}}^{2})$,
\begin{equation}
\nabla_\theta\mathcal L_{\mathrm{GRPO}}^{\mathrm{temp}}
=-\frac{\chi_{\mathrm{GRPO}}A\bar\rho_{\mathrm{GRPO}}}{\Nel\sigma_{\mathrm{k}}^{2}}
\left(\frac{\partial\mu_\theta}{\partial\theta}\right)^{\!\top}\big(x_r-\mu_\theta\big).
\label{eq:app-ppo-grad}
\end{equation}
Substituting \eqref{eq:grpo-decomp} into \eqref{eq:master} gives
$\mathcal L_{\mathrm{rwd}}=W\msp{\big(\xz-\sg(\xz)-g\big)^{2}}$ with $\xz=\mu_\theta$
and $g=\omega\,(x_r-\sg(\mu_\theta))$; since
$\xz-\sg(\xz)$ vanishes in value and has the identity as Jacobian,
\[
\nabla_\theta\mathcal L_{\mathrm{rwd}}
=-\frac{2W\omega}{\Nel}
\left(\frac{\partial\mu_\theta}{\partial\theta}\right)^{\!\top}\big(x_r-\mu_\theta\big),
\]
which, with $W=\bar\rho_{\mathrm{GRPO}}$ and
$\omega=\chi_{\mathrm{GRPO}}A/(2\sigma_{\mathrm{k}}^{2})$, is
\eqref{eq:app-ppo-grad} term by term. The equality holds for gradients; the
value of $\mathcal L_{\mathrm{rwd}}$ is $W\msp{g^{2}}$, as for every trainer
(Stage~4).

\paragraph{Justification of each factor.}
\begin{itemize}
\item \emph{The per-element mean.} A Gaussian log-density is a sum over the
$\Nel$ non-batch elements of one latent; taking it as the mean
$\msp{\cdot}$ divides the exponent by $\Nel$, which is why
$\bar\rho_{\mathrm{GRPO}}$ is the $1/\Nel$ power of the density ratio, or
equivalently a score at temperature $\Nel\sigma_{\mathrm{k}}^{2}$. Here
$\Nel=65{,}536$ for the image latent and
$\Nel=2{,}096{,}640$ for the video latent. Using the mean keeps the ratio within
floating-point range (with the sum, the exponent can overflow for
differences allowed by the clip), follows FlowGRPO, and is consistent with a
clipping radius of order $10^{-3}$.
\item \emph{The factor $1/(2\sigma_{\mathrm{k}}^{2})$ in $\omega$.} The reward term of
\eqref{eq:master} contributes $2W\omega/\Nel$ times the displacement, so
$\omega$ must contain the factor $1/(2\sigma_{\mathrm{k}}^{2})$ of the log-score for
\eqref{eq:app-ppo-grad} to hold. Without it, each transition would be weighted
by $2\sigma_{\mathrm{k}}^{2}$, which varies by a factor of $2.7$ across the four transitions of
the grid; a single learning rate cannot compensate for such a
transition-dependent scale.
\item \emph{$\xbase=\sg(\xz)$.} The score of the Gaussian at the action,
$(x_r-\mu_\theta)/\sigma_{\mathrm{k}}^{2}$, is measured from the current mean, so the origin of
the reward direction has to follow the policy. A fixed origin
$\mu_{\mathrm{old}}$, i.e.\ a translation as in DiffusionNFT, coincides with it only at
$\theta=\theta_{\mathrm{old}}$.
\item \emph{$\xreg=\varnothing$.} The self-regression role is played by the ratio and
its clip, and by the velocity penalty of \eqref{eq:kl-uref}; a self-regression term
would add to \eqref{eq:app-ppo-grad} a gradient that the policy-gradient
estimator does not contain.
\item \emph{Method-specific $c_{\mathrm{adv}}$ scaling.} The implemented NFT
and general OPSD trainers multiply both loss weights $W$ by
$c_{\mathrm{adv}}$ to undo the compression of \eqref{eq:rhat}; the factor is
written explicitly in \eqref{eq:nft-weights} and \eqref{eq:opsd-weights}.
ReFL keeps it although it fixes $\hat r\equiv1$. FewStepGRPO does not use it, since $\omega$ already contains
the gated signed advantage $\chi_{\mathrm{GRPO}}A$.
\end{itemize}

\section{FewStepOPSD}\label{app:opsd}

This appendix proves that, in the v-form of \S\ref{sec:FewStepOPSD}, the loss
and its gradient do not depend on the detached term $J$. After dropping $J$,
the current prediction and its anchors reduce to the $J$-free $(t,r)$ query of
\S\ref{sec:fewstep}, i.e., the average velocity queried at $(t,r)$ and
transported to the clean end without correction, while the reward-gradient steps still start from the $r=0$
prediction. We follow one member and suppress the index $(i)$.
Every prediction is evaluated at the query state $\xt$ of level $t$ and at the
$r$ drawn during the reward-coupled construction and reused at fitting
(\S\ref{sec:FewStepOPSD}, \emph{(i)}). Four properties of the implementation are
used:
\begin{enumerate}
\item[(a)] the same detached estimate $J=\Jterm{\uold}{t}{r}$ is added to every
prediction (Stage~3\,\emph{(iii)});
\item[(b)] the old policy $\uold$ is the same in the reward-coupled construction and
in fitting, being refreshed only after the policy update of the outer iteration
(Stage~1);
\item[(c)] the points $\tilde x_\pm$ are constructed by \eqref{eq:opsd-shift}; the gradient
steps start from the $r=0$ prediction, and only the resulting displacement is
added to the anchor;
\item[(d)] the velocity penalty, where present, is that of \eqref{eq:kl-uref}.
\end{enumerate}

\subsection{The quantities}\label{app:opsd-quantities}

By \eqref{eq:P-v} and \eqref{eq:opsd-v-decomp}, the current prediction and the anchors
are
\begin{equation}
\begin{aligned}
\xz&=\xt-t\big(\uth(\xt,t,r)+J\big),\\
\xreg=\tilde x_{\mathrm{base},1}=\tilde x_{\mathrm{base},2}=\xind
&=\xt-t\big(\uold(\xt,t,r)+J\big).
\end{aligned}
\label{eq:app-opsd-points}
\end{equation}
By (c), the iteration \eqref{eq:opsd-target} starts at $\xind+\Delta=\xoldu$,
the $r=0$ prediction of \eqref{eq:upred-old}; that is, $\tilde x_{\mathrm{old}}$
in \eqref{eq:opsd-target} is replaced by $\xoldu$ in the starting point
$\tilde x^{[0]}_\pm=\xoldu$, in the step length $h=\eta\kappa\norm{\xoldu}/M_{\mathrm{tgt}}$
and in the trust region, the ball of radius $\kappa\norm{\xoldu}$ around $\xoldu$.
The resulting points are $\tilde x_\pm=\tilde x^{[M_{\mathrm{tgt}}]}_\pm-\Delta$, so that their
increments over the anchor, in the notation of \eqref{eq:opsd-increment}, are
\begin{equation}
\Delta_{\mathrm{rwd}}\tilde x_\pm:=\tilde x_\pm-\xind
=\tilde x^{[M_{\mathrm{tgt}}]}_\pm-\xoldu .
\label{eq:app-opsd-incr}
\end{equation}

\subsection{Cancellation of \texorpdfstring{$J$}{J}}\label{app:opsd-cancel}

\textit{(1) the increments do not contain $J$.}
\begin{lemma}\label{lem:opsd-incr}
$\Delta_{\mathrm{rwd}}\tilde x_\pm$ is a function of $\xoldu$, of the reward
$R(\dec(\cdot),c)$ and of the constants $\eta,\kappa,M_{\mathrm{tgt}},\varepsilon_g$
alone; in particular it does not depend on $J$.
\end{lemma}
\begin{proof}
$\xoldu=\xt-t\uold(\xt,t,0)$ contains no $J$. If $\tilde x^{[m]}_\pm$ does
not, neither does the reward gradient evaluated at it, nor the step length $h$,
nor the centre and radius of the trust region, all of which are evaluated at
$\xoldu$; hence $\tilde x^{[m+1]}_\pm$ does not. By induction
$\tilde x^{[M_{\mathrm{tgt}}]}_\pm$ does not, and neither does
\eqref{eq:app-opsd-incr}.
\end{proof}
The shift $\Delta=\xoldu-\xind=t\big(\uold(\xt,t,r)+J-\uold(\xt,t,0)\big)$
does contain $J$; it enters $\tilde x_\pm$ only as the offset subtracted from
the end point of the iteration, and cancels in \eqref{eq:app-opsd-incr}.

\textit{(2) the self-regression residual.} By (a) the two copies of $J$
in \eqref{eq:app-opsd-points} are one detached vector, and
\begin{equation}
\xz-\xreg=-t\big(\uth-\uold\big)(\xt,t,r),
\label{eq:app-opsd-dev}
\end{equation}
which is \eqref{eq:opsd-v-lreg}.

\textit{(3) the weights.} The normalisers of \eqref{eq:nft-gamma},
evaluated at $\tilde x_\pm$ as in \S\ref{sec:FewStepOPSD}\,\emph{(iv)}, take the
branch residuals
\[
\xbase\pm\beta\big(\sg(\xz)-\xreg\big)-\tilde x_\pm
=\pm\beta\big(\sg(\xz)-\xreg\big)-\Delta_{\mathrm{rwd}}\tilde x_\pm
=\mp\beta t\big(\uth-\uold\big)-\Delta_{\mathrm{rwd}}\tilde x_\pm ,
\]
by $\xbase=\xind$, \eqref{eq:app-opsd-incr} and \eqref{eq:app-opsd-dev}. By
Lemma~\ref{lem:opsd-incr} neither term contains $J$; hence neither do
$\gamma_\pm$, $a_+=\hat r/\gamma_+$, $a_-=(1-\hat r)/\gamma_-$, nor the weights
$W$, $\omega_1$, $\omega_2$ of \eqref{eq:opsd-weights}. $\hat r$ depends only on
the rewards of the rollouts (Stage~2).

\textit{(4) the reward displacements.} From \eqref{eq:opsd-decomp},
\begin{equation}
g_1=\omega_1\big(\tilde x_+-\xind\big)=\omega_1\,\Delta_{\mathrm{rwd}}\tilde x_+,
\qquad
g_2=\omega_2\big(\tilde x_--\xind\big)=\omega_2\,\Delta_{\mathrm{rwd}}\tilde x_- ,
\label{eq:app-opsd-g}
\end{equation}
neither of which contains $J$.

\textit{(5) the Jacobian.} $J$ is detached, so by \eqref{eq:mf-jac}
$\partial\xz/\partial\theta=-t\,\partial\uth(\xt,t,r)/\partial\theta$.

\textit{(6) the loss and its gradient.} Each reward square of
\eqref{eq:opsd-decoupled} equals $W\msp{g_k^{2}}$ in value (Stage~4), so the
value of the loss is
\begin{equation}
\mathcal L=W\msp{(\xz-\xreg)^{2}}+W\msp{g_1^{2}}+W\msp{g_2^{2}},
\label{eq:app-opsd-value}
\end{equation}
and its gradient, which is \eqref{eq:opsd-gradx} written with the per-sample
mean, is by Steps~2--5
\begin{equation}
\nabla_\theta\mathcal L
=-\frac{2tW}{\Nel}
\left(\frac{\partial\uth(\xt,t,r)}{\partial\theta}\right)^{\!\top}
\Big[-t\big(\uth-\uold\big)
-\omega_1\Delta_{\mathrm{rwd}}\tilde x_+
-\omega_2\Delta_{\mathrm{rwd}}\tilde x_-\Big].
\label{eq:app-opsd-grad}
\end{equation}

\begin{proposition}\label{prop:opsd-J}
The value and the gradient of the v-form FewStepOPSD loss,
\eqref{eq:app-opsd-value} and \eqref{eq:app-opsd-grad}, do not depend on $J$.
In particular, they are those obtained with $J=0$.
\end{proposition}
\begin{proof}
Every factor of \eqref{eq:app-opsd-value} and \eqref{eq:app-opsd-grad} is one of
$\xz-\xreg$, $g_1$, $g_2$, $W$, $\omega_1$, $\omega_2$,
$\Delta_{\mathrm{rwd}}\tilde x_\pm$ and $\partial\xz/\partial\theta$, none of which
contains $J$ by Steps~1--5. The penalty of \eqref{eq:kl-uref} compares $\uth$ and
$u_{\mathrm{ref}}$ at the same $(\xt,t,r)$ and contains no $J$ either.
\end{proof}

\begin{corollary}\label{cor:opsd-uv}
Dropping $J$ in \eqref{eq:app-opsd-points} gives
\[
\begin{gathered}
\xz=\Xmap{\xt}{\uth(\xt,t,r)}{0},\qquad
\xreg=\xbase=\xoldur=\Xmap{\xt}{\uold(\xt,t,r)}{0},\\
\tilde x_\pm=\xbase+\Delta_{\mathrm{rwd}}\tilde x_\pm ,
\end{gathered}
\]
with $\Delta_{\mathrm{rwd}}\tilde x_\pm$ computed from $\xoldu$. These are the
$J$-free $(t,r)$ queries of \S\ref{sec:fewstep}; they coincide with the
u-form quantities of \S\ref{sec:FewStepOPSD}\,(ii) except that the current
prediction and its anchors are queried at $(t,r)$ instead of $(t,0)$; the
reward displacement is still computed from $\xoldu$. At $r=0$ the two
constructions coincide, since then $\xbase=\xoldu$ and $\Delta=0$. Freeing $r$
therefore trains every $\uth(\cdot,t,r)$ while the decoder and the reward
gradient are still evaluated at the clean prediction $\xoldu$; the u-form
trains only $\uth(\cdot,t,0)$.
\end{corollary}

\subsection{Remarks}\label{app:opsd-remarks}

\begin{remark}[The cancellation as a translation]
Steps~1--4 state that $J$ shifts the five points $\xz$, $\xreg$, $\xbase$,
$\tilde x_+$, $\tilde x_-$ by the same detached vector $-tJ$ (with
$\tilde x_\pm$ shifted through the offset $\Delta$ of \eqref{eq:opsd-shift}), whereas the loss depends
on them only through $\xz-\xreg$ and $\tilde x_\pm-\xbase$. The two quantities
of the construction that depend on an absolute position, the points at which
$\nabla R$ is evaluated and the radius of the trust region, are evaluated at
$\xoldu$, which does not depend on $J$.
\end{remark}

\begin{remark}[Where the cancellation fails]
(i)~If the gradient steps were started at $\xind$ itself instead of at
$\xoldu$ as prescribed by \eqref{eq:opsd-shift}, $\nabla R$ would be evaluated at, and the
radius computed from, points displaced by $-tJ$, and $J$ would remain in
$\Delta_{\mathrm{rwd}}\tilde x_\pm$; this is why \refex{ex:opsd}\,(iii)
starts the steps at the $J$-free prediction. (ii)~If the constructed points were left at the end point of the
iteration, $\tilde x^{[M_{\mathrm{tgt}}]}_\pm$, instead of being shifted back by
$\Delta$,
the differences $\tilde x_\pm-\xbase$ would contain $\Delta$ and hence $J$.
(iii)~The penalty of \eqref{eq:kl-vref} contains $J$ in its target. (iv)~An
old policy refreshed between point construction and fitting would make
$\xbase$ at fitting differ from the anchor around which $\tilde x_\pm$ were placed,
by a term that the cancellation does not remove.
\end{remark}

\begin{remark}[Why the argument does not extend to the other trainers]
In FewStepNFT, FewStepAWM and FewStepGRPO the rewarded point is a sample (the
rollout sample $\xh$ or the action $x_r$) and does not contain $J$, whereas the
baseline does. Dropping $J$ changes the reward displacement of FewStepNFT,
$g=\omega(\xh-\xind)$, by $-\omega\,tJ$ (Stage~3\,(iii)), and the residual
$x_r-\mu_\theta$ of FewStepGRPO by $-b_t\,J$. This shift can be large; for
MeanFlowNFT, the logged mean of $|tJ|$ is $2.7$ in the mixed-reward setting and
$3.6$ in the OCR setting, compared with a root-mean-square entry of the clean
latent of $1.06$ and $1.46$, respectively. In FewStepOPSD, both ends of every
reward displacement are constructed from the same anchor, and the gradient
steps start from the $J$-free prediction $\xoldu$ (\eqref{eq:opsd-shift});
together, these two properties give the cancellation. The same holds for ReFL,
whose single reward square is the first one of \eqref{eq:opsd-decoupled}.
\end{remark}

\subsection{Cost, and an empirical check}\label{app:opsd-cost}

$J$ requires one directional-difference estimate \eqref{eq:cd}, i.e.\ two no-grad
forwards of $\uold$. With $J$ retained, the v-form evaluates it twice per member and outer
iteration: in the reward-coupled construction, to form $\xind$, where the members are
processed in micro-batches of six members to bound the memory of the reward backward pass; and in
fitting, to form $\xz$ and $\xreg$. By \Cref{cor:opsd-uv} neither evaluation is
needed. \Cref{tab:cd} compares two runs of the v-form that differ only in this
respect, and the same pair for ReFL with $\nabla R$ of the full reward. The
training reward agrees to within $0.002$ for the v-form at every evaluated
iteration, and to within $0.006$ for ReFL at iteration $400$, below the $1\%$
(about $0.016$) by which two evaluation draws of one checkpoint may differ;
held-out OCR accuracy differs more.

\begin{table}[H]
\centering
\footnotesize
\setlength{\tabcolsep}{4pt}
\renewcommand{\arraystretch}{1.15}
\resizebox{\textwidth}{!}{%
\begin{tabular}{@{}llcccccc@{}}
\toprule
trainer & it. & training reward & HPSv2 & PickScore & CLIPScore & aesthetic & OCR \\
\midrule
\multirow{4}{*}{FewStepOPSD}
& 100 & 1.5692 / 1.5704 & 0.3520 / 0.3515 & 0.9205 / 0.9230 & 0.2968 / 0.2959 & 6.092 / 6.093 & 0.400 / 0.405 \\
& 200 & 1.6052 / 1.6055 & 0.3643 / 0.3633 & 0.9338 / 0.9356 & 0.3071 / 0.3066 & 6.068 / 6.054 & 0.433 / 0.441 \\
& 300 & 1.6258 / 1.6243 & 0.3705 / 0.3698 & 0.9410 / 0.9416 & 0.3144 / 0.3129 & 5.984 / 5.983 & 0.474 / 0.486 \\
& 400 & 1.6406 / 1.6409 & 0.3733 / 0.3734 & 0.9464 / 0.9458 & 0.3208 / 0.3217 & 5.929 / 5.906 & 0.550 / 0.493 \\
ReFL & 400 & 1.6166 / 1.6225 & 0.3651 / 0.3675 & 0.9355 / 0.9378 & 0.3160 / 0.3172 & 5.893 / 5.891 & 0.468 / 0.546 \\
\midrule
& & s/iter & sampling (s) & update (s) & & & \\
\cmidrule(lr){3-5}
FewStepOPSD & & 499 / 268 & 437.4 / 224.7 & 58.4 / 43.3 & & & \\
ReFL & & 303 / 219 & 240.0 / 174.6 & 60.4 / 44.2 & & & \\
\bottomrule
\end{tabular}}
\caption{Comparison of retaining $J$ and setting $J=0$ in FewStepOPSD and in
ReFL with $\nabla R$ of the full reward. For each trainer, the two runs differ
only in whether \eqref{eq:cd} is evaluated. Each cell
reads ``$J$ retained / $J=0$''; the runs with $J=0$ are those of
\S\ref{sec:exp-main}, and their entries are those of \Cref{tab:main},
\Cref{tab:clip-traj} and \Cref{fig:mix}. Top: the runs with $J$ retained are
evaluated offline ($256$ images, seed $12345$, DrawBench prompts, OCR prompts
for OCR accuracy), with the same protocol as the runs with $J=0$ at iteration $400$. Bottom:
cost per iteration, medians over the training logs on $8$ GPUs; the runs with
$J$ retained were not timed on the same machine as those with $J=0$.}
\label{tab:cd}
\end{table}

\section{Supplementary Experiments}\label{app:experiments}

\subsection{Shared hyperparameters}\label{app:hparams}

The following values are shared by every trainer of \S\ref{sec:exp-setup}
and \S\ref{sec:exp-setup-video} and complement the settings listed there.

\paragraph{Old-policy EMA.} The old policy $\uold$ is refreshed once per
outer iteration $i$ by
$\theta_{\mathrm{old}}\leftarrow\eta_i\,\theta_{\mathrm{old}}+(1-\eta_i)\,\theta$,
with the decay schedule of the official DiffusionNFT
implementation \citep{zheng2026diffusionnft}: the decay ratio
$\eta_i=\min\{10^{-3}i,\,0.5\}$ increases linearly with the outer iteration
$i$ to a cap of $0.5$. Since $\eta_i>0$ for $i\ge1$, the condition
$\theta_{\mathrm{old}}=\theta$ of \refex{ex:refl} holds only at the first
outer iteration.

\paragraph{Floors.} The floors are:
the residual-scale floor $\tau_{\min}=10^{-2}$ of FewStepAWM
(\eqref{eq:awm-statistic}), the directional-difference step
$\varepsilon_{\mathrm{cd}}=5\times10^{-3}$ (\eqref{eq:cd}), the adaptive-normaliser
floor $\varepsilon_\gamma=10^{-5}$ (\eqref{eq:nft-gamma}), the
reward-gradient-norm floor $\varepsilon_g=10^{-12}$ (\eqref{eq:opsd-target})
and the advantage standard-deviation floor $\varepsilon_{\mathrm{std}}=10^{-4}$
(\eqref{eq:adv}).

\paragraph{GenEval2.} Following MeanFlowNFT \citep{huang2026meanflownft}, we
adopt the official GenEval2 evaluation setting \citep{kamath2025geneval2}.

\begin{table}[H]
\centering
\scalebox{0.94}{%
\begin{minipage}{0.90\linewidth}
\centering
\footnotesize
\setlength{\tabcolsep}{3.5pt}
\renewcommand{\arraystretch}{1.08}
\sbox{\videosemanticbox}{%
\begin{tabular}{@{}lccccccccc|c@{}}
\toprule
method & OC & MO & HA & Co & SR & Sc$^{*}$ & AS$^{*}$ & TS$^{*}$ & OCs$^{*}$ & semantic \\
\midrule
FewStepGRPO & 90.33 & 82.25 & 93.10 & 88.23 & 81.43 & \textbf{53.20} & 72.81 & 62.42 & 69.02 & 76.98 \\
MeanFlowAWM & 90.36 & 82.04 & 93.60 & 88.67 & 81.24 & 52.96 & 72.84 & \textbf{62.52} & \textbf{69.13} & \textbf{77.04} \\
DMDAWM & 89.35 & 81.36 & 93.00 & 88.31 & \textbf{82.12} & 52.61 & 72.65 & 62.47 & 68.97 & 76.76 \\
MeanFlowNFT & 88.91 & 81.10 & \textbf{94.20} & \textbf{89.25} & 80.92 & 52.20 & 73.00 & 62.16 & 68.97 & 76.75 \\
DMDNFT & 90.00 & \textbf{84.02} & 93.00 & 88.37 & 80.38 & 50.95 & 72.01 & 62.15 & 68.81 & 76.63 \\
\midrule
FewStepOPSD & 90.38 & 81.07 & 92.00 & 88.16 & 80.11 & 52.24 & \textbf{74.53} & 61.58 & 68.71 & 76.53 \\
ReFL & \textbf{90.57} & 81.55 & 92.80 & 87.79 & 81.22 & 53.16 & 73.32 & 62.27 & 69.02 & 76.85 \\
\midrule
mean & 89.99 & 81.91 & 93.10 & 88.40 & 81.06 & 52.47 & 73.02 & 62.22 & 68.95 & 76.79 \\
spread (max$-$min) & 1.66 & \textbf{2.95} & 2.20 & 1.46 & 2.01 & 2.25 & 2.52 & 0.94 & 0.42 & 0.51 \\
\bottomrule
\end{tabular}}
\textbf{(a) Quality dimensions (Q)}\par\smallskip
\resizebox{\linewidth}{!}{%
\makebox[\wd\videosemanticbox][c]{%
\begin{tabular}{@{}lccccccc|c@{}}
\toprule
method & DD & SC$^{*}$ & MS$^{*}$ & TF$^{*}$ & BC$^{*}$ & AQ & IQ & quality \\
\midrule
FewStepGRPO & \textbf{59.62} & 97.38 & 95.85 & 96.67 & 95.07 & 69.97 & 69.54 & 85.27 \\
MeanFlowAWM & 58.89 & 97.52 & 96.04 & 96.83 & 95.35 & 70.02 & \textbf{69.65} & 85.36 \\
DMDAWM & 58.61 & 97.57 & 96.24 & 96.98 & 95.24 & 69.99 & 69.47 & 85.35 \\
MeanFlowNFT & 58.06 & 97.62 & 96.35 & 97.15 & 95.37 & 69.99 & 69.20 & 85.34 \\
DMDNFT & 53.33 & \textbf{97.83} & \textbf{97.21} & \textbf{97.44} & 95.27 & 69.77 & 68.75 & 85.07 \\
\midrule
FewStepOPSD & 59.44 & 97.57 & 96.02 & 96.82 & \textbf{95.68} & \textbf{70.87} & 68.21 & 85.37 \\
ReFL & 59.17 & 97.55 & 96.02 & 96.84 & 95.45 & 70.36 & 69.27 & \textbf{85.40} \\
\midrule
mean & 58.16 & 97.58 & 96.25 & 96.96 & 95.35 & 70.14 & 69.16 & 85.31 \\
spread (max$-$min) & \textbf{6.29} & 0.45 & 1.36 & 0.77 & 0.61 & 1.10 & 1.44 & 0.33 \\
\bottomrule
\end{tabular}}}

\bigskip
\textbf{(b) Semantic dimensions (S)}\par\smallskip
\resizebox{\linewidth}{!}{\usebox{\videosemanticbox}}

\bigskip
\textbf{(c) Overall}\par\smallskip
\resizebox{\linewidth}{!}{%
\makebox[\wd\videosemanticbox][c]{%
\begin{tabular}{@{}lcc|ccc|c@{}}
\toprule
& \multicolumn{2}{c|}{training reward} & \multicolumn{4}{c}{VBench} \\
method & sum & rank & quality & semantic & overall & rank \\
\midrule
FewStepGRPO & 15.89 & 7 & 85.27 & 76.98 & 83.62 & 4 \\
MeanFlowAWM & 16.66 & 4 & 85.36 & \textbf{77.04} & \textbf{83.70} & 1 \\
DMDAWM & 16.78 & 3 & 85.35 & 76.76 & 83.64 & 3 \\
MeanFlowNFT & 16.80 & 2 & 85.34 & 76.75 & 83.62 & 4 \\
DMDNFT & \textbf{17.93} & 1 & 85.07 & 76.63 & 83.38 & 7 \\
\midrule
FewStepOPSD & 16.26 & 6 & 85.37 & 76.53 & 83.60 & 6 \\
ReFL & 16.46 & 5 & \textbf{85.40} & 76.85 & 83.69 & 2 \\
\midrule
spread (max$-$min) & 2.04 & & 0.33 & 0.51 & 0.32 & \\
\bottomrule
\end{tabular}}}

\smallskip
{\raggedright\scriptsize Quality dimensions: DD dynamic degree, SC subject
consistency, MS motion smoothness, TF temporal flickering, BC background
consistency, AQ aesthetic quality, IQ imaging quality. Semantic dimensions: OC
object class, MO multiple objects, HA human action, Co color, SR spatial
relationship, Sc scene, AS appearance style, TS temporal style, OCs overall
consistency. $^{*}$~marks the eight dimensions rescaled by fixed min--max ranges
beforehand; their raw values differ substantially and are not comparable with
these. quality: the seven Q dimensions aggregated with DD at weight $0.5$ and the other six at weight $1$, divided by $6.5$ (the VBench convention); semantic:
the nine S dimensions; overall $=0.8\,$quality$\,+\,0.2\,$semantic.\par}
\end{minipage}}
\caption{All $16$ VBench dimensions at iteration $250$, normalised (\%), i.e.\
the values entering the aggregates of \Cref{tab:video-main}.
(a)~Quality and (b)~semantic dimensions, each with its aggregate and, below,
the mean and spread over the seven methods. (c)~Training reward against the
VBench aggregates. Only DD ($6.29$) and MO ($2.95$) have a spread above $2.6$;
the remaining dimension spreads are at most $2.52$. In the method rows, bold
marks the highest per column; in the spread rows, it marks values above $2.6$.}
\label{tab:video-dims}
\end{table}

\begin{table}[H]
\centering
\footnotesize
\setlength{\tabcolsep}{4pt}
\renewcommand{\arraystretch}{1.15}
\resizebox{\textwidth}{!}{%
\begin{tabular}{@{}lrcccccrrrrrr@{}}
\toprule
& & \multicolumn{4}{c}{HPSv3 at $M_{\mathrm{step}}$}
& \multicolumn{7}{c}{\% change at $M_{\mathrm{step}}=32$ against $M_{\mathrm{step}}=4$} \\
\cmidrule(lr){3-6}\cmidrule(l){7-13}
trainer & it. & 4 & 8 & 16 & 32
& HPSv3 & ImgRwd & GenEval2 & Aesth. & CLIP & HPSv2 & Pick \\
\midrule
FewStepGRPO & 400 & 11.58 & 11.15 & 10.49 & 10.09 & $-12.85$ & $-4.26$ & $-2.89$ & $-1.76$ & $+0.58$ & $-3.82$ & $-0.85$ \\
MeanFlowAWM & 100 & 12.30 & 12.36 & 12.17 & 11.85 & $-3.71$ & $-0.58$ & $-6.46$ & $-2.98$ & $-0.08$ & $+0.57$ & $+0.07$ \\
DMDAWM & 400 & 13.10 & 13.24 & 12.98 & 12.77 & $-2.53$ & $+1.50$ & $+6.02$ & $-2.87$ & $-2.47$ & $+0.20$ & $+0.19$ \\
MeanFlowNFT & 300 & 13.26 & 13.18 & 12.96 & 12.85 & $-3.08$ & $-1.67$ & $+6.81$ & $-1.27$ & $-0.11$ & $+0.27$ & $+0.15$ \\
DMDNFT & 400 & 13.49 & 13.49 & 13.34 & 13.13 & $-2.67$ & $+0.08$ & $+5.59$ & $-2.09$ & $-0.66$ & $-0.24$ & $-0.05$ \\
\midrule
FewStepOPSD & 400 & 13.06 & 13.27 & 13.29 & 13.19 & $+0.97$ & $+0.12$ & $+1.65$ & $-0.08$ & $-0.68$ & $-0.01$ & $+0.03$ \\
ReFL & 200 & 13.80 & 13.50 & 13.12 & 12.87 & $-6.76$ & $-1.55$ & $+0.05$ & $-1.38$ & $-0.58$ & $-2.97$ & $-0.74$ \\
\bottomrule
\end{tabular}}
\caption{Test-time scaling in the mixed-reward setting. Left: HPSv3, a
held-out evaluator with a clear signal, at each step count. Right: the
percentage change of every evaluated dimension between $M_{\mathrm{step}}=4$
and $32$. CLIPScore, HPSv2 and PickScore are the three summands of the training
reward; HPSv3, ImageReward, GenEval2 and the aesthetic score are held out.
The GenEval2 column is reported for
completeness only, since its largest change from $4$ to $32$ steps ($0.028$) and
its largest intermediate change ($0.032$) are within or at the ${\sim}0.03$
resolution stated in \S\ref{sec:exp-setup} for $256$ images.
Trainer labels are as in \Cref{tab:arms}; ``it.'' is the checkpoint used.}
\label{tab:tts}
\end{table}

\begin{figure}[H]
\centering
\includegraphics[width=\textwidth]{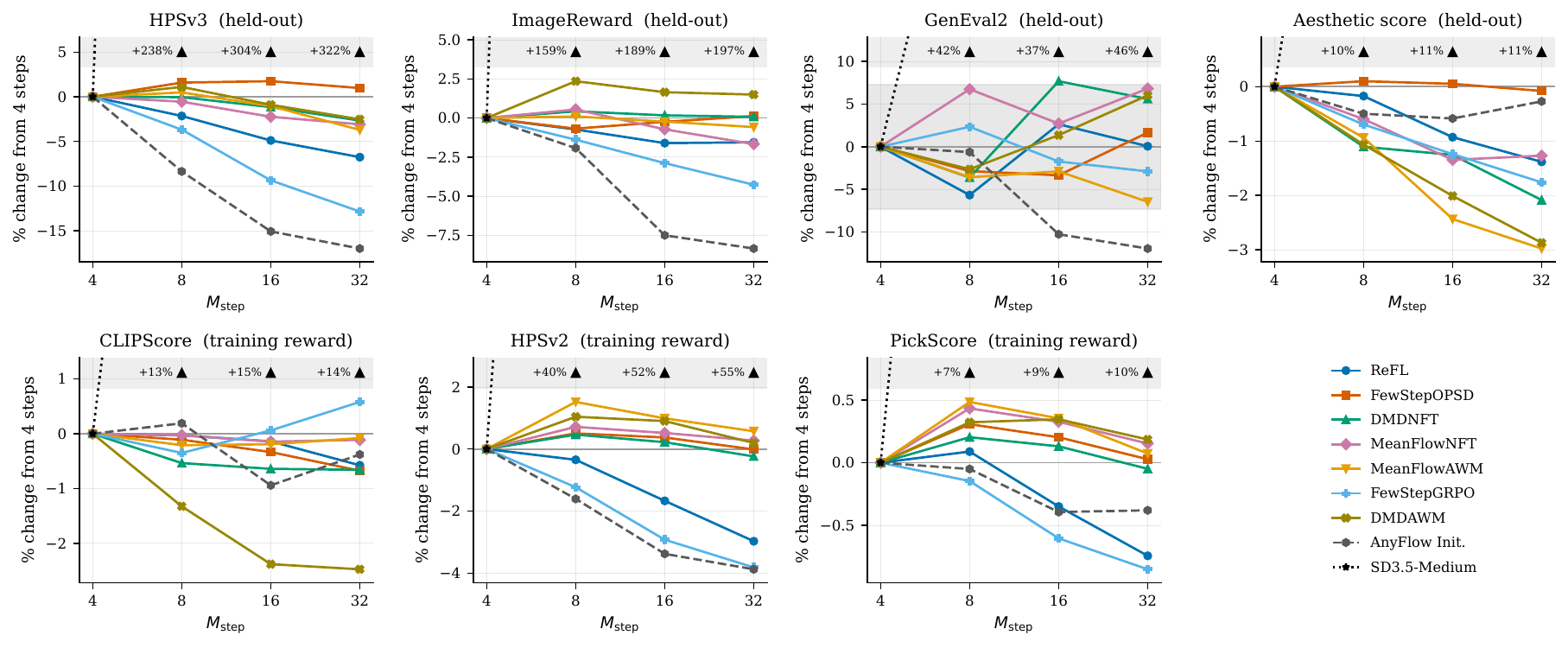}
\caption{Percentage change of each evaluated dimension from its four-step
value against $M_{\mathrm{step}}$, one panel per evaluator and one line per
model, including the AnyFlow initialisation and the pretrained SD3.5-Medium.
Values of SD3.5-Medium outside the plotted range are printed at the top of each
panel. The shaded band in the GenEval2 panel marks the ${\sim}0.03$ resolution
of \S\ref{sec:exp-setup}; the changes of the fine-tuned models stay within or
at this resolution, whereas both references move beyond it.}
\label{fig:tts-trend-rel}
\end{figure}

\begin{figure}[H]
\centering
\includegraphics[width=\textwidth]{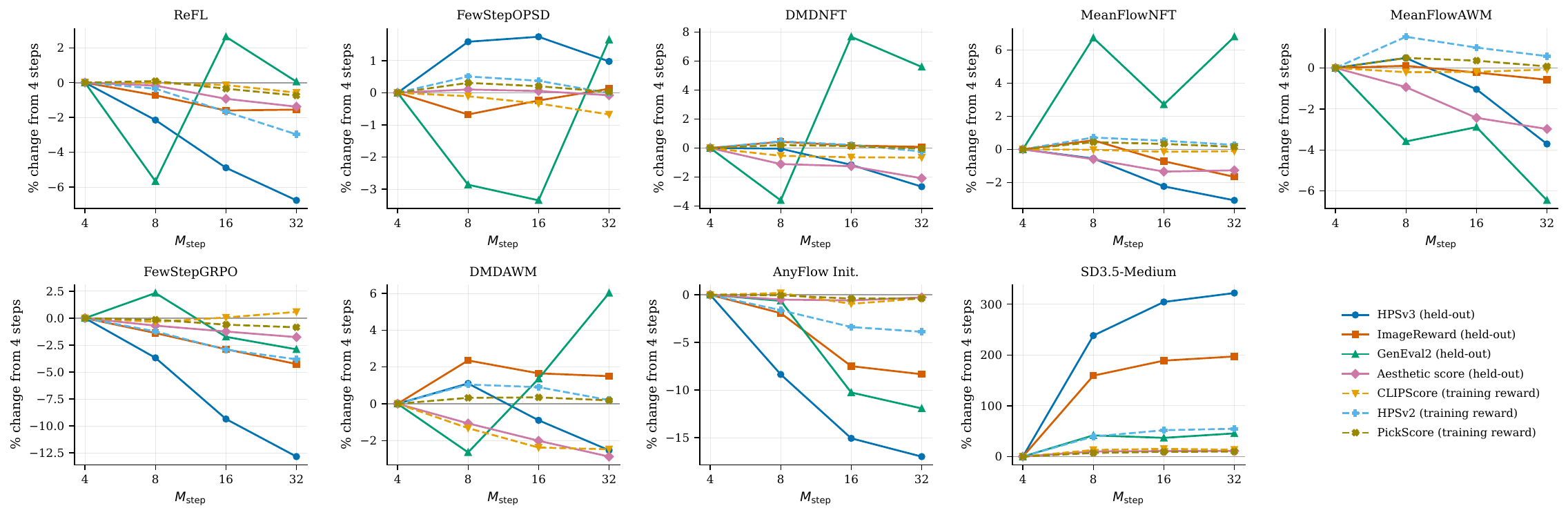}
\caption{The percentage changes of \Cref{fig:tts-trend-rel}, one panel per
model and one line per evaluator. Solid lines are held-out evaluators, and
dashed lines are the summands of the training reward.}
\label{fig:tts-trend-arm}
\end{figure}

\subsection{Analysis of no gain in CLIPScore for group-advantage runs}\label{app:clipscore}
In the mixed-reward setting, \emph{all five} default group-advantage runs end
below the pre-RL CLIPScore. The two runs with the full reward gradient instead
increase CLIPScore at every listed $100$-iteration checkpoint, to $0.3217$ and
$0.3172$. In contrast, all three CLIP-focused MeanFlowNFT ablations in the lower
block of \Cref{tab:clip-traj} end above the pre-RL value.

The training reward weights its three summands equally, but CLIPScore varies
little within a prompt group, where the advantage is computed; at the pre-RL
policy, its within-group standard deviation is only about $0.01$ ($0.010$ and
$0.014$ on the first iteration of the two CLIPScore-only runs below), roughly a
quarter of its global-batch standard deviation and $3.5$--$5\%$ of its value. It
accounts for $30.4\%$ of the variance of the three summands, against $53.6\%$
for PickScore and $16.0\%$ for HPSv2, so an equal weight does not give it an
equal share of the signal.

We therefore tested three ways of changing that share, the first two on
MeanFlowNFT (\Cref{tab:clip-traj}, lower block).
\emph{(i)}~Raising the weight of CLIPScore: the weights
$(w_{\mathrm{CLIP}},w_{\mathrm{HPS}},w_{\mathrm{Pick}})=(2.4,0.3,0.3)$, i.e.\
$8{:}1{:}1$, and $(3.0,0,0)$, CLIPScore alone, the limit of an unbounded
relative weight; both keep $K=24$.
\emph{(ii)}~Enlarging the group: CLIPScore alone with $K=72$ images
per prompt instead of $24$.
\emph{(iii)}~Removing it from the gradient: ReFL differentiating HPSv2 alone.
The first two increase CLIPScore, and the increase grows with its share; by
iteration $400$, it gains $0.0133$ at $8{:}1{:}1$, $0.0191$ with CLIPScore
alone, and $0.0363$ with $K=72$, more than either gradient-based trainer
($0.0286$--$0.0331$), although this run draws three times as many images per
prompt. In all three runs, however, the other two summands stop improving, and
HPSv2 and PickScore oscillate within $[0.2473,\,0.2974]$ and
$[0.8459,\,0.8691]$, the range they occupy at the start of training; under
CLIPScore alone the held-out aesthetic score and OCR accuracy also decrease, by
$0.64$--$0.98$ and $0.18$--$0.19$. Conversely, ReFL with the
gradient of HPSv2 alone leaves CLIPScore near its pre-RL value ($0.2844$--$0.2883$ over
iterations $100$--$400$) while raising HPSv2 to $0.3946$, the highest of any
reported variant.

In these ablations, increasing CLIPScore under a scalar advantage coincides
with the gains of the other summands weakening or vanishing. The scalar is
shared according to the covariance of each summand with the group reward, so
reweighting redistributes the observed signal. The reward gradient instead
contains a direction from every summand; in the reported FewStepOPSD and ReFL
runs, CLIPScore, HPSv2 and PickScore rise together by
$0.0286$--$0.0331$, approximately $0.0705$--$0.0764$, and approximately
$0.0738$--$0.0818$, respectively.
As noted in the main text, this approach requires a reward that is
differentiable with respect to the clean sample.

\begin{table}[tbp]
\centering
\footnotesize
\setlength{\tabcolsep}{4.5pt}
\renewcommand{\arraystretch}{1.15}
\begin{tabular}{@{}lccccc@{}}
\toprule
& \multicolumn{5}{c}{CLIPScore at iteration} \\
\cmidrule(l){2-6}
method & 0 & 100 & 200 & 300 & 400 \\
\midrule
FewStepOPSD, $\nabla R$          & 0.2886 & 0.2959 & 0.3066 & 0.3129 & 0.3217 \\
ReFL, $\nabla$ of $R$            & 0.2886 & 0.2967 & 0.3076 & 0.3127 & 0.3172 \\
ReFL, $\nabla$ of HPSv2          & 0.2886 & 0.2845 & 0.2844 & 0.2863 & 0.2883 \\
\midrule
\multicolumn{6}{l}{MeanFlowNFT, weights $w_{\mathrm{CLIP}}{:}w_{\mathrm{HPS}}{:}w_{\mathrm{Pick}}$} \\
\midrule
$1{:}1{:}1$ (default), $K=24$                  & 0.2886 & 0.2800 & 0.2815 & 0.2806 & 0.2814 \\
$8{:}1{:}1$ $(2.4,0.3,0.3)$, $K=24$           & 0.2886 & 0.2914 & 0.2963 & 0.3010 & 0.3019 \\
CLIPScore only $(3.0,0,0)$, $K=24$             & 0.2886 & 0.2948 & 0.3029 & 0.3049 & 0.3077 \\
CLIPScore only $(3.0,0,0)$, $K=72$             & 0.2886 & 0.3031 & 0.3094 & 0.3182 & 0.3249 \\
\bottomrule
\end{tabular}

\caption{CLIPScore against the iteration in the mixed-reward setting, the
summand of the training reward on which the two groups of trainers disagree.
Upper block: the update direction contains $\nabla R$. Lower block: the reward
enters as the scalar $A$ or $\hat r$. Evaluation protocol as in
\S\ref{sec:exp-setup}; the HPSv2-only ReFL row uses a different training
reward.
Iteration $0$ is the shared pre-RL checkpoint, CLIPScore $0.2886$.}
\label{tab:clip-traj}
\end{table}

\paragraph{Conclusion.} In the reported group-advantage ablations, raising the
share of CLIPScore or enlarging the group raises CLIPScore while the gains of
the other summands are reduced or vanish. In these runs, the weighting of the
summands therefore affects which rewards improve, not only how fast they
improve. Reward-gradient-based methods provide a direction from every summand,
and all three summands rise together in the reported runs; however, these
methods apply only when the reward is differentiable with respect to the clean
sample, which excludes the OCR setting (\S\ref{sec:exp}).

\subsection{Why FewStepOPSD resembles ReFL}\label{app:opsd-refl}

In FewStepOPSD, the signed reward coefficients satisfy
$|\omega_1|+|\omega_2|=1/\beta$; $\hat r$ redistributes this magnitude between
the ascent and descent terms and can also change their common weight $W$
(\eqref{eq:opsd-weights}). Both terms move the prediction in the direction of
increasing reward, because $g_1=\omega_1\Delta_{\mathrm{rwd}}\tilde x_+$ and
$g_2=\omega_2\Delta_{\mathrm{rwd}}\tilde x_-$ with $\omega_2\le0$, while the two
increments follow $+\nabla R$ and $-\nabla R$, respectively. The logged mean
magnitudes $|\omega_1|=5.12$ and $|\omega_2|=4.88$ differ by $0.24$, i.e.\ $2.4\%$ of their fixed sum $1/\beta=10$; they do
not by themselves determine the imbalance of the two reward-augmentation
vectors. FewStepOPSD is therefore close to the ReFL reward augmentation of
\refex{ex:refl}, with an added self-regression, a descent term and several
gradient steps; this is consistent with the similar curves in \Cref{fig:mix}.

\subsection{The velocity penalty: KL-u against KL-v}
\label{app:ablations}\label{app:klablation}

The velocity penalty, labelled KL in the run names although it is a squared
velocity residual (\S\ref{sec:impl-stages}), admits two constructions: the fixed initial
average-velocity reference of \eqref{eq:kl-uref}, which compares $\uth$ and
$u_{\mathrm{ref}}$ at the same $(\xt,t,r)$ (KL-u), and the frozen
instantaneous teacher of \eqref{eq:kl-vref}, converted through
$J[\uold,t,r]$, which depends on the old policy (KL-v). The variants therefore differ in both
reference construction and parameterisation space, and
the weight is $10^{-4}$ in both; the main text reports KL-u under the name MeanFlowNFT.

In the mixed-reward setting, the two variants are close at iteration $400$;
both reach a training reward of $1.509$, and the held-out evaluators differ by
about $1$--$1.5\%$ in either direction (HPSv3 $13.23$ against $13.36$,
ImageReward $1.385$ against $1.365$), with neither variant ahead on both.
GenEval2 is $0.407$ against $0.400$ at iteration $400$. Its average over
iterations $100$--$400$ is $0.393$ against $0.365$; the $0.028$ gap is near the
stated ${\sim}0.03$ resolution, so we do not interpret it as a clear
difference.

In the OCR setting, KL-u reaches OCR accuracy $0.85$ earlier, at iteration
$100$ instead of $150$, with $0.785$, $0.850$, $0.945$
at iterations $50$, $100$, $150$ (the values of \Cref{tab:main}a) against $0.693$, $0.834$, $0.897$. At matched
OCR accuracy the difference is mixed; at the first saved checkpoint with OCR
accuracy of at least $0.82$, HPSv3 is $10.64$ against $10.24$ and ImageReward
$0.887$ against $0.896$, each variant ahead on one of the two. At this weight,
the choice of reference therefore has a limited effect in these runs; it changes the iteration
at which the OCR threshold is reached but has no clear effect on the quality at that point, and
both variants have nearly the same cost per iteration ($435$ against $436$\,s in the
mixed-reward setting). The main text therefore reports only KL-u.

\subsection{Partial differentiability of the video reward}
\label{app:exp-video-reward}

This restriction does not occur in the image setting and affects only
FewStepOPSD and ReFL.

Every method is \emph{evaluated} on the same four-term reward. A method that
requires $\nabla R$ can only use the differentiable terms, which on this
backbone are HPSv3-general ($39\%$ of the sum) and HPSv3-percentile ($57\%$);
VideoAlign MQ ($6\%$) and TA ($-2\%$) are not differentiable. The FewStepOPSD
and ReFL runs in \Cref{tab:video-main} use the gradients of both HPSv3
reductions, together $96\%$ of the aggregate under the logged normalisation,
but no gradient of motion quality or text alignment. On SD3.5, the gradient
covers $100\%$ of the training reward, since all three summands are
differentiable image scorers.

This restriction is not visible in the reported scalar rewards, since the
reward, the online evaluation and the advantage of FewStepOPSD all use the full
four-term scorer (ReFL has no advantage). The restriction affects the direction
of the gradient and not necessarily the norm of the update.

The released \texttt{bfloat16} HPSv3 scorer changes the reward value by only
${\sim}0.003$ relative to fp32 but rotates its gradient by $76.2$ degrees on
average. The runs in \Cref{tab:video-main} therefore use fp32 with the
\texttt{sdpa} attention backend, so this bf16 error does not affect the
reported results. One remaining constraint is that the clip is decoded one
latent frame at a time to limit memory; since the VAE is causal, a frame
decoded alone is treated as the first frame of a clip. Compared with a joint
decode of the same two frames, the gradient deviates by $79.2$ degrees, and the
gradient of the second frame is nearly orthogonal to the joint-decode reference.

{The FewStepOPSD and ReFL rows of \Cref{tab:video-main} therefore use the
gradient of the two HPSv3 terms with per-frame decoding, and not the gradient of
the full reward.} Removing the causal-decode constraint requires a joint window
of at least two frames, which needs $126.1$\,GB of GPU memory without
activation offloading on a $144$\,GB card with $25.6$\,GB already in use, or
$189$\,s per iteration with offloading. Across $48$ GPUs and $250$ iterations this
amounts to $26.25$ GPU-days (about $0.547$ days of wall-clock time if fully
parallel). This limitation comes from the memory of a single GPU and not
from the method.

\end{document}